\documentclass{article} 
\usepackage{iclr2027_conference,times}

\usepackage{amsmath,amsfonts,bm}

\def\eqref#1{equation~\ref{#1}}

\def\1{\bm{1}}

\DeclareMathAlphabet{\mathsfit}{\encodingdefault}{\sfdefault}{m}{sl}
\SetMathAlphabet{\mathsfit}{bold}{\encodingdefault}{\sfdefault}{bx}{n}

\usepackage{bbm}
\usepackage{amssymb}
\usepackage{hyperref}
\usepackage{url}
\usepackage{booktabs}
\usepackage{tabularx}
\usepackage{caption}
\usepackage{enumitem}
\usepackage{array}          
\usepackage{booktabs}       
\usepackage{multirow}       
\usepackage[table]{xcolor} 
\usepackage{colortbl}
\usepackage{adjustbox}
\usepackage{fontawesome5}
\usepackage{graphicx}
\usepackage{wrapfig}
\usepackage{tcolorbox}
\usepackage{comment}
\usepackage{pifont}
\usepackage{longtable}
\usepackage{amsmath}
\usepackage{siunitx}
\usepackage{cleveref}
\crefname{section}{\S}{\S\S}
\Crefname{section}{\S}{\S\S}

\definecolor{darkergreen}{RGB}{198,230,198}
\definecolor{lightgreen}{RGB}{230,244,230}
\definecolor{darkerred}{RGB}{255,204,204}
\definecolor{lightred}{RGB}{255,230,230}

\definecolor{deltared}{HTML}{9C1F14}
\definecolor{deltagreen}{HTML}{176B3A}
\definecolor{deltagray}{HTML}{666666}

\definecolor{deltaredbg}{RGB}{252,232,230}
\definecolor{deltagreenbg}{RGB}{230,242,232}

\providecommand{\posdelta}[1]{}
\providecommand{\negdelta}[1]{}
\providecommand{\zerodelta}[1]{}

\renewcommand{\posdelta}[1]{%
  \textcolor{deltagreen}{\ensuremath{+}#1}}
\renewcommand{\negdelta}[1]{%
  \textcolor{deltared}{\ensuremath{-}#1}}
\renewcommand{\zerodelta}[1]{%
  \textcolor{deltagray}{\ensuremath{+}#1}}
\providecommand{\hyp}[1]{\textit{#1}}

\newcommand{\sepwithci}[3]{%
  \ensuremath{#1_{\scriptscriptstyle[#2,\,#3]}}%
}

\newtcolorbox[auto counter]{promptbox}[2][]{
  colback=gray!5,
  colframe=gray!50!black,
  fonttitle=\bfseries,
  title={Prompt \thetcbcounter: #2},
  boxrule=0.5pt,
  arc=2pt,
  left=6pt, right=6pt, top=4pt, bottom=4pt,
  #1
}

\definecolor{ppl}{HTML}{C2185B}

\title{What Makes Something Hard(er)? Explaining Question Difficulty in Natural Language}

\author{%
\mdseries
\begin{tabular}[t]{@{}p{0.5\textwidth}@{}l@{}}
\textbf{Peng Cui}\thanks{Equal contribution.} & \textbf{Qiaoyuan Zheng}\footnotemark[1] \\
Department of Computer Science & Department of Computer Science \\
ETH Zurich & ETH Zurich \\
\texttt{pencui@ethz.ch} & \texttt{zqiaoyuan@ethz.ch} \\[1em]
\textbf{Rudolf Debelak} & \textbf{Mrinmaya Sachan} \\
Machine Learning and Optimization Lab & Department of Computer Science \\
EPFL & ETH Zurich \\
\texttt{rudolf.debelak@epfl.ch} & \texttt{msachan@ethz.ch}
\end{tabular}%
}

\begin{document}

\maketitle
\lhead{Preprint}

\begin{abstract}
Difficulty is one of the most fundamental properties of a question: it determines whether the question can meaningfully discriminate between models of differing ability. 
Although a variety of methods can now estimate or predict difficulty automatically, they yield only a single descriptive number, with no account of the underlying factors that make a question difficult in the first place. 
In this work, we propose a data-driven approach that automatically generates and validates natural-language hypotheses explaining what makes one question harder than another. 
We first estimate each item's difficulty from the responses of a large pool of LLMs using Item Response Theory. 
We then sample contrasting sets of easy and hard questions and prompt an LLM to propose candidate explanations of the difference, which are subsequently validated and selected on held-out questions. Experimental results across three datasets spanning mathematical, logical, and commonsense reasoning show that our method produces interpretable and predictive hypotheses. 
On their own, they predict the difficulty of unseen questions competitively with, or better than, advanced black-box difficulty regressors; used as additional features, they further improve those regressors, implying that they discover difficulty signals that existing models fail to capture. 
Moreover, we demonstrate that editing questions according to a hypothesis can shift their measured difficulty in the expected direction, indicating that the discovered hypotheses are causally valid difficulty factors rather than post-hoc descriptions. 
Our approach thus turns a purely descriptive difficulty score into actionable statements.
\end{abstract}

\vspace{-1ex}
\begin{center}
\faGithub~ \href{https://github.com/pengcuix/explain-difficulty}{\texttt{github.com/pengcuix/explain-difficulty}}
\end{center}

\section{Introduction}
The rapid growth of LLM capabilities has led to saturation on existing benchmarks \citep{akhtar2026when}, driving continuous benchmark iteration: harder, more challenging datasets are needed to keep discriminating between increasingly capable models \citep{rein2024gpqa,wang2024mmlupro,white2025livebench}. 
Yet what determines the difficulty of a question in the first place remains poorly understood.
Prior work on question difficulty has largely focused on \textit{measuring} difficulty and \textit{predicting} difficulty. 
Measurement typically relies on the responses of test-takers. 
The representative approach in this field is the Item Response Theory (IRT) family of models \citep{hambleton2013item,hambleton1991fundamentals}, which decomposes observed performance into item properties and test-taker ability.
IRT has increasingly been adopted for LLM evaluation \citep{zouhar-etal-2025-select,cui-sachan-2025-investigating}. 
However, it only yields a descriptive number per item but does not reveal the underlying factors that make an item hard in the first place. 

Some follow-up work \citep{fischer1973linear,janssen2004models,boeck2004explanatory} attempted to represent an item's difficulty as a linear combination of hand-specified item features, such as the cognitive operations it requires. 
This turns difficulty from a free parameter into an outcome of question properties, and thereby shifts the question from measuring difficulty to predicting and explaining it.
A line of feature-based difficulty prediction followed across domains \citep{benedetto2023survey,alkhuzaey2024text}.
Such features, however, must be designed by domain experts and coded by hand, which makes these approaches hard to scale across domains or to large question pools. 
More recent work instead predicts difficulty with neural network or pre-trained LLMs that read the item difficulty directly \citep{question2017huang,benedetto2020r2de,loginova2021towards}. 
While these methods usually enjoy stronger performance and greater flexibility  they predict difficulty from representations that are themselves uninterpretable, and therefore provide little insight into what makes a question difficult.

On the other hand, a recent line of work uses LLMs to generate natural-language explanations for surface regularities in unstructured data \citep{zhou-etal-2024-hypothesis,pmlr-v162-zhong22a,zhong2023goal,zhong2024explaining,hollmann2023large}.
These methods typically group data that are likely to share a property, and then prompt an LLM to describe a pattern the group has in common. 
So far, these methods have mostly been applied to properties that are semantically explicit and directly observable in the text, for example, the topic, stance, or style (e.g., \textit{"is about sports"}, \textit{"is written in the first person"} \citep{pmlr-v162-zhong22a}. 
Such properties are relatively easy to discover. 
In this work, we target \textit{difficulty}, a property that is latent rather than annotated, continuous rather than categorical, and the product of many interacting factors, and that is correspondingly harder to explain. 
We first estimate item difficulty with an IRT model from the responses of a large pool of LLMs.
We then repeatedly sample contrasting data of easy and hard questions and present them to an LLM to propose plausible explanations of the difference. 
We explore various sampling strategies to make the discovery more effective, and validate and select the most informative hypotheses on a held-out set of questions\footnote{We defer a detailed discussion of related work to Appendix \ref{related_work}.}. 

We apply our approach to generate hypotheses for various benchmarks, including mathematical reasoning, logical deduction, and commonsense reasoning. 
Experimental results show that our approach can generate higher-quality and valid explanations than other baseline generators.
Individually, our generated explanations separate difficulty more sharply than explanations from baseline approaches, and collectively they predict difficulty of unseen questions as well as or better than finetuned difficulty predictors and few-shot LLM judges. 
Moreover, incorporating the generated explanations as features into these predictors further improves the performance, indicating that they capture difficulty-related information the existing models can not. 
Finally, we use the hypotheses to curate synthetic questions: we edit questions so that they satisfy or violate a given hypothesis and observe that their measured difficulty shifts in the expected direction. 
This demonstrates that the discovered hypotheses are not merely post-hoc descriptions but carry causal effect, which yields practical guidance for designing and iterating on benchmarks.

\section{Method}\label{method}

\begin{figure}[t]
\centering
\includegraphics[width=\linewidth]{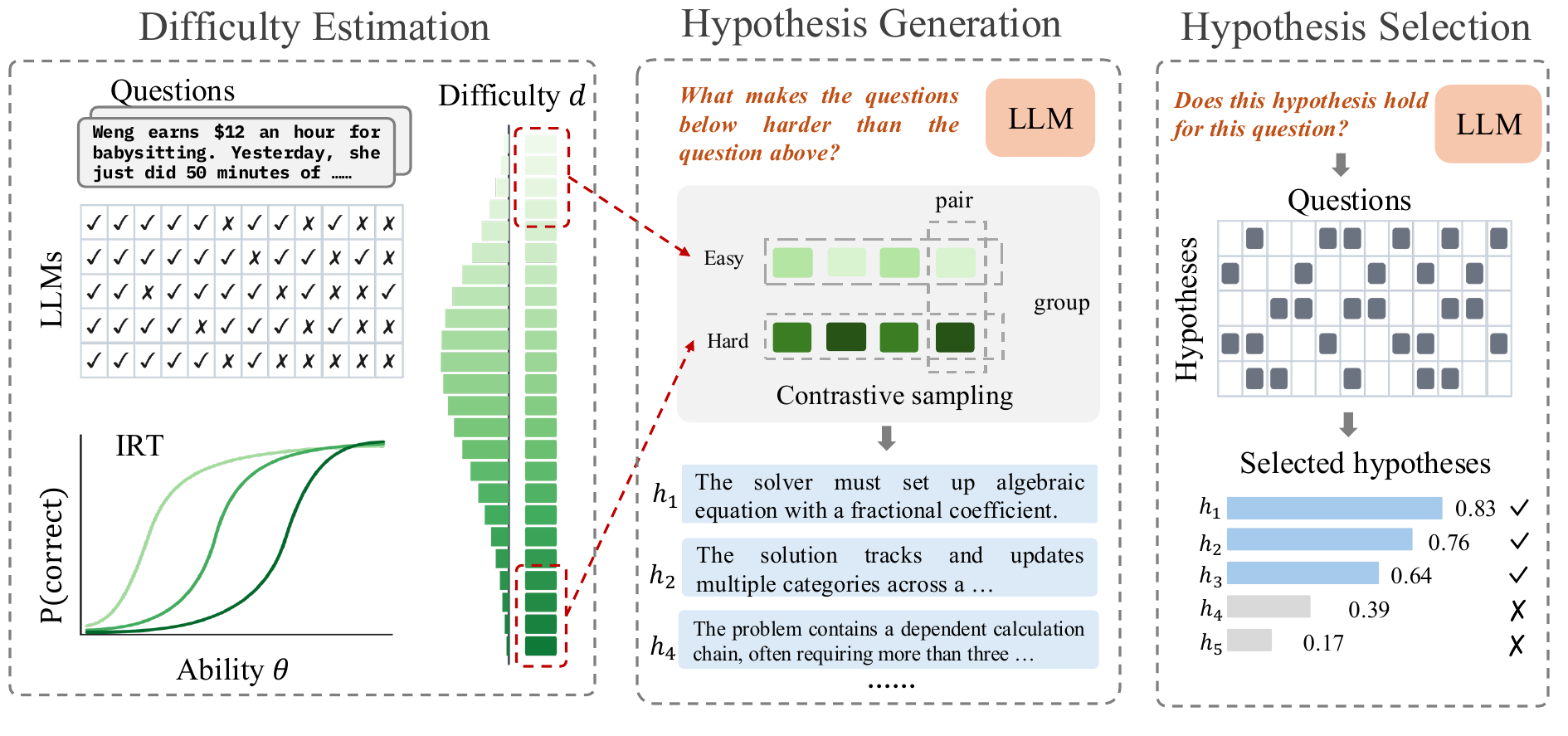}
\caption{We first estimate question difficulty from LLMs' responses using an IRT model. We then present an LLM with contrasting pairs or groups of hard and easy questions and prompt it to propose candidate hypotheses that explain the difference between them. Finally, we retain the hypotheses that best explain the difficulty of held-out questions.}
\label{fig:main} \label{fig:method}
\end{figure}

Our method consists of three steps. 
First, we use IRT to estimate each question's difficulty. 
Second, we construct contrastive samples of hard and easy questions and prompt LLMs to generate candidate hypotheses explaining the difficulty gap. 
Third, we refine and select from these candidate hypotheses. 
The full pipeline is illustrated in Figure \ref{fig:method}.

\subsection{Question Difficulty Estimation}
Let $\mathcal{D}$ be a dataset of $N$ questions. 
To measure the difficulty of each question in $\mathcal{D}$, we first collect $M$ LLMs' responses to the questions, forming a response matrix $R \in \{0,1\}^{M \times N}$, where each entry $R_{ij}=1$ indicates that the $i$-th model correctly answered the $j$-th question, and $0$ otherwise.

We then use this matrix $R$ to fit an IRT model, which provides an invariant and noise-aware measurement of item difficulty. 
Specifically, we use the one parameter logistic (1PL, Rasch) model \citep{rasch1993probabilistic}, where each item's difficulty $b_j$ and each test-taker's (LLM's) ability $\theta_i$ are learned jointly:
\begin{equation}
    P(R_{ij}=1) = \frac{1}{1+{\rm exp}(-(\theta_i - b_j))}.
\end{equation}

\subsection{Hypothesis Generation}
Our core idea is to present LLMs with \textit{contrastive} sets of difficult and easy items, and instruct an LLM to generate hypotheses explaining the difficulty gap between them.
We split the dataset into three disjoint sets: a generation set $\mathcal{G}$ used to generate candidate hypotheses, a verification set $\mathcal{V}$ to verify and select them, and a test set $\mathcal{T}$ for the final evaluation.
Each set is constructed via difficulty-stratified sampling, ensuring coverage across the full difficulty range.
We repeat the sample–generate procedure for multiple rounds on $\mathcal{G}$ to collect a sufficient pool of candidate hypotheses.

\paragraph{Contrastive Sampling.}
We sample either \emph{group-wise} or \emph{pair-wise}. 
For group-wise sampling, we first cut $\mathcal{G}$ into $10$ difficulty-rank buckets $ \mathcal{B}_{\{1\cdots10\}}$ of comparable size.
At round $t$, we first draw a bucket combination $(\mathcal{B}_{h}, \mathcal{B}_e )$ with
\begin{equation}
    w_t(\mathcal{B}_h, \mathcal{B}_e) \;\propto\; (h-e)^{\rho}\,\big(1+n^{<t}_{(\mathcal{B}_h,\mathcal{B}_e)}\big)^{-\gamma_c},
    \qquad h-e \ge \delta , \label{eq:sample_bucket}
\end{equation}
where $w(\cdot)$ is the sampling weight, $h, e$ index the high and low difficulty buckets, $n_{(\cdot)}^{<t}$ denote the number of times it has been sampled in earlier rounds and $\delta$ is the minimum gap.
$\rho$ tilts the draw toward larger difficulty gaps, which are otherwise underrepresented because few bucket combinations are far apart, and $\gamma_c$ discourages combinations already used. 
We then draw questions independently from each selected bucket with weights $w_t(q) \;\propto\;(1+n_q^{<t})^{-\gamma_i} \label{eq:sample_in_bucket}$ to down-weight questions selected in previous rounds, resulting in two groups of questions $\mathcal{Q}_{h}$ and $\mathcal{Q}_l$. 

For pair-wise sampling, we draw $k$ pairs ($q_h$, $q_e$) from $\mathcal{G}$ with $b(q_h)-b(q_e)\geq\Delta_{\min}$ from all candidate pairs that have not been used in this round, with
\begin{equation}
    w_t(q_h,q_e) \;\propto\; \big(1+n^{<t}_{q_h}\big)^{-\gamma_i}\big(1+n^{<t}_{q_e}\big)^{-\gamma_i}\big(1+n^{<t}_{(q_h, q_e)}\big)^{-\gamma_p},
\end{equation}
which down-weights both previously selected questions and pairs to promote diversity.

\paragraph{Generation.}
Accordingly, we construct two forms of prompt, one by pair and one by group. 
For the pair form, we present the sampled question pairs $(q_h, q_e)$ to an LLM (\textsc{Gemini-3.1-pro-preview}) directly and instruct it to explain the difficulty gap within each individual pair. 
For the group form, We prompt the LLM with $\mathcal{Q}_h$ and $\mathcal{Q}_l$ to generate more general hypotheses that explain the difference between the two groups and cover as many questions as possible. 
We combine the two strategies to discover difficulty factors at different granularities.

For the pair form, we show 5 pairs at a time;
for the group form, we use a larger window of 40 questions at a time (20 hard and 20 easy); 
For both prompting formats, we provide the question text, its reference answer, and its difficulty $b$ for every question shown.
Although the two forms are intended to capture different granularities, we explicitly instruct the model in both cases to avoid generating overly specific hypotheses.
See our prompt templates in Appendix \ref{app:prompt}.

\subsection{Hypothesis Selection}\label{sec:method_selection}
\paragraph{Refinement.} 
We repeat the above sample-and-propose process for $K$ rounds and obtain a list of candidate hypotheses. 
We compute pairwise similarities between hypotheses using cosine
similarity over \textsc{Qwen3-Embedding-4B} embeddings. 
Hypotheses whose
similarity exceeds 0.85 are grouped and passed to an LLM for deduplication and refinement: 
given a group of semantically similar or overlapping hypotheses, the LLM either retains those that are semantically distinct or merges near-equivalent ones into a single new hypothesis.

\paragraph{Match.} We first prompt a Judge LLM (\textsc{Gemini-3.7-Flash}) to annotate, for every refined hypothesis, whether each question satisfies it, yielding three binary matching matrices $U^{\mathcal{G}} \in \{0, 1\}^{|\mathcal{G}| \times |\mathcal{H}|}, U^{\mathcal{V}} \in \{0, 1\}^{|\mathcal{V}| \times |\mathcal{H}|}$ and $ U^{\mathcal{T}} \in \{0,1\}^{|\mathcal{T}| \times |\mathcal{H}|}$, where $U_{ij}=1$ indicates that the $i$-th question satisfies the $j$-th hypothesis, and $0$ otherwise. 
These matrices will be used for the subsequent selection and evaluation. 

\paragraph{Selection.}
Finally, we select the most informative hypotheses from the refined candidates $\mathcal{H}$ on the validation set $\mathcal{V}$ using $L_1$-regularized regression.
We fit a separate LASSO on each of 100 random subsamples containing 75\% of $\mathcal{V}$, with the L1 penalty weight chosen by 5-fold cross-validation within each subsample, and retain the top hypotheses by selection frequency.

\section{Experimental Setup}
\subsection{Datasets}

We evaluate our method on three popular benchmarks across various tasks: 
(1) \textbf{GSM8K} \citep{cobbe2021gsm8k} ($N=1,319$) is a widely used benchmark of grade-school math word problems that require multi-step arithmetic reasoning;  
(2) \textbf{BBH-structured} ($N=1,396$) comprises six subtasks in the BBH dataset \citep{suzgun-etal-2023-challenging} that require logical reasoning: \textit{"logical deduction (three, five, and seven)"}, \textit{"reasoning about colored objects"}, \textit{temporal sequences}, and \textit{"penguins in a table"};
(3) \textbf{Winogrande} \citep{sakaguchi2021winogrande} ($N=1,267$) is a dataset to test commonsense reasoning with pronoun resolution problems while minimizing the surface-level statistical biases present in the original Winograd Schema Challenge \citep{winograd2012}. 
The difficulty of this dataset is known to be difficult to predict \citep{ding2024easy2hard}.
We use it as a challenging testbed to test the limit of baselines and our method. 
We collect LLM response records from \textsc{RouterEval} \citep{huang-etal-2025-routereval}, including 5,000 LLMs for GSM8K and WinoGrande, and 3,811 LLMs for BBH-structured.
We fit a 1PL IRT model implemented in py-irt \citep{Lalor_2023}, which estimates the model by stochastic variational inference.

\subsection{Evaluation}\label{sec:metrics}
We evaluate the quality of the generated hypotheses along two dimensions: 
\begin{itemize}[leftmargin=15pt]
    \item \textit{Predictive effect} -- to what extent do they explain and predict the difficulty of unseen questions? 
    We evaluate the selected hypotheses in two ways. The first is an \emph{in-sample} evaluation, in which we use OLS to fit difficulty directly on the test pool $\mathcal{T}$ using its matching matrix ${U}^{\mathcal{T}}$. We report the adjusted $R^2$ \citep{ezekiel1930methods}, which corrects for the upward bias of $R^2$ as the number of predictors grows, to measure the goodness of fit of the hypotheses as features and their explanatory power.  
    The second is an \emph{out-of-sample} evaluation, in which we fit an OLS model on ${U}^{\mathcal{G}\cup \mathcal{V}}$ and predict the difficulty of the questions in $\mathcal{T}$. 
    This setting tests whether the features can predict the difficulty of new questions.

    \item \textit{Causal effect} -- does editing a question according to a hypothesis change its difficulty in the expected direction? For a given hypothesis $h$, we edit questions according to $h$, modifying only the attribute that $h$ describes while keeping the rest of the question unchanged wherever possible. 
    We check whether the measured difficulty shifts significantly in the direction predicted by $h$ using \emph{difficulty shift} ($\Delta b_{h}$) and \emph{success rate} ($\mathrm{SR}_{h}$), defined as:
    \begin{equation}
        \Delta b_{h} = \mathbb{E}_{q \sim \mathcal{D}_h}[b(q'_h) - b(q)], \quad
        \mathrm{SR}_h = \mathbb{E}_{q \sim \mathcal{D}_h}\!\left[\mathbf{1}\!\left[b(q'_h) > b(q)\right]\right], \label{eq:casual_eval}
    \end{equation}
    where $\mathcal{D}_h$ is the set of questions that can be meaningfully edited according to $h$, $b(\cdot)$ is a question's difficulty, $q'_h$ is the edited question. While $\Delta b_h$ captures the magnitude of the effect, $\mathrm{SR}_h$ measures the fraction of edits that move difficulty in the expected direction.
\end{itemize}

\subsection{Baselines}\label{sec:baselines}
We compare our method with the following hypothesis generation baselines:
\begin{itemize}[leftmargin=15pt]
    \item \textbf{{\textsc{NLParam}}} \citep{zhong2024explaining} learns natural-language predicates as model parameters by optimizing a continuous relaxation of each predicate in embedding space and then prompting an LLM to verbalize it, iteratively replacing the least useful predicate. We use its classification variant with the logistic layer replaced by linear regression on IRT difficulty.
    \item \textbf{{\textsc{HypotheSAEs}}} \citep{pmlr-v267-movva25a} applies SAEs to turn the embeddings into sparse interpretable features which are then are selected via Lasso and verbalized into a hypothesis. 
    We obtain embeddings from the hidden-state activations of \textsc{Gemma-3-4B}; because the evaluation set we use is not sufficient to finetune an SAE, we use the pretrained SAEs from Gemma-Scope-2, and have an LLM (\textsc{Gemini-3.7-Flash}) refine the verbal description of each selected feature using the questions from our dataset on which it activates. 

    \item \textbf{{\textsc{HypoGeniC}}} \citep{zhou-etal-2024-hypothesis} generates hypotheses from labeled examples with an LLM and improves them iteratively using a bandit-style reward. As \textsc{HypoGeniC} is defined for classification, we adapt it to difficulty by discretizing IRT difficulty into 3 buckets and treating level prediction as the classification task. 

    \item \textbf{{\textsc{Edisa}}} \citep{hua2023discover} is a slice detection model that discovers coherent groups of error-prone examples by fitting a mixture Gaussian model using embedding, label, and error distance. We adapt it to our setting by replacing the error distance of a single model with the IRT difficulty, and by verbalizing each discovered slice with an LLM prompted with its representative members. We set the number of slices to the hypothesis budget $\mathcal{H}$, and use the same embeddings as for \textsc{Qwen3-Embedding-4B}.
\end{itemize}
In addition, to measure the predictive power of the generated hypotheses when used as features, we compare our method against a range of difficulty predictors: 
(1) \textbf{Fine-tuned models}, for which we use RoBERTa \citep{liu2019roberta}, a competitive model for this task as reported in \cite{yaneva-etal-2024-findings}; 
(2) \textbf{Frozen representations} with a trained adapter, for which we take the last-token representation of \textsc{Gemma-3-4B} \citep{team2025gemma} and \textsc{Qwen3-Embedding-4B} \citep{zhang2025qwen3} and train a Lasso regressor on top (a dense regressor would otherwise overfit as the dimension of embeddings is greater than the number of examples); 
(3) \textbf{Few-shot prompting} of frontier LLMs, for which we use \textsc{Gemini-3.1-pro-preview} and \textsc{Claude-Sonnet-5}. 
We split the difficulty range into 5 equal-frequency buckets and randomly draw 2 questions with their answers from each bucket as demonstrations; 
(4) \textbf{Hypothesis-based predictors}, for which we first use a judge model to determine whether each question satisfies each hypothesis as in Section \ref{sec:method_selection}, and then train an OLS regressor on these binary features\footnote{More details on the datasets and the implementation of all methods are provided in Appendix \ref{app:implementation}.}.

\begin{table*}[t]
\caption{
Top hypotheses by our method on three datasets.
${\rm SEP}$ is the difference in difficulty between matched and unmatched test items (95\% bootstrap CI); Cohen's $d$ is the standardized effect size, where $^\ddagger$ marks a large effect ($d \ge 0.8$) and $^\dagger$ a medium effect ($d \ge 0.5$) \citep{cohen2013statistical}; Coverage is the matched-item percentage. The full set of hypotheses by all methods are in Appendix \ref{app:all_hypotheses}.
}
\label{tab:top5-selected-hypotheses-three-datasets}
\centering
\scriptsize
\setlength{\tabcolsep}{3pt}
\renewcommand{\arraystretch}{1.14}

\begin{tabular}{cp{8.2cm}ccc}
\toprule
\textbf{\#} &
\multicolumn{1}{c}{\textbf{Top-5 Generated Hypotheses}} &
\(\mathbf{SEP}_{\scriptscriptstyle[95\%\,\mathrm{CI}]}\) &
\textbf{Cohen's $d$} &
\textbf{Cov. (\%)} \\
\midrule
\multicolumn{5}{c}{\textbf{GSM8K}} \\
\midrule

1 &
\hyp{The problem requires multistep bookkeeping in which intermediate results feed subsequent calculations.} &
\sepwithci{1.498}{1.043}{1.959} &
0.899$^\ddagger$ &
51.0 \\

2 &
\hyp{The solver must derive a rate, ratio, or percentage rather than directly apply one.} &
\sepwithci{2.035}{1.341}{2.742} &
1.174$^\ddagger$ &
9.0 \\

3 &
\hyp{The solver must set up and solve an algebraic equation rather than perform only sequential arithmetic.} &
\sepwithci{1.428}{0.520}{2.320} &
0.790$^\dagger$ &
4.0 \\

4 &
\hyp{The reference answer contains a logical error or contradiction that conflicts with the correct solution.} &
\sepwithci{3.696}{3.043}{4.351} &
2.063$^\ddagger$ &
1.0 \\

5 &
\hyp{The solver must aggregate multiple time intervals or convert between time units to obtain a final duration.} &
\sepwithci{1.057}{-0.238}{2.519} &
0.584$^\dagger$ &
6.5 \\

\midrule
\multicolumn{5}{c}{\textbf{BBH-structured}} \\
\midrule

1 &
\hyp{The solver must order at least five entities from relational constraints or attributes.} &
\sepwithci{1.429}{1.157}{1.695} &
1.419$^\ddagger$ &
23.6 \\

2 &
\hyp{The solver must aggregate or transform multiple elements rather than retrieve one stated value.} &
\sepwithci{0.449}{0.192}{0.720} &
0.387 &
27.5 \\

3 &
\hyp{The answer choices contain many plausible alternatives rather than a small, easily separable set.} &
\sepwithci{0.645}{0.336}{0.953} &
0.566$^\dagger$ &
29.3 \\

4 &
\hyp{The requested relative position or rank depends on reconstructing the full ordering.} &
\sepwithci{1.345}{1.029}{1.643} &
1.264$^\ddagger$ &
16.8 \\

5 &
\hyp{The solver must filter a set and then aggregate or count the remaining elements.} &
\sepwithci{0.768}{0.332}{1.238} &
0.664$^\dagger$ &
8.2 \\

\midrule
\multicolumn{5}{c}{\textbf{Winogrande}} \\
\midrule

1 &
\hyp{Resolving the blank requires a multistep causal chain with an unstated intermediate state or action.} &
\sepwithci{0.916}{0.446}{1.386} &
0.559$^\dagger$ &
26.4 \\

2 &
\hyp{Resolving the blank requires implicit physical or spatial commonsense about object interactions.} &
\sepwithci{0.739}{0.221}{1.258} &
0.445 &
20.9 \\

3 &
\hyp{The sentence describes interpersonal causality and requires separating roles such as cause and experiencer.} &
\sepwithci{0.742}{0.199}{1.287} &
0.447 &
19.7 \\

4 &
\hyp{The sentence contains a contrast or concession whose resolution crosses the contrastive boundary.} &
\sepwithci{0.596}{0.062}{1.177} &
0.356 &
17.3 \\

5 &
\hyp{A contrastive or concessive marker requires inverting the expected relationship or attribute.} &
\sepwithci{0.111}{-0.507}{0.771} &
0.066 &
13.8 \\

\bottomrule
\end{tabular}

\end{table*}

\section{Hypothesis Evaluation}
\subsection{Main Results}

\paragraph{What makes a question hard?}
We present top hypotheses generated by our method in Table \ref{tab:top5-selected-hypotheses-three-datasets}.
Every hypothesis is plausible and interpretable. Most of them concern the cognitive demands that the corresponding task actually imposes on the solver---the operations that have to be carried out and the information that has to be held and manipulated---rather than surface features of the text such as question length.
On GSM8K, our method also captures issues of data quality: items whose reference answer is itself flawed, which affect only a small fraction of the data but carry a very large difficulty effect. 
In terms of the statistical measures, the $95\%$ CI of ${\rm SEP}$ has a lower bound above zero for 13/15 hypotheses, showing that they genuinely discriminate between hard and easy items. 
Except on Winogrande, whose difficulty is known to be hard to perceive even for humans \citep{ding2024easy2hard}, 9/10 hypotheses on the two other datasets reach at least a medium effect$^\dagger$, and half of them reach a large effect$^\ddagger$.

\begin{figure}[t]
\centering
\includegraphics[width=\linewidth]{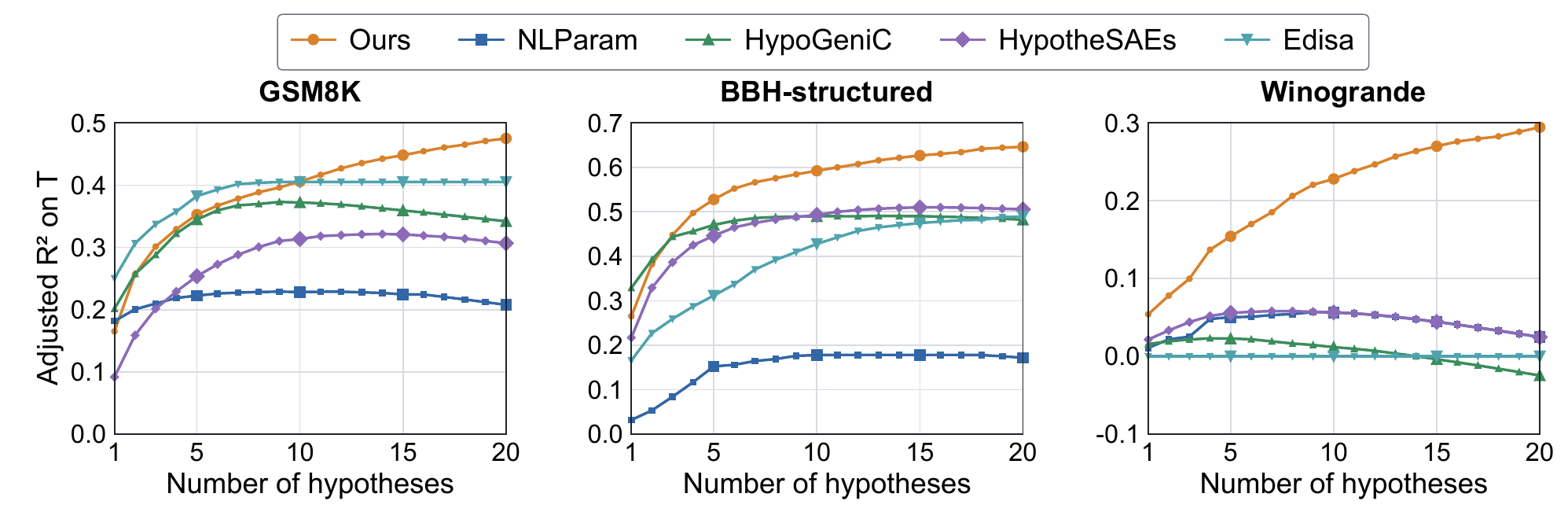}
\caption{We fit the test set using hypotheses generated by different methods as features to test their explanatory power, and report the adjusted $R^2$ for varying numbers of hypotheses.
We rank the features by greedily picking, at each step, the hypothesis that yields the largest incremental $R^2$.
}
\label{fig:main} \label{fig:in_dist_r2}
\end{figure}

\paragraph{Our approach discovers more effective hypotheses.} 
We fit an OLS regression of item difficulty using the generated hypotheses as features on the test set, and report adjusted $R^2$ as a goodness-of-fit measure of their explanatory power (i.e., the in-sample setting in \cref{sec:metrics}).
The results are present in Figure \ref{fig:in_dist_r2}, where our method achieves the best fit at most hypothesis budgets on three datasets. 
More importantly, the baselines saturate quickly: they yield only 5--10 effective hypotheses, beyond which adjusted $R^2$ declines, indicating that the additional hypotheses carry noise rather than real signal. 
Our method, in contrast, keeps producing informative hypotheses. 
We attribute this to the fact that the baselines generate patterns directly from the data and therefore discover only the most salient signals, whereas our sampling strategy allows the model to explore new difficulty factors by repeatedly contrasting difficult and easy questions.

\paragraph{The generated hypotheses predict the difficulty on par with black-box predictors.}
We now examine how well $\mathcal{H}$ predicts the difficulty of unseen questions. 
We train an OLS regressor on $\mathcal{G}$ and $\mathcal{V}$ using the hypotheses as features and evaluate its $R^2$ on the test set, comparing our method against the baseline difficulty predictors introduced in \cref{sec:baselines}. 
The results are presented in Table~\ref{tab:regression}. 
Overall, our model achieves the best performance on two datasets, GSM8K and WinoGrande, and the second best on BBH-structured, showing that these features alone already constitute a strong difficulty predictor. 
We note that all models perform poorly on WinoGrande. 
This again echoes the finding that the difficulty of WinoGrande is hard to predict \citep{ding2024easy2hard}\footnote{On Winogrande, human annotators agree with the IRT-estimated difficulty on only $73.4\%$ of item pairs. Under a bivariate-normal assumption, this agreement level implies $R^2 \approx 0.45$ as a ceiling.}.
Nevertheless, our method still extracts some signal on this dataset, while all baselines except \textsc{Claude-Sonnet-5} fail completely, attaining $R^2$ values around or even below $0$.

\begin{table}[h]
\caption{Performance of different difficulty prediction methods on three datasets. 
All methods are trained on $\mathcal{G} \cup \mathcal{V}$, and we report $\mathrm{R^2}$ and $\mathrm{RMSE}$ on $\mathcal{T}$. Results of \textsc{RoBerta-base} are averaged over 5 seeds, and prompting of LLMs over 3 set of demonstrations. The number of hypotheses $|\mathcal{H}|$ for the baselines is selected on the verification pool over a sweep of $\{10, 20, 50\}$; for \textsc{Edisa}, we filtered out clusters with $\leq5$ members. Subscripts denote standard deviations across seeds. Lasso and OLS are deterministic given the data, so no $\mathrm{s.d.}$ is reported for them. Best results are in \textbf{bold} and second-best are \underline{underlined}.}
\label{tab:regression}
\small
\centering
\setlength{\tabcolsep}{5pt}
\begin{tabular}{lcccccc}
\specialrule{1.0pt}{0pt}{2pt}
\multirow{2}{*}{\textbf{Approaches}} & \multicolumn{2}{c}{\textbf{GSM8K}}   & \multicolumn{2}{c}{\textbf{BBH-structured}}     & \multicolumn{2}{c}{\textbf{Winogrande}}     
\\ \cmidrule(l){2-3} \cmidrule(l){4-5} \cmidrule(l){6-7}
& $\mathrm{R^2} \uparrow$ & $\mathrm{RMSE} \downarrow$ & $\mathrm{R^2} \uparrow$ & $\mathrm{RMSE} \downarrow$ & $\mathrm{R^2} \uparrow$ & $\mathrm{RMSE} \downarrow$ \\ 
\midrule
\textsc{RoBerta-base}\textsubscript{finetuned}           &   \num{0.362} \textsubscript{0.02}  &   1.454 \textsubscript{0.02}  & \textbf{0.646} \textsubscript{0.01}  &  \textbf{0.698} \textsubscript{0.01} &  -0.038 \textsubscript{\phantom{0.09}} & 1.711 \textsubscript{\phantom{0.07}}     \\ 
\textsc{Gemma-3-4B}\textsubscript{last-token}    &   0.191 \textsubscript{\phantom{0.00}}  & 1.637 \textsubscript{\phantom{0.00}}     & 0.554 \textsubscript{\phantom{0.00}} & 0.783 \textsubscript{\phantom{0.00}}  & -0.019 \textsubscript{\phantom{0.00}}    & 1.697 \textsubscript{\phantom{0.00}}   \\
\textsc{Qwen3-Embedding-4B}          &      0.200 \textsubscript{\phantom{0.00}}     &    1.629 \textsubscript{\phantom{0.00}}    &    0.533 \textsubscript{\phantom{0.00}}   & 0.802 \textsubscript{\phantom{0.00}}     &  -0.000 \textsubscript{\phantom{0.00}}   &    1.681 \textsubscript{\phantom{0.00}}      \\ \midrule
\textsc{Gemini-3.1-pro}\textsubscript{10-shot}        &  0.346 \textsubscript{0.03}     & 1.472 \textsubscript{0.04}      & 0.139 \textsubscript{0.172}     & 1.085 \textsubscript{0.107}         &   -0.134 \textsubscript{0.14}      & 1.788 \textsubscript{0.11}               \\
\textsc{Claude-sonnet-5}\textsubscript{10-shot}       &    0.041 \textsubscript{0.06}   & 1.782 \textsubscript{0.05}    &   0.043 \textsubscript{0.284}    & 1.139 \textsubscript{0.165}        &  \underline{0.080} \textsubscript{0.03}      &  \underline{1.618} \textsubscript{0.03}      \\ \midrule
\textsc{HypotheSAE} \textsubscript{$|\mathcal{H}|$=\{50, 50, 50\}}    &  0.351 \textsubscript{\phantom{0.00}}   & 1.467 \textsubscript{\phantom{0.00}}   &      0.477 \textsubscript{\phantom{0.00}}    & 0.848 \textsubscript{\phantom{0.00}}      & -0.160 \textsubscript{\phantom{0.00}}    & 1.810 \textsubscript{\phantom{0.00}}              \\
\textsc{HypoGeniC} \textsubscript{$|\mathcal{H}|=\{20, 20, 20\}$}  &    0.327 \textsubscript{\phantom{0.00}}     &  1.494 \textsubscript{\phantom{0.00}}    &  0.465 \textsubscript{\phantom{0.00}}        &  0.861 \textsubscript{\phantom{0.00}}    &    -0.004 \textsubscript{\phantom{0.00}}    & 1.684 \textsubscript{\phantom{0.00}}             \\
\textsc{NLParam}$_{|\mathcal{H}|=\{20, 20, 20\}}$    & 0.240 \textsubscript{\phantom{0.00}}    &   1.587 \textsubscript{\phantom{0.00}}    &   0.088 \textsubscript{\phantom{0.00}}  & 1.120 \textsubscript{\phantom{0.00}}     &  0.015 \textsubscript{\phantom{0.00}}    & 1.668 \textsubscript{\phantom{0.00}} \\
\textsc{Edisa}$_{|\mathcal{H}|=\{44, 47, 37\}}$    & \underline{0.368} \textsubscript{\phantom{0.00}}    &   \underline{1.448} \textsubscript{\phantom{0.00}}    &   0.485 \textsubscript{\phantom{0.00}}  & 0.842 \textsubscript{\phantom{0.00}}     &  -0.014 \textsubscript{\phantom{0.00}}    & 1.693 \textsubscript{\phantom{0.00}}   
\\ \midrule
Ours \textsubscript{$|\mathcal{H}|$=\{40, 12, 44\}}      & \textbf{0.373} 
\textsubscript{\phantom{0.00}}        & \textbf{1.441} \textsubscript{\phantom{0.00}}        & \underline{0.580} \textsubscript{\phantom{0.00}}     &   \underline{0.760} \textsubscript{\phantom{0.00}}       &     \textbf{0.090} \textsubscript{\phantom{0.00}}        &  \textbf{1.603} \textsubscript{\phantom{0.00}}  \\            
\specialrule{1.0pt}{2pt}{0pt}
\end{tabular}

\end{table}

\begin{figure}[h]
\centering
\includegraphics[width=\linewidth]{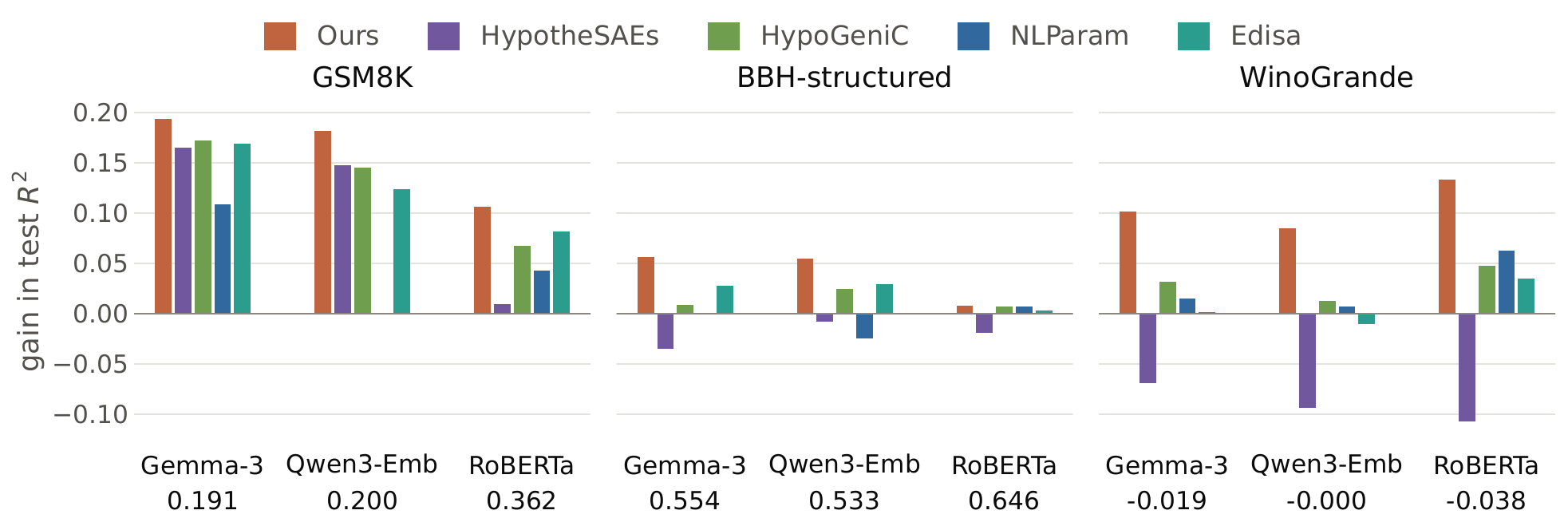}
\caption{Comparison of $R^2$ gains from incorporating hypotheses generated by different methods into different regression models (Base-model performance is shown below).}
\label{fig:augment}
\end{figure}
\paragraph{The hypotheses capture difficulty signals that baselines fail to.}
So far we have compared our hypothesis-based difficulty predictor against strong baselines. 
In practice, however, our method is complementary to them rather than competing with them: the hypotheses it discovers can be supplied to existing predictors as additional features. 
We test this on three semantic predictors--fine-tuned \textsc{RoBERTa-base}, \textsc{Gemma-3-4B} and \textsc{Qwen3-Embedding-4B}---by adding each method's hypothesis match vector as a second feature block.
Figure~\ref{fig:augment} reports the gain in test $R^2$. 
Our hypotheses give the largest gain of any hypothesis source in all settings.
The gains are largest on GSM8K: $+0.19$ and $+0.18$ over the two frozen embeddings, and $+0.11$ even over the fine-tuned RoBERTa ($0.362 \rightarrow 0.468$), by far the strongest baseline there. 
On BBH-structured the gain over the strongest model is small: RoBERTa-base already reaches $R^2 = 0.646$ there, and our hypotheses add only $+0.008$ on top of it (they still add $+0.05$ to the two frozen embeddings), implying that on this dataset the difficulty signal is largely already captured by the encoder itself.
On WinoGrande, all baselines sit at around $0$, yet our hypotheses lift each of them to $R^2$ between $0.08$ and $0.10$.

\subsection{Impact of Sampling Strategy}

\begin{figure}[h]
    \centering
    \begin{minipage}[t]{0.48\textwidth}
        \vspace{0pt}
        \centering
        \includegraphics[width=\linewidth]{
            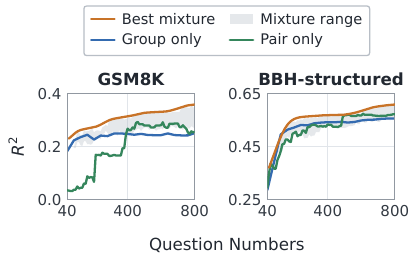
        }
        \caption{
            Group and pair sampling on GSM8K and BBH-structured.
            Shading shows the range across mixed allocations;
            orange marks the best mixture at each question budget.
        }
        \label{fig:sampling}
    \end{minipage}\hfill%
    \begin{minipage}[t]{0.48\textwidth}
        \vspace{0pt}
        \centering
        \includegraphics[width=\linewidth]{
            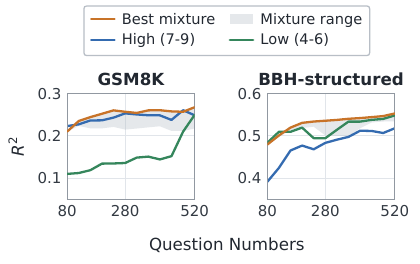
        }
        \caption{
        Contrast-strength ablation on GSM8K and BBH-structured.
        High and low use bucket gaps of 7--9 and 4--6.
        Shading shows the mixture range; orange marks its upper bound.
        }
        \label{fig:contrast-ablation}
    \end{minipage}
\end{figure}

\paragraph{Combining group and pair contrasts outperforms either alone.}
Figure \ref{fig:sampling} compares three prompting strategies: group-only, pair-only, and a mixture of group and pair prompts. 
The first two are designed to discover hypotheses at different granularities. 
For the mixture, we keep the total number of questions fixed and vary the proportion of group versus pair prompts for fair comparison, with the range of results across these proportions shown in the shaded region.
On GSM8K, group-only converges faster, although pair-only reaches comparable or better performance with a larger budget. 
On BBH, the two perform similarly.
On both datasets, however, interpolating between them under the same budget yields better results in most cases (shadow area), scales better as the budget grows, and reaches a higher upper bound (orange line). 
This suggests that mixing the two strategies indeed discovers complementary hypotheses.

\paragraph{Larger gaps are not always better.}
Figure~\ref{fig:contrast-ablation} compares the effect of contrast strength: sampling high-contrast questions (bucket gap $ \delta \geq 7$; Eq. \ref{eq:sample_bucket}), sampling low-contrast questions ($ 4\le \delta \le 6$), and mixing the two. 
The former is designed to capture more pronounced difficulty patterns, while the latter more nuanced signals. 
The two strategies behave in opposite ways on the two datasets: high contrast is the stronger choice on GSM8K, while low contrast wins on BBH-structured. 
The mixed strategy serves as a more stable option: it does not beat the better single strategy at every budget, but it stays close to it in most cases. 
We therefore recommend the mixed strategy as the more reliable option.

\paragraph{A small budget suffices for strong performance, with potential for more.} Overall, with a suitable sampling and prompting strategy, our method reaches a good level of performance within $200$ questions (a proxy for the annotation budget) and retains the potential to scale further when the budget allows, as none of the mixed-strategy curves in Figure~\ref{fig:sampling} has fully converged. 
This echoes the results in Figure~\ref{fig:in_dist_r2}, and again indicates the ability of our method to keep discovering diverse and effective hypotheses.
\section{Causal Analysis}


\definecolor{rewritered}{RGB}{170,0,0}
\providecommand{\rewriteedit}[1]{%
  \textcolor{rewritered}{\bfseries #1}%
}

\providecommand{\tabanswer}[1]{%
  \hfill\textit{Ans.}\ #1%
}

\providecommand{\tabanswernewline}[1]{%
  \par\noindent\hfill\textit{Ans.}\ #1%
}


\definecolor{rewritered}{RGB}{170,0,0}

\providecommand{\rewriteedit}[1]{%
  \textcolor{rewritered}{\bfseries #1}%
}

\providecommand{\tabanswer}[1]{%
  \hfill\mbox{\textit{Ans.}\ #1}%
}

\begin{table*}[t]
\caption{
Bidirectional editing for a hypothesis with an explicit
entity-count threshold.
\rewriteedit{Red text} marks additions or replacements;
the decrease rewrite also removes the third dog's stated distance.
The increase edit expands three clothing categories to four,
whereas the decrease edit reduces four dogs to three.
Both preserve the task and answer units.
}
\label{tab:counterfactual-bidirectional-example}
\centering
\footnotesize
\setlength{\tabcolsep}{5pt}
\renewcommand{\arraystretch}{1.08}

\begin{tabularx}{\textwidth}{
    >{\raggedright\arraybackslash\bfseries}p{0.155\textwidth}
    >{\raggedright\arraybackslash}X
}
\toprule

Hypothesis
&
\emph{Requires tracking and calculating values for four or more
distinct entities or categories.}
\\

\midrule

Increase: original
&
Mishka bought 3 pairs of shorts, 3 pairs of pants, and 3 pairs
of shoes. One pair of shorts costs \$16.50. One pair of pants
costs \$22.50 and one pair of shoes costs \$42. How many dollars
did Mishka spend on all the clothing items?
\tabanswer{\$243}
\\[1pt]

Increase: \\ edited
&
Mishka bought 3 pairs of shorts, 3 pairs of pants, 3 pairs of shoes,
\rewriteedit{and 3 shirts}. One pair of shorts costs \$16.50.
One pair of pants costs \$22.50, one pair of shoes costs \$42,
\rewriteedit{and one shirt costs \$15}. How many dollars did
Mishka spend on all the clothing items?
\tabanswer{\$288}
\\

\midrule

Decrease: original
&
Bernie has 4 dogs. They each need a certain amount of exercise
per day. The first needs to walk 1 mile. The second needs to
walk 4 miles. The third needs to walk 3 miles. On average,
they need to walk 3 miles per day. How many miles does the
last dog need?
\tabanswernewline{4 miles}
\\[1pt]

Decrease:  edited
&
Bernie has \rewriteedit{3 dogs}. They each need a certain amount
of exercise per day. The first needs to walk 1 mile. The second
needs to walk 4 miles. On average, they need to walk 3 miles
per day. How many miles does the last dog need?
\tabanswer{4 miles}
\\

\bottomrule
\end{tabularx}
\end{table*}

\paragraph{Counterfactual question editing.}
\label{sec:counterfactual-rewriting}
We test whether a hypothesis captures a difficulty-bearing property through targeted question editing (see examples in  Table~\ref{tab:counterfactual-bidirectional-example}). 
Because the difficulty signal in WinoGrande is inherently weak, we focus here on GSM8K and BBH-structured.
We randomly sample 50 test questions, determine which hypotheses apply to each of them, and prompt an LLM to edit and verify the questions, changing only what the hypothesis describes while leaving the rest intact. Details are given in Appendix \ref{app:question-rewriting-details}.

\paragraph{Evaluation protocol.}
We evaluate the causal effect using the difficulty shift and the success rate defined in \cref{sec:metrics}. 
Measuring the difficulty of an edited question would require re-evaluating it on thousands of LLMs, which is prohibitively costly. 
We therefore re-run only a subset of LLMs, chosen to cover a range of performance levels (from below 10\% to above 80\% accuracy) and limited to models under 30B due to the restriction of our compute resource. 
A reliable difficulty re-estimation is not attainable from this subset, so throughout this section we use accuracy as a proxy for difficulty\footnote{Under the 1PL model an item's mean accuracy over a fixed respondent set is a strictly monotone function of its difficulty, so the direction of a within-panel accuracy change matches the direction of the difficulty shift.}.
Because each hypothesis applies to a different set of questions (those it can meaningfully rewrite), and induces a different kind of edit, comparing hypotheses against one another in a controlled, fair setting is difficult, which in turn makes a comparison against the baselines difficult. 
In this section we therefore focus on validating the effectiveness of our own hypotheses.

\paragraph{Results.}
\label{sec:counterfactual-results}
In Table~\ref{tab:counterfactual-results} we report the difficulty shift and success rate averaged over all hypotheses, which shows consistent bidirectional changes on both datasets. 
Increase-difficulty edits reduce mean accuracy by 15.27 and 23.88 percentage points on GSM8K and BBH-structured, respectively, while decrease-difficulty edits improve it by 32.18 and 28.52 points.
Question-level alignment reaches 70.9\% and 92.2\% for increase and decrease edits on GSM8K, compared with 77.6\% and 85.5\% on BBH-structured. 
Thus, the aggregate changes are accompanied by directional consistency across most edited questions, rather than only a few large shifts.

In addition, we present the full model list and per-model result in Appendix
Table~\ref{tab:counterfactual-routereval}.
The results show that these patterns extend across model families: 14/17 GSM8K models and $15/16$ BBH-structured models lose accuracy after increase edits, whereas $16/17$ and $16/16$ gain accuracy after decrease edits.
The changes are not confined to weak models.
On BBH-structured, \textsc{Yi-1.5-9B-Chat} falls from 88.1\% to 37.3\% after increase edits; on GSM8K, \textsc{Llama-3-8B-Instruct} rises from 43.1\% to 94.1\% after decrease edits.
Conversely, all three models with reversed increase-direction changes start below 15\% accuracy on the corresponding original questions.
On the one hand, their poor performance leaves little room for a further drop. 
On the other, most of them probably lack genuine reasoning ability and perform close to random guessing, so they may be insensitive to changes in difficulty.
Finally, Table \ref{tab:hypothesis-rewrite-transitions} provides the per-hypothesis results, where we can see that most individual hypotheses also achieve a solid difficulty shift and success rate.
Together, these results support a causal effect of the generated hypotheses on difficulty that holds across models.
\begin{table*}[t]
\caption{
Hypothesis-guided rewriting results on both datasets.
{Orig.}\ and {Rewrite} are mean accuracies (\%) across pairs and models;
$\mathrm{\Delta Acc}_{\mathcal{H}}$ and $\mathrm{SR_{\mathcal{H}}}$ are the accuracy shift and success rate (counts in parentheses) averaged over all hypotheses, respectively (Eq. \ref{eq:casual_eval}).
}
\label{tab:counterfactual-results}
\centering
\small
\setlength{\tabcolsep}{3pt}
\renewcommand{\arraystretch}{1.15}

\begin{tabular}{lcccc cccc}
\toprule
&
\multicolumn{4}{c}{\textbf{GSM8K}} &
\multicolumn{4}{c}{\textbf{BBH-structured}} \\
\cmidrule(lr){2-5}\cmidrule(lr){6-9}

\textbf{Intervention} &
{Orig.} &
{Edited} &
${\Delta} \mathrm{Acc_{\mathcal{H}}}$ &
 $\mathrm{SR}_{\mathcal{H}}$ &
{Orig.} &
{Edited} &
${\Delta} \mathrm{Acc_{\mathcal{H}}}$ &
$\mathrm{SR}_{\mathcal{H}}$ \\
\midrule

Increase difficulty
& 45.01 & 29.75
& \cellcolor{deltaredbg}$-15.27$
& 70.9\% (100/141)
& 60.63 & 36.75
& \cellcolor{deltaredbg}$-23.88$
& 77.6\% (52/67) \\

Decrease difficulty
& 25.61 & 57.79
& \cellcolor{deltagreenbg}$+32.18$
& 92.2\% (47/51)
& 36.82 & 65.34
& \cellcolor{deltagreenbg}$+28.52$
& 85.5\% (47/55) \\

\bottomrule
\end{tabular}
\end{table*}

\section{Conclusion}
In this work, we presented a hypothesis generation method for explaining what makes a question difficult. 
Rather than treating difficulty as a number to be estimated, we take it as a quantity to be explained: we measure difficulty with IRT from the responses of thousands of LLMs, sample contrasting sets of easy and hard questions, and prompt an LLM to propose natural-language hypotheses. 
Experiments on three datasets show that the resulting hypotheses are interpretable and of high quality. They capture difficulty signals that semantic encoders do not, predict the difficulty of unseen questions as well as or better than black-box predictors, and further improve those predictors when added as features. Beyond prediction, they also guide the editing of questions to manipulate their difficulty in a controlled direction, which suggests that they identify genuine sources of difficulty rather than post-hoc correlations. We hope this offers a step toward benchmarks whose difficulty can be understood and designed, rather than only measured.

\clearpage
\bibliography{iclr2027_conference}
\bibliographystyle{iclr2027_conference}

\clearpage

\appendix

\section{Related Work}\label{related_work}

\textbf{Difficulty Modeling.}
Question difficulty is a central quantity wherever questions are used to measure ability. 
In educational testing it determines which items enter an exam, how scores are compared across test forms, and which item an adaptive test presents next \citep{van2000computerized}; it also has to be known for newly written questions before they can be administered, which has motivated predicting and controlling the difficulty of generated items \citep{cui-sachan-2023-adaptive,jiao2023automatic}. 
In LLM evaluation it has become equally consequential: benchmarks are filtered or rebuilt around the items that remain hard \citep{kipnis2025metabench,zouhar-etal-2025-select}, and item-level difficulty is used to pick informative subsets, to test models adaptively at a fraction of the cost \citep{hofmann2025fluid,polo2024tinybenchmarks}, or to weight items when ranking models by competency \citep{hu2026eip}.
How difficulty is obtained falls into two families. 
They first estimates it from responses: descriptive IRT models treat each item's difficulty as a free parameter, while explanatory item response models constrain it to be a function of item covariates, classically a linear combination of the cognitive operations an item requires \citep{fischer1973linear, janssen2004models}. 
These covariates are specified by experts, so the explanation is only as complete as the expert's list; our approach can be seen as an explanatory model whose covariates are discovered from data.
The second family predicts difficulty from the item text alone. 
Early work relied on hand-designed linguistic and cognitive features \citep{benedetto2023survey, alkhuzaey2024text}, which were then replaced by neural encoders of the item \citep{question2017huang, loginova2021towards}, the approach that dominated the recent shared task on medical exam items \citep{yaneva-etal-2024-findings}; \citet{byrd-srivastava-2022-predicting} extend this line beyond difficulty to discrimination, relating both parameters to textual and contextual properties of the question. 
With the arrival of LLMs, difficulty has been estimated by prompting them to judge items against IRT-derived levels, by scoring the plausibility of the candidate answers \citep{mozafari-etal-2026-question}, and by reading it off the hidden states of the answering model itself \citep{lugoloobi2025llms}. 
Closest to our goal, \citet{atlanta2024assessing} use LLMs to analyze which features are associated with question difficulty; there the features are drawn from a predefined inventory and examined post hoc, whereas we let the model propose candidate explanations, validate each of them on held-out items, and select among them by their joint predictive power.

\textbf{Automated Error Analysis.}
A related area, slice discovery, seeks to discover the data on which an AI model is prone to err and to explain the patterns behind those errors, which is in essence also a form of explanation of difficulty. 
Early work assumed structured inputs and searched over conjunctions of feature values for slices on which a model performs significantly worse than average, with the slice definition itself serving as the explanation \citep{chung2019slice, sagadeeva2021sliceline}; 
Later methods instead look for coherent, underperforming regions of a model's representation space and attach natural-language descriptions to the regions they find, by matching them to captions in a shared embedding space or by prompting an LLM with their members \citep{spotlight2022, eyuboglu2022domino, hua2023discover}. 
What these methods explain, however, is the difficulty an item poses to one particular model. 
The errors of a single model, however, are likely to confound the difficulty of an item with the weaknesses and biases of that particular model, rather than reflecting an intrinsic property of the item. 
While effective for model debugging, the weaknesses these methods uncover are therefore unlikely to transfer to other models, and offer limited insight into the broader questions of how a dataset should be curated and iterated on.

\textbf{Hypothesis Generation with LLMs.}
A growing line of work uses LLMs to describe, in natural language, the patterns shared by a group of data. 
These methods first partition the data into groups that are likely to share a pattern or behavior, for example, by taking samples with different labels or from different corpora \citep{pmlr-v162-zhong22a,zhong2023goal}, or by clustering or otherwise exploiting structure in embedding space \citep{zhong2024explaining, pmlr-v267-movva25a}. 
They then prompt an LLM with samples from these groups to generate descriptions of what the samples have in common, typically followed by some form of validation of the generated descriptions.
Across previous approaches, the property being explained is usually one that is observable in the text or supplied as a label: a topic, stance, or writing style, the aspects that drive a product rating, membership of a given group, or an annotated outcome such as engagement or persuasiveness \citep{pmlr-v162-zhong22a}. Our work keeps the propose-and-verify structure of this line but changes the target and the evaluation: the property is a latent, continuous quantity inferred from model responses rather than annotated, the contrasting groups are constructed by sampling from that quantity, and hypotheses are scored by how much of its variance they explain, individually and jointly, on items never used for their discovery.
\section{Prompts}

\begin{promptbox}[label={prompt:group_proposer}]{Group Proposer Prompt}

\textbf{System:} 
\par\smallskip
You study why some questions in a benchmark are harder for language
models to answer correctly than others. ``Harder'' is measured, not
guessed: every question has an IRT (Item Response Theory) difficulty
parameter fit from many models' responses. Each question below is
annotated with that parameter: larger values mean harder questions and
smaller values mean easier questions. The values are positions on this
benchmark's relative IRT scale, not probabilities or percentages.

\{Dataset-specific context\}
You will see two GROUPS of questions from the same benchmark: a HARDER
group, whose questions all have much higher measured difficulty, and an
EASIER group. Your task is to propose hypotheses about WHAT MAKES A
QUESTION HARD in this benchmark: short, testable natural-language
descriptions of a property that the harder questions tend to have and
the easier ones tend not to, or the reverse.

How to work:

\begin{itemize}
    \item First read BOTH groups in full, before writing anything. Look
    for patterns that recur across several questions in the HARDER group
    and are absent or rarer in the EASIER group, rather than what is
    special about any one question.

    \item Use the individual IRT values as graded evidence. Prefer a
    pattern when questions exhibiting it tend to have larger values
    across the examples, and do not let one extreme-valued outlier stand
    in for a recurring pattern.

    \item Write one hypothesis for each pattern actually found. There is
    no target count. Do not pad the list.
\end{itemize}

Guidelines:

\begin{enumerate}
    \item Stay on difficulty. Each hypothesis must describe something
    that plausibly makes a question harder or easier to answer correctly.
    Avoid topic labels and properties specific to one question.

    \item Consider, but are not limited to:
    \{Dataset-specific aspects\}.

    \item Generalize. A hypothesis should be a property that many unseen
    questions in the benchmark could exhibit and that can be checked
    without knowing the question's difficulty.

    \item Prefer breadth. A pattern shared by several HARDER questions is
    preferable to one that fits only a single question.

    \item Propose distinct hypotheses covering different plausible
    explanations, not rewordings of one another.

    \item State each hypothesis as a specific, checkable yes/no criterion,
    rather than a vague label such as ``is conceptually difficult'' or
    ``requires careful reasoning.''

    \item Keep each hypothesis concise and atomic. If two properties can
    be judged separately, state them as separate hypotheses rather than
    joining them with ``and'' or ``or.''
\end{enumerate}

Respond with ONLY a JSON object:

\texttt{\{"hypotheses": ["...", "...", ...]\}}

\par\medskip
\noindent\textbf{User:}\par\smallskip

HARDER group (much higher measured difficulty):

\{Numbered harder questions with IRT difficulties and
reference answers/explanations\}

EASIER group (much lower measured difficulty):

\{Numbered easier questions with IRT difficulties and
reference answers/explanations\}

Read both groups first. Then propose the NEW hypotheses you find, one
per pattern actually observed---as many or as few as the groups
genuinely support, with no fixed count.

\end{promptbox}

\begin{promptbox}[label={prompt:pair_proposer}]{Pair-Moderate Proposer Prompt}

\textbf{System:} 
\par\smallskip

You study why some benchmark questions are harder for language models
than others. Difficulty is an IRT parameter estimated from many models'
responses; larger values mean harder questions.

\{Dataset-specific context\}

You will see exactly 5 randomly sampled pairs. Within each pair, HARDER
has an IRT difficulty at least \{minimum gap\} logits above EASIER.
Pairs were not matched by topic and were not drawn from difficulty
buckets.

Read all pairs before answering. Propose short, testable hypotheses
about properties that may make questions harder or easier. For each
hypothesis, list every pair number that supports it. A hypothesis may
be supported by a single pair, but it must describe a reusable property
that could naturally occur in unseen questions.

IMPORTANT: Include both broad recurring factors and narrower reusable
subtypes when the pairs support them; a subtype does not need to recur
across many shown pairs. Prefer concrete yes/no criteria, but do not
copy names, objects, answers, incidental constants, distinctive wording,
or a complete solution path from a shown question. Do not create a long
conjunction whose main purpose is to identify one displayed item.
Numerical thresholds are allowed only when they define a reusable
complexity class.

Consider, but are not limited to: \{Dataset-specific aspects\}. Return
as many well-supported, non-redundant hypotheses as the evidence
justifies, up to 10. Do not pad the list.

Return ONLY valid JSON:

\texttt{\{"hypotheses": [\{"hypothesis": "...", "pairs": [1, 4]\},}\\
\texttt{\phantom{\{"hypotheses": [}\{"hypothesis": "...", "pairs": [2]\}]\}}

\par\medskip
\noindent\textbf{User:}\par\smallskip

5 random hard/easy pairs (each IRT gap is at least \{minimum gap\}):

Pair 1

\quad HARDER [IRT difficulty: \{value\}]:
\{Question with reference answer/explanation\}

\quad EASIER [IRT difficulty: \{value\}]:
\{Question with reference answer/explanation\}

\(\vdots\)

Pair 5

\quad HARDER [IRT difficulty: \{value\}]: \{Question\}

\quad EASIER [IRT difficulty: \{value\}]: \{Question\}

Read all pairs first, then return only the required JSON object.

\end{promptbox}

\begin{promptbox}[label={prompt:match_judge}]{Match Judge Prompt}

\textbf{System:} 
\par\smallskip

You judge whether a benchmark question exhibits each of several
candidate properties.

Candidate properties:

1. \{Hypothesis 1\}

2. \{Hypothesis 2\}

\(\vdots\)

\(K\). \{Hypothesis \(K\)\}

For the question given by the user, answer whether it exhibits each
candidate property above. When uncertain, answer 0 (do not guess yes).

Respond with ONLY a JSON object:

\texttt{\{"scores": [0 or 1, 0 or 1, ...]\}}

Return exactly one integer per candidate property, in the same order
they were listed.

\par\medskip
\noindent\textbf{User:}\par\smallskip

Question and reference answer/explanation:

\{Item text\}

Scores:

\end{promptbox}

\label{app:prompt}

\section{Details of Experiment Setup}\label{app:implementation}

\subsection{Dataset statistics}
\label{app:dataset-statistics}
Table~\ref{tab:dataset-statistics} summarizes the dataset splits,
the number of models used for IRT estimation, and the resulting
item-difficulty distributions.


\begin{table*}[t]
\caption{
Dataset statistics.
\#Models denotes the number of models whose responses are used
to estimate IRT difficulty.
$\mathcal{G}$, $\mathcal{V}$, and $\mathcal{T}$ denote our
hypothesis-generation, validation, and test sets, respectively.
Difficulty statistics summarize unstandardized 1PL IRT difficulty $d$
over $\mathcal{G}\cup\mathcal{V}\cup\mathcal{T}$.
SD denotes the sample standard deviation across items.
}
\label{tab:dataset-statistics}
\centering
\small
\setlength{\tabcolsep}{7pt}
\renewcommand{\arraystretch}{1.12}

\begin{tabular}{lccccccc}
\toprule
& &
\multicolumn{3}{c}{\textbf{Split}} &
\multicolumn{3}{c}{\textbf{IRT difficulty} $d$} \\
\cmidrule(lr){3-5}\cmidrule(lr){6-8}
\textbf{Dataset} &
\textbf{\#Models} &
$|\mathcal{G}|$ &
$|\mathcal{V}|$ &
$|\mathcal{T}|$ &
\textbf{Min} &
\textbf{Mean $\pm$ SD} &
\textbf{Max} \\
\midrule
GSM8K
& 5,000 & 719 & 400 & 200
& $-3.40$ & $1.13 \pm 1.85$ & $7.29$ \\

BBH-structured
& 3,811 & 558 & 558 & 280
& $-2.83$ & $0.14 \pm 1.19$ & $4.86$ \\

Winogrande
& 5,000 & 759 & 254 & 254
& $-3.92$ & $-0.70 \pm 1.67$ & $5.09$ \\
\bottomrule
\end{tabular}
\end{table*}

\subsection{Hypothesis Discovery}
\label{app:impl-hypgen}
All generation, feature selection and model fitting take place inside the generation and verification pools; the test pool is used once, for the reported metrics. Two models recur throughout: \textsc{Gemini-3.1-pro-preview} proposes hypotheses, and \textsc{Gemini-3.7-flash} labels items against them. Both run at temperature $0$ with JSON-constrained output and the model's lowest thinking setting.
\paragraph{Our method.}
For each dataset we run 30 rounds of group-based generation and 30 rounds of pair-based generation. Each group sample contains 40 questions, 20 hard and 20 easy. For group-based sampling, we partition $\mathcal{G}$ into 10 difficulty-rank buckets and set the minimum bucket gap to $\delta=3$ and the gap-weighting exponent to $\rho=2$. The reuse-penalty exponents are $\gamma_c=1$ for bucket combinations and $\gamma_i=1$ for individual questions. Each pair prompt presents 10 questions in 5 disjoint hard/easy pairs. For pair-based sampling, we set the minimum IRT difficulty gap to 4.0 on GSM8K and Winogrande and 3.5 on BBH-structured, with $\gamma_i=2$ for individual-question reuse and $\gamma_p=1$ for exact-pair reuse. Questions are sampled without replacement within each round; reuse counts are accumulated across rounds. Figure~\ref{fig:sampling} shows how performance changes with the question budget. The candidate pool is reduced by stability selection with a Lasso base learner: we draw 100 subsamples of 75\% the training items, record how often each candidate receives a non-zero coefficient, and keep the most frequently selected ones. The group and pair candidate hypothesis pools contain a combined total of \num{376}, \num{235}, and \num{171} hypotheses on GSM8K, BBH-structured and Winogrande. After deduplication and refinement, the corresponding pools contain \num{233}, \num{127}, and \num{146} hypotheses.

\paragraph{\textsc{HypoGeniC}.}
The method presumes a classification target, so we discretize difficulty into three equal-frequency levels and map a predicted level back to that level's mean difficulty when regression metrics are needed. 
Apart from these adaptations, we follow the original implementations as closely as possible.
10 initial items seed ten initial hypotheses. 
A bank of 20 hypotheses is then maintained by an upper-confidence-bound rule with exploration weight $0.5$, the 10 highest-scoring hypotheses vote on each new item, and a pool of ten mispredicted items triggers a generation of five new hypotheses whose rewards are initialized on the most recently scored items. 
We also ran a stricter variant in which generation only ever sees the generation pool $\mathcal{G}$ and rewards only ever come from the validation pool $\mathcal{T}$.
We stop generation once the model has seen a fixed number of questions, matching the number seen by our method, as a proxy for the budget to ensure a fair comparison.

\paragraph{\textsc{NLParam}.}
We follow the reference implementation closely. 
Questions are embedded with \textsc{Instructor-XL} \citep{su-etal-2023-one} exactly as that code embeds them.
Predicate directions are initialized by $k$-means\texttt{++} and refined by 10 alternations between the regression weights and the relaxed predicate representation, each solved with Adam \citep{kingma2015iclr-adam} at learning rate $10^{-2}$ for at most a thousand steps. 
Refinement then repeats 10 times, each time ablating the least useful predicate, refitting that direction and replacing it with the best of five candidates under the full-data loss. 
We report the results of 20 predicates.

\paragraph{\textsc{HypotheSAEs}.}
We use the Gemma-Scope-2\footnote{\url{https://huggingface.co/google/gemma-scope-2}} autoencoders for \textsc{google/gemma-3-4b-pt} at the final (33) layer, width 16 thousand, running them over every token of an item and max-pooling per item; features active on fewer than 20 training questions are discarded. 
Following the original, a Lasso with cross-validated penalty selects features, the non-zero ones are ranked by coefficient magnitude.

\paragraph{\textsc{Edisa}.}
We follow the implementation in \citet{hua2023discover} and replace the error distance with the standardized IRT difficulty; the model-confidence term is dropped, as it has no counterpart for a population of models.
Questions are embedded together with their reference answers by \textsc{Qwen3-Embedding-4B} and reduced to 128 dimensions with PCA fitted on $\mathcal{G}\cup\mathcal{V}$. 
A mixture of $k$ diagonal Gaussians over the embedding and the difficulty is fitted by EM, initialized by $k$-means, with the two likelihood terms weighted by exponents $\gamma$ and $\lambda$ as in the original; at inference difficulty is marginalized out. 
We set ($\gamma, \lambda$) to $(0.15, 1)$ for GSM8K and BBH, and $(0.15, 100)$ for WinoGrande, selected by 5-fold cross-validation.
For each remaining slice, \textsc{Gemini-3.1-Pro} is shown the 5 members with the highest posterior probability together with their difficulties and the slice's mean difficulty, and writes one predicate that most of them share; no contrast items are shown.

\subsection{Fine-tuned Difficulty Predictors}
\label{app:impl-repr}

\paragraph{Fine-tuned encoder.}
We fine-tune \textsc{roberta-base} end-to-end with a scalar regression head on the item text (question and answer), optimizing squared error against the standardized difficulty and mapping predictions back to the logit scale before scoring. 
We use AdamW \citep{loshchilov2018decoupled} at learning rate $2\times10^{-5}$ with weight decay $0.01$, $10\%$ linear warm-up, batch size $16$ and a maximum sequence length of $512$ tokens, for at most 10 epochs with early stopping on a per-seed tenth of the training pool held out for that purpose, with patience three. 
Every reported number is the mean over 5 seeds, since end-to-end fine-tuning on roughly a thousand items is seed-sensitive. 
Training runs on a single NVIDIA RTX 4090 GPU and takes a few minutes per seed.


\paragraph{Frozen representations.}
\textsc{Gemma-3-4B} is run in \texttt{bfloat16} with a maximum length of $2,176$ tokens and tail truncation, and we take the last-token hidden state of the final layer.
\textsc{Qwen3-Embedding-4B} is used through its sentence-embedding interface
with $\ell_2$-normalised outputs and no instruction prefix.
For both models, features are standardized using training-pool statistics, a Lasso penalty is chosen by five-fold cross-validation inside the training pool, and the resulting model is evaluated once on the test pool. 

\subsection{Prompted Judges}
\label{app:impl-judge}

The prompted baselines receive no training set. 
A context of 10 labelled demonstrations is drawn once from the training pool by splitting it into five equal-frequency difficulty buckets and sampling 2 questions from each, and the same context precedes every test item. 
The system message explains that difficulty is an IRT parameter on a logit scale, gives the benchmark's observed range, mean and standard deviation, and lists the demonstrations with their measured values;
the model replies with a single number, which we use unchanged. 
\textsc{Gemini-3.1-pro-preview} runs at temperature $0$ with its lowest thinking setting and JSON-constrained output. 
The \textsc{Claude-Sonnet-5} API rejects an explicit temperature for this model, so the request omits it and the model decodes at its default setting, with extended thinking disabled; the shared system block is marked for prompt caching. 
Both judges see item text truncated to $1,500$ characters and are run with three demonstration draws, seeds 0 through 2, over which we report mean and standard
deviation. 
Because the draw depends only on the seed and the pools, both judges see identical contexts. 
To ensure reproducibility, we provide the identifiers of the sampled questions in Table~\ref{tab:app-demos}.

\begin{table}[t]
\centering
\small
\caption{Demonstrations used by the prompted judges, listed in increasing order of measured
difficulty.}
\label{tab:app-demos}
\begin{tabular}{lcp{0.58\columnwidth}}
\toprule
Dataset & Seed & Question IDs \\
\midrule
\multirow{3}{*}{GSM8K}
 & 0 & 3344, 3268, 2478, 3542, 3198, 2561, 2761, 3407, 3132, 2894 \\
 & 1 & 2823, 2876, 3727, 3336, 3078, 3406, 2555, 2866, 2806, 3139 \\
 & 2 & 2946, 2602, 3689, 2950, 3273, 3149, 3272, 3306, 3005, 2880 \\
\midrule
\multirow{3}{*}{BBH-structured}
 & 0 & 36661, 34883, 36010, 36121, 34851, 36773, 35797, 36779, 35888, 35503 \\
 & 1 & 36235, 36068, 35911, 37867, 36206, 36670, 37800, 37794, 35495, 34780 \\
 & 2 & 36664, 37919, 36648, 35527, 35981, 35567, 34958, 36780, 35475, 34813 \\
\midrule
\multirow{3}{*}{WinoGrande}
 & 0 & 860, 422, 1006, 894, 139, 1146, 1225, 10, 581, 273 \\
 & 1 & 1139, 1054, 918, 744, 100, 1031, 1019, 699, 1124, 135 \\
 & 2 & 338, 1083, 881, 499, 292, 255, 215, 1239, 80, 1100 \\
\bottomrule
\end{tabular}
\end{table}

\subsection{Question Editing}
\label{app:question-rewriting-details}

We construct hypothesis-guided question edits in three stages.
First, we determine whether the original question
satisfies the target hypothesis using its binary match label.
If the property is present, we request its removal or simplification.
If it is absent, an applicability judge assesses whether it can be
introduced naturally while preserving the question's broad setting
and answer format; inapplicable interventions are skipped. Second, a rewriter produces a separate candidate for each applicable
intervention (including questions and reference answers), targeting the specified property while preserving the
question's broad setting and answer format.
Third, \textsc{Gemini} and \textsc{Astra} verify the candidates in sequence, each
solving the rewritten question from scratch and checking its
solvability, intended property change, and preservation of the
original setting. Malformed or incomplete responses are retried, and
only rewrites accepted by both verifiers enter the evaluation.

\clearpage
\section{More Results}
\begin{table*}[h]
\caption{
Counterfactual evaluation of hypothesis-guided question edits on GSM8K and
BBH-structured across open-weight language models. Values are accuracies
(\%); $\Delta {\rm Acc.}$ is rewritten-item accuracy minus original-item accuracy in
percentage points. For increase-difficulty edits, a negative $\Delta {\rm Acc.}$ is
the intended outcome; for decrease-difficulty edits, a positive $\Delta {\rm Acc.}$
is intended. GSM8K contains 141 increase and 51 decrease edits;
BBH-structured contains 67 increase and 55 decrease edits.
}
\label{tab:counterfactual-routereval}
\centering
\scriptsize
\renewcommand{\arraystretch}{1.10}

\begin{tabular}{llccc|ccc}
\toprule
& &
\multicolumn{3}{c|}{\textbf{Increase-difficulty edits}} &
\multicolumn{3}{c}{\textbf{Decrease-difficulty edits}} \\
\cmidrule(lr){3-5}\cmidrule(l){6-8}
\textbf{\#} &
\textbf{Model} &
\textbf{Orig.} & \textbf{Edited} & $\boldsymbol{\Delta} \mathbf{Acc.}$ &
\textbf{Orig.} & \textbf{Edited} & $\boldsymbol{\Delta} \mathbf{Acc.}$ \\
\midrule

\multicolumn{8}{c}{\textbf{GSM8K}} \\
\midrule
1  & \textsc{TinyLlama-1.1B} \citep{zhang2024tinyllama}                  & 1.4  & 1.4  & \zerodelta{0.0} & 0.0  & 0.0  & \zerodelta{0.0} \\
2  & \textsc{Gemma-2B-IT} \citep{team2024gemma}                     & 4.3  & 6.4  & \posdelta{2.1}  & 2.0  & 25.5 & \posdelta{23.5} \\
3  & \textsc{Qwen1.5-0.5B-Chat} \citep{bai2023qwen}               & 10.6 & 2.8  & \negdelta{7.8}  & 0.0  & 13.7 & \posdelta{13.7} \\
4  & \textsc{Phi-1.5} \citep{li2023textbooks}                         & 13.5 & 19.9 & \posdelta{6.4}  & 3.9  & 43.1 & \posdelta{39.2} \\
5  & \textsc{Qwen1.5-1.8B-Chat} \citep{bai2023qwen}               & 22.0 & 2.1  & \negdelta{19.9} & 7.8  & 9.8  & \posdelta{2.0} \\
6  & \textsc{Mistral-7B-Instruct-v0.1} \citep{jiang2023mistral7b}        & 22.7 & 17.0 & \negdelta{5.7}  & 11.8 & 60.8 & \posdelta{49.0} \\
7  & \textsc{Mistral-7B-Instruct-v0.2} \citep{jiang2023mistral7b}        & 39.7 & 19.9 & \negdelta{19.9} & 7.8  & 60.8 & \posdelta{52.9} \\
8  & \textsc{Qwen1.5-MoE-A2.7B-Chat} \citep{bai2023qwen}          & 41.8 & 31.2 & \negdelta{10.6} & 29.4 & 66.7 & \posdelta{37.3} \\
9  & \textsc{Gemma-1.1-7B-IT} \citep{team2024gemma}                 & 47.5 & 38.3 & \negdelta{9.2}  & 19.6 & 74.5 & \posdelta{54.9} \\
10 & \textsc{Qwen2-0.5B} \citep{yang2024qwen2technicalreport}                      & 51.1 & 19.1 & \negdelta{31.9} & 25.5 & 64.7 & \posdelta{39.2} \\
11 & \textsc{Yi-9B} \citep{young2024yi}                           & 61.7 & 34.8 & \negdelta{27.0} & 31.4 & 82.4 & \posdelta{51.0} \\
12 & \textsc{Qwen1.5-4B} \citep{bai2023qwen}                      & 61.7 & 34.0 & \negdelta{27.7} & 39.2 & 60.8 & \posdelta{21.6} \\
13 & \textsc{Llama-3-8B-Instruct} \citep{grattafiori2024llama}             & 73.0 & 57.4 & \negdelta{15.6} & 43.1 & 94.1 & \posdelta{51.0} \\
14 & \textsc{Qwen2-1.5B} \citep{yang2024qwen2technicalreport}                      & 73.8 & 41.1 & \negdelta{32.6} & 43.1 & 74.5 & \posdelta{31.4} \\
15 & \textsc{Yi-1.5-9B} \citep{young2024yi}                       & 77.3 & 51.1 & \negdelta{26.2} & 60.8 & 80.4 & \posdelta{19.6} \\
16 & \textsc{Yi-1.5-9B-Chat} \citep{young2024yi}                  & 80.9 & 58.9 & \negdelta{22.0} & 56.9 & 82.4 & \posdelta{25.5} \\
17 & \textsc{Phi-3-mini-4k-instruct} \citep{abdin2024phi3technicalreporthighly}          & 82.3 & 70.2 & \negdelta{12.1} & 52.9 & 88.2 & \posdelta{35.3} \\
\midrule
& \textbf{Mean across models}
& --- & --- & \textbf{\negdelta{15.3}}
& --- & --- & \textbf{\posdelta{32.2}} \\
\midrule

\multicolumn{8}{c}{\textbf{BBH-structured}} \\
\midrule
1  & \textsc{Qwen2.5-0.5B-Instruct} \citep{qwen2025qwen25technicalreport}           & 13.4 & 17.9 & \posdelta{4.5}  & 29.1 & 38.2 & \posdelta{9.1} \\
2  & \textsc{TinyLlama-1.1B} \citep{zhang2024tinyllama}                  & 14.9 & 9.0  & \negdelta{6.0}  & 10.9 & 27.3 & \posdelta{16.4} \\
3  & \textsc{Gemma-2B-IT} \citep{team2024gemma}                     & 38.8 & 17.9 & \negdelta{20.9} & 9.1  & 25.5 & \posdelta{16.4} \\
4  & \textsc{Qwen1.5-4B} \citep{bai2023qwen}                      & 43.3 & 16.4 & \negdelta{26.9} & 12.7 & 45.5 & \posdelta{32.7} \\
5  & \textsc{Gemma-2-2B-IT} \citep{team2024gemma}                   & 44.8 & 17.9 & \negdelta{26.9} & 23.6 & 50.9 & \posdelta{27.3} \\
6  & \textsc{Mistral-7B-Instruct-v0.3} \citep{jiang2023mistral7b}        & 56.7 & 37.3 & \negdelta{19.4} & 32.7 & 69.1 & \posdelta{36.4} \\
7  & \textsc{Qwen1.5-MoE-A2.7B-Chat} \citep{bai2023qwen}          & 59.7 & 25.4 & \negdelta{34.3} & 14.5 & 38.2 & \posdelta{23.6} \\
8  & \textsc{Qwen2.5-3B-Instruct} \citep{qwen2025qwen25technicalreport}             & 61.2 & 26.9 & \negdelta{34.3} & 27.3 & 65.5 & \posdelta{38.2} \\
9  & \textsc{Llama-3-8B-Instruct} \citep{grattafiori2024llama}             & 65.7 & 38.8 & \negdelta{26.9} & 38.2 & 80.0 & \posdelta{41.8} \\
10 & \textsc{Llama-3.1-8B-Instruct} \citep{grattafiori2024llama}           & 68.7 & 47.8 & \negdelta{20.9} & 43.6 & 72.7 & \posdelta{29.1} \\
11 & \textsc{Phi-3.5-mini-instruct} \citep{abdin2024phi3technicalreporthighly}           & 77.6 & 59.7 & \negdelta{17.9} & 45.5 & 80.0 & \posdelta{34.5} \\
12 & \textsc{Phi-3-medium-4k-instruct} \citep{abdin2024phi3technicalreporthighly}        & 80.6 & 68.7 & \negdelta{11.9} & 63.6 & 94.5 & \posdelta{30.9} \\
13 & \textsc{Qwen2.5-7B-Instruct} \citep{qwen2025qwen25technicalreport}             & 85.1 & 53.7 & \negdelta{31.3} & 45.5 & 87.3 & \posdelta{41.8} \\
14 & \textsc{Gemma-2-9B-IT} \citep{team2024gemma}                   & 85.1 & 53.7 & \negdelta{31.3} & 60.0 & 92.7 & \posdelta{32.7} \\
15 & \textsc{Qwen2.5-14B-Instruct} \citep{qwen2025qwen25technicalreport}            & 86.6 & 59.7 & \negdelta{26.9} & 74.5 & 90.9 & \posdelta{16.4} \\
16 & \textsc{Yi-1.5-9B-Chat} \citep{young2024yi}                  & 88.1 & 37.3 & \negdelta{50.7} & 58.2 & 87.3 & \posdelta{29.1} \\
\midrule
& \textbf{Mean across models}
& --- & --- & \textbf{\negdelta{23.9}}
& --- & --- & \textbf{\posdelta{28.5}} \\
\bottomrule
\end{tabular}

\end{table*}

\begin{table*}[h]
\caption{
Hypothesis-level rewriting outcomes. $Q$ counts accepted rewrites;
transition counts pool 17 models on GSM8K and 16 on BBH-structured.
For each hypothesis $h$, let $n_{CC}$, $n_{CW}$, $n_{WW}$, and
$n_{WC}$ denote correct-to-correct, correct-to-incorrect,
incorrect-to-incorrect, and incorrect-to-correct counts, respectively,
and let $N=n_{CC}+n_{CW}+n_{WW}+n_{WC}$.
$\Delta\mathrm{Acc}_h=100(n_{WC}-n_{CW})/N$ is rewritten minus
original accuracy on the percentage scale, not a relative percentage change.
The success rate $\mathrm{SR}_h$ is
$100(n_{CW}+n_{WW})/N$ for Increase and
$100(n_{WC}+n_{CC})/N$ for Decrease, including unchanged outcomes.
Overall pools all observations within each dataset and direction.
A dash denotes no accepted rewrites.
Original outcomes come from historical RouterEval evaluations.
}
\label{tab:hypothesis-rewrite-transitions}
\medskip
\centering
\small
\setlength{\tabcolsep}{4pt}
\renewcommand{\arraystretch}{1.12}

\begin{tabular}{crrrccrrrcc}
\toprule
&
\multicolumn{5}{c}{\textbf{Increase}} &
\multicolumn{5}{c}{\textbf{Decrease}} \\
\cmidrule(lr){2-6}\cmidrule(lr){7-11}

\textbf{\#} &
\textbf{Q} &
\textcolor{black}{\ding{51}\,$\to$\,\ding{55}} &
\textcolor{black}{\ding{55}\,$\to$\,\ding{51}} &
$\Delta\mathbf{Acc}_h$ &
$\mathbf{SR}_h$ &
\textbf{Q} &
\textcolor{black}{\ding{51}\,$\to$\,\ding{55}} &
\textcolor{black}{\ding{55}\,$\to$\,\ding{51}} &
$\Delta\mathbf{Acc}_h$ &
$\mathbf{SR}_h$ \\

\midrule
\multicolumn{11}{c}{\textbf{GSM8K}} \\
\midrule
1  & 11 & 46  & 9  & \negdelta{19.7} & 75.9
   & 0  & 0   & 0   & --- & --- \\
2  & 11 & 48  & 5  & \negdelta{22.99} & 83.4
   & 0  & 0   & 0   & --- & --- \\
3  & 9  & 37  & 11 & \negdelta{16.99} & 68.0
   & 2  & 0   & 10  & \posdelta{29.41} & 29.4 \\
4  & 28 & 97  & 40 & \negdelta{11.97} & 68.3
   & 5  & 2   & 25  & \posdelta{27.06} & 36.5 \\
5  & 3  & 22  & 1  & \negdelta{41.18} & 90.2
   & 3  & 1   & 18  & \posdelta{33.33} & 56.9 \\
6  & 32 & 102 & 55 & \negdelta{8.64} & 61.2
   & 8  & 4   & 60  & \posdelta{41.18} & 66.9 \\
7  & 5  & 39  & 0  & \negdelta{45.88} & 80.0
   & 25 & 22  & 156 & \posdelta{31.53} & 63.8 \\
8  & 1  & 5   & 1  & \negdelta{23.53} & 76.5
   & 3  & 3   & 13  & \posdelta{19.61} & 39.2 \\
9  & 10 & 37  & 9  & \negdelta{16.47} & 84.7
   & 0  & 0   & 0   & --- & --- \\
10 & 31 & 103 & 39 & \negdelta{12.14} & 67.0
   & 5  & 3   & 32  & \posdelta{34.12} & 57.6 \\
\midrule
\textbf{Overall}
& \textbf{141} & \textbf{536} & \textbf{170}
& \negdelta{\textbf{15.27}} & \textbf{70.3}
& \textbf{51} & \textbf{35} & \textbf{314}
& \posdelta{\textbf{32.18}} & \textbf{57.8} \\

\midrule
\multicolumn{11}{c}{\textbf{BBH-structured}} \\
\midrule
1 & 8  & 43 & 10 & \negdelta{25.78} & 60.2
  & 8  & 10 & 42 & \posdelta{25.00} & 50.8 \\
2 & 11 & 36 & 13 & \negdelta{13.07} & 55.1
  & 2  & 0  & 15 & \posdelta{46.88} & 56.2 \\
3 & 5  & 29 & 5  & \negdelta{30.00} & 62.5
  & 7  & 5  & 31 & \posdelta{23.21} & 65.2 \\
4 & 9  & 38 & 8  & \negdelta{20.83} & 68.1
  & 12 & 9  & 59 & \posdelta{26.04} & 67.2 \\
5 & 15 & 87 & 23 & \negdelta{26.67} & 66.7
  & 8  & 12 & 41 & \posdelta{22.66} & 61.7 \\
6 & 3  & 16 & 7  & \negdelta{18.75} & 52.1
  & 2  & 4  & 15 & \posdelta{34.38} & 62.5 \\
7 & 14 & 71 & 15 & \negdelta{25.00} & 64.3
  & 3  & 0  & 15 & \posdelta{31.25} & 64.6 \\
8 & 2  & 18 & 1  & \negdelta{53.13} & 84.4
  & 13 & 4  & 77 & \posdelta{35.10} & 76.9 \\
\midrule
\textbf{Overall}
& \textbf{67} & \textbf{338} & \textbf{82}
& \negdelta{\textbf{23.88}} & \textbf{63.2}
& \textbf{55} & \textbf{44} & \textbf{295}
& \posdelta{\textbf{28.52}} & \textbf{65.3} \\
\bottomrule
\end{tabular}
\end{table*}
\clearpage
\section{All Generated Hypotheses}\label{app:all_hypotheses}
\begingroup
\scriptsize
\setlength{\tabcolsep}{3pt}
\renewcommand{\arraystretch}{1.14}
\setlength{\LTcapwidth}{\textwidth}
\begin{longtable}{@{}c>{\raggedright\arraybackslash}p{0.56\textwidth}ccc@{}}
\caption{All 20 final HypoGenic hypotheses on GSM8K, in exported order. SEP is the mean IRT difficulty of matched minus unmatched test items (95\% bootstrap CI); Cohen's $d$ is the standardized difference, and coverage is the matched-item percentage. Undefined statistics are shown as ---; CI and $d$ are omitted when either group has fewer than two items.}\label{tab:all-hypotheses-hypogenic-gsm8k}\\
\toprule
\textbf{\#} & \multicolumn{1}{c}{\textbf{Hypotheses}} &
$\mathbf{SEP}_{\scriptscriptstyle[95\%\,\mathrm{CI}]}$ &
\textbf{Cohen's $d$} & \textbf{Cov. (\%)} \\
\midrule

\multicolumn{5}{c}{\textbf{GSM8K}}\\\midrule
\endfirsthead
\multicolumn{5}{l}{\footnotesize HypoGenic -- GSM8K (continued)}\\
\toprule
\textbf{\#} & \multicolumn{1}{c}{\textbf{Hypotheses}} &
$\mathbf{SEP}_{\scriptscriptstyle[95\%\,\mathrm{CI}]}$ &
\textbf{Cohen's $d$} & \textbf{Cov. (\%)} \\
\midrule

\endhead
\midrule\multicolumn{5}{r}{\scriptsize Continued on next page}\\\endfoot
\bottomrule\endlastfoot
1 & \hyp{Questions involving counter-intuitive operations or implicit variable relationships, such as returned items or equating frequency of work to earnings, are more likely to be rated as difficulty level 3.} & \sepwithci{2.036}{1.079}{2.975} & 1.154 & 6.0 \\[2pt]
2 & \hyp{Questions that require tracking and aggregating multiple entities or sub-totals over several steps tend to be rated as moderate to high difficulty.} & \sepwithci{1.442}{0.957}{1.905} & 0.855 & 40.5 \\[2pt]
3 & \hyp{The inclusion of negative numbers, unit conversions, or complex ratios significantly increases the difficulty level of a question.} & \sepwithci{1.052}{0.263}{1.883} & 0.584 & 10.5 \\[2pt]
4 & \hyp{Questions that require tracking and combining multiple sub-totals before reaching the final answer are more difficult than those with a single linear calculation path.} & \sepwithci{1.291}{0.798}{1.774} & 0.753 & 41.0 \\[2pt]
5 & \hyp{Problems that involve implicit conversions or understanding of units, such as 'dozen' or 'a week', often have a higher difficulty level.} & \sepwithci{0.718}{-0.130}{1.607} & 0.396 & 11.5 \\[2pt]
6 & \hyp{Questions requiring more than three distinct mathematical operations tend to have a higher difficulty level.} & \sepwithci{1.520}{1.058}{1.996} & 0.912 & 42.5 \\[2pt]
7 & \hyp{Questions with a larger number of distinct steps or operations required to reach the final answer generally have a higher difficulty level.} & \sepwithci{1.762}{1.282}{2.259} & 1.080 & 32.5 \\[2pt]
8 & \hyp{Questions that require tracking multiple overlapping categories or states, such as different genders and species of fish, tend to have the highest difficulty level.} & \sepwithci{1.502}{0.174}{2.718} & 0.828 & 2.5 \\[2pt]
9 & \hyp{Questions that require simple, sequential addition and subtraction operations without complex conversions or intermediate reasoning steps tend to have a lower difficulty level.} & \sepwithci{-1.465}{-1.939}{-1.008} & -0.851 & 23.0 \\[2pt]
10 & \hyp{Questions with a larger number of distinct arithmetic operations or intermediate steps generally have a higher difficulty level.} & \sepwithci{1.584}{1.125}{2.053} & 0.960 & 44.5 \\[2pt]
11 & \hyp{Questions involving four or more distinct mathematical operations or intermediate steps are more likely to be rated as difficulty level 3.} & \sepwithci{1.572}{1.118}{2.046} & 0.951 & 44.0 \\[2pt]
12 & \hyp{Questions with more than three distinct calculation steps tend to be more difficult.} & \sepwithci{1.479}{0.995}{1.934} & 0.884 & 46.5 \\[2pt]
13 & \hyp{Questions with a larger number of sequential arithmetic steps (e.g., calculating values for five consecutive days) tend to be more difficult.} & \sepwithci{2.000}{1.354}{2.704} & 1.182 & 14.0 \\[2pt]
14 & \hyp{Questions with multiple sequential steps involving different operations (e.g., addition followed by multiplication and then subtraction) are more difficult.} & \sepwithci{1.203}{0.731}{1.677} & 0.696 & 54.5 \\[2pt]
15 & \hyp{Questions involving percentages or interest rates are more likely to be rated as difficulty level 3.} & \sepwithci{0.340}{-0.339}{1.033} & 0.186 & 15.0 \\[2pt]
16 & \hyp{Questions that require setting up algebraic equations to solve for unknown variables, particularly involving time shifts like future ages, are associated with higher difficulty levels.} & \sepwithci{1.175}{0.522}{1.833} & 0.649 & 5.5 \\[2pt]
17 & \hyp{The presence of conditional logic or threshold-based rules, such as applying an operation only if a certain value is exceeded, increases the difficulty level of the question.} & \sepwithci{0.366}{-1.520}{1.669} & 0.200 & 1.5 \\[2pt]
18 & \hyp{Questions that require multiple unit conversions or rate calculations across different time scales tend to have a higher difficulty level.} & \sepwithci{1.060}{-0.113}{2.348} & 0.586 & 6.5 \\[2pt]
19 & \hyp{Questions that require more than three distinct mathematical operations to reach the final answer are generally harder.} & \sepwithci{1.328}{0.853}{1.816} & 0.779 & 43.5 \\[2pt]
20 & \hyp{Problems that can be solved with a direct, single-layer sum of products are generally the easiest and receive a difficulty level of 1.} & \sepwithci{-1.178}{-1.753}{-0.611} & -0.659 & 12.0 \\[2pt]
\end{longtable}
\endgroup

\clearpage
\begingroup
\scriptsize
\setlength{\tabcolsep}{3pt}
\renewcommand{\arraystretch}{1.14}
\setlength{\LTcapwidth}{\textwidth}
\begin{longtable}{@{}c>{\raggedright\arraybackslash}p{0.56\textwidth}ccc@{}}
\caption{All 20 final HypoGenic hypotheses on BBH-structured, in exported order. SEP is the mean IRT difficulty of matched minus unmatched test items (95\% bootstrap CI); Cohen's $d$ is the standardized difference, and coverage is the matched-item percentage. Statistics exclude missing labels (275 of 280 test items have usable labels). Undefined statistics are shown as ---; CI and $d$ are omitted when either group has fewer than two items.}\label{tab:all-hypotheses-hypogenic-bbh-structured}\\
\toprule
\textbf{\#} & \multicolumn{1}{c}{\textbf{Hypotheses}} &
$\mathbf{SEP}_{\scriptscriptstyle[95\%\,\mathrm{CI}]}$ &
\textbf{Cohen's $d$} & \textbf{Cov. (\%)} \\
\midrule

\multicolumn{5}{c}{\textbf{BBH-structured}}\\\midrule
\endfirsthead
\multicolumn{5}{l}{\footnotesize HypoGenic -- BBH-structured (continued)}\\
\toprule
\textbf{\#} & \multicolumn{1}{c}{\textbf{Hypotheses}} &
$\mathbf{SEP}_{\scriptscriptstyle[95\%\,\mathrm{CI}]}$ &
\textbf{Cohen's $d$} & \textbf{Cov. (\%)} \\
\midrule

\endhead
\midrule\multicolumn{5}{r}{\scriptsize Continued on next page}\\\endfoot
\bottomrule\endlastfoot
1 & \hyp{Questions where the final answer is explicitly stated verbatim in the prompt text are easier than those requiring implicit logical deduction.} & \sepwithci{-0.949}{-1.219}{-0.672} & -0.865 & 30.2 \\[2pt]
2 & \hyp{Questions that require tracking and ordering five or more entities are generally harder than those involving fewer entities, unless the answer is explicitly stated.} & \sepwithci{0.824}{0.555}{1.096} & 0.743 & 44.0 \\[2pt]
3 & \hyp{Logical deduction tasks that require combining multiple relative constraints to deduce a complete sequence are typically rated at difficulty level 3.} & \sepwithci{0.946}{0.634}{1.253} & 0.859 & 28.4 \\[2pt]
4 & \hyp{Questions that require sorting numerical values or combining multiple relational constraints to find an intermediate position tend to be rated at the highest difficulty level.} & \sepwithci{1.322}{1.029}{1.616} & 1.261 & 21.5 \\[2pt]
5 & \hyp{Questions that require deducing the relative order of objects from multiple constraints are generally more difficult than questions that explicitly state the answer or involve simple counting.} & \sepwithci{0.170}{-0.112}{0.450} & 0.144 & 48.4 \\[2pt]
6 & \hyp{Questions involving a larger number of objects or constraints (e.g., five or seven objects) tend to have a higher difficulty level than those with fewer objects (e.g., three objects), unless the answer is explicitly stated.} & \sepwithci{0.692}{0.427}{0.960} & 0.613 & 47.6 \\[2pt]
7 & \hyp{Logical deduction tasks involving a larger number of objects or more complex relative positioning constraints are associated with higher difficulty levels.} & \sepwithci{1.060}{0.750}{1.365} & 0.978 & 26.9 \\[2pt]
8 & \hyp{Logical deduction tasks that require chaining multiple relative position statements (e.g., 'A is left of B', 'B is left of C') to find an unanchored object's position are more difficult (level 3) than those with fewer steps.} & \sepwithci{0.937}{0.614}{1.260} & 0.849 & 27.6 \\[2pt]
9 & \hyp{Questions with a larger number of objects to order (e.g., 5 or 7) are generally more difficult than those with fewer objects (e.g., 3), unless the answer is explicitly given.} & \sepwithci{0.677}{0.400}{0.951} & 0.598 & 45.8 \\[2pt]
10 & \hyp{Questions that require combining multiple constraints to deduce a full sequence or count items are more difficult than those that ask for a single explicit fact.} & \sepwithci{0.452}{0.169}{0.722} & 0.390 & 53.8 \\[2pt]
11 & \hyp{Questions involving a larger number of objects or events to track tend to have a higher difficulty level.} & \sepwithci{1.365}{1.139}{1.587} & 1.413 & 56.0 \\[2pt]
12 & \hyp{Questions where the answer is explicitly stated in the text without requiring any deductive reasoning are typically assigned a lower difficulty level.} & \sepwithci{-1.295}{-1.546}{-1.040} & -1.225 & 20.7 \\[2pt]
13 & \hyp{Questions where the answer is explicitly stated in the prompt (e.g., 'The cardinal is the fourth from the left') are easier, but might still be rated as difficulty 2 if they are embedded in a complex paragraph.} & \sepwithci{-1.121}{-1.388}{-0.853} & -1.015 & 16.7 \\[2pt]
14 & \hyp{Questions where the answer is explicitly stated in the prompt without requiring any logical deduction or inference tend to have lower difficulty levels (1-2).} & \sepwithci{-1.295}{-1.552}{-1.045} & -1.225 & 20.7 \\[2pt]
15 & \hyp{Questions requiring the model to chain multiple logical constraints together to deduce an unstated sequence are harder than those where the answer is explicitly stated in the text.} & \sepwithci{0.110}{-0.167}{0.384} & 0.093 & 53.8 \\[2pt]
16 & \hyp{Questions where the correct answer is explicitly stated verbatim in the premise are the easiest and have a difficulty level of 1.} & \sepwithci{-1.273}{-1.526}{-1.025} & -1.193 & 19.6 \\[2pt]
17 & \hyp{Questions that require identifying an object based on a relative position (e.g., 'second from the right') and then deducing the remaining positions are harder than those with direct positional statements.} & \sepwithci{1.037}{0.725}{1.351} & 0.943 & 21.8 \\[2pt]
18 & \hyp{Questions that require combining multiple constraints to find an unstated missing interval are harder than those that ask for a direct lookup.} & \sepwithci{0.201}{-0.032}{0.433} & 0.170 & 22.9 \\[2pt]
19 & \hyp{Tasks that require deducing the position of an object through a chain of relative spatial or temporal constraints are more likely to be rated at the highest difficulty level (level 3).} & \sepwithci{0.872}{0.560}{1.184} & 0.784 & 29.5 \\[2pt]
20 & \hyp{Questions with a large number of distractors or irrelevant information to filter through before counting tend to be more difficult.} & \sepwithci{0.562}{-0.023}{1.199} & 0.479 & 6.2 \\[2pt]
\end{longtable}
\endgroup

\clearpage
\begingroup
\scriptsize
\setlength{\tabcolsep}{3pt}
\renewcommand{\arraystretch}{1.14}
\setlength{\LTcapwidth}{\textwidth}
\begin{longtable}{@{}c>{\raggedright\arraybackslash}p{0.56\textwidth}ccc@{}}
\caption{All 20 final HypoGenic hypotheses on Winogrande, in exported order. SEP is the mean IRT difficulty of matched minus unmatched test items (95\% bootstrap CI); Cohen's $d$ is the standardized difference, and coverage is the matched-item percentage. Undefined statistics are shown as ---; CI and $d$ are omitted when either group has fewer than two items.}\label{tab:all-hypotheses-hypogenic-winogrande}\\
\toprule
\textbf{\#} & \multicolumn{1}{c}{\textbf{Hypotheses}} &
$\mathbf{SEP}_{\scriptscriptstyle[95\%\,\mathrm{CI}]}$ &
\textbf{Cohen's $d$} & \textbf{Cov. (\%)} \\
\midrule

\multicolumn{5}{c}{\textbf{Winogrande}}\\\midrule
\endfirsthead
\multicolumn{5}{l}{\footnotesize HypoGenic -- Winogrande (continued)}\\
\toprule
\textbf{\#} & \multicolumn{1}{c}{\textbf{Hypotheses}} &
$\mathbf{SEP}_{\scriptscriptstyle[95\%\,\mathrm{CI}]}$ &
\textbf{Cohen's $d$} & \textbf{Cov. (\%)} \\
\midrule

\endhead
\midrule\multicolumn{5}{r}{\scriptsize Continued on next page}\\\endfoot
\bottomrule\endlastfoot
1 & \hyp{Questions requiring the model to infer physical constraints or spatial relationships, such as a bulky object breaking a band or a minuscule ring not fitting a gem, are typically harder.} & \sepwithci{0.258}{-0.400}{0.933} & 0.153 & 10.6 \\[2pt]
2 & \hyp{Increased syntactic distance or the use of complex clause structures between the target blank and the contextual clues increases the difficulty level.} & $0.263$$_{[\text{---}]}$ & --- & 0.4 \\[2pt]
3 & \hyp{Questions containing explicit causal conjunctions like 'because' or 'so' that directly link an action to its immediate outcome are more likely to be rated as difficulty level 1.} & \sepwithci{0.058}{-0.385}{0.486} & 0.034 & 65.4 \\[2pt]
4 & \hyp{The presence of negation, sarcasm, or counter-intuitive scenarios increases the difficulty level of the question.} & \sepwithci{0.070}{-0.748}{0.947} & 0.041 & 4.7 \\[2pt]
5 & \hyp{Questions that rely on direct, common-sense word associations, such as 'chainsaw' and 'tree cutter', are easier (level 1) than those requiring multi-step logical deductions.} & \sepwithci{-0.919}{-1.573}{-0.138} & -0.550 & 7.1 \\[2pt]
6 & \hyp{Questions where the pronoun resolution depends on understanding basic emotional reactions to direct actions, such as lying to avoid someone's anger, are generally easier.} & \sepwithci{0.287}{-0.614}{1.277} & 0.170 & 4.7 \\[2pt]
7 & \hyp{Questions where the causal link is explicitly stated using simple, direct adjectives (e.g., 'too large', 'tasteless', 'too awkward') are generally the easiest (level 1).} & \sepwithci{0.205}{-0.499}{0.935} & 0.122 & 11.0 \\[2pt]
8 & \hyp{Questions that explicitly contrast two entities with well-known attributes, like clear versus dark liquors or heavy versus light objects, are easier to resolve (level 2).} & \sepwithci{-0.475}{-0.942}{0.032} & -0.283 & 15.7 \\[2pt]
9 & \hyp{Questions where the blank can be resolved by matching a simple descriptive trait to a stated outcome are typically of moderate difficulty.} & \sepwithci{-0.246}{-0.704}{0.224} & -0.146 & 18.9 \\[2pt]
10 & \hyp{Questions involving human names and straightforward cause-and-effect relationships regarding personal actions tend to be of moderate difficulty (level 2).} & \sepwithci{-0.236}{-0.684}{0.208} & -0.140 & 64.2 \\[2pt]
11 & \hyp{Questions that rely on specific external commonsense knowledge about objects, such as stencils or seeing-eye dogs, usually have moderate difficulty (level 2).} & \sepwithci{-0.152}{-0.720}{0.423} & -0.090 & 15.0 \\[2pt]
12 & \hyp{Questions where the correct referent can be resolved through simple syntactic parallelism or explicit contrast in the sentence structure tend to be easier (level 1).} & \sepwithci{-0.073}{-0.574}{0.451} & -0.043 & 13.4 \\[2pt]
13 & \hyp{Questions involving direct physical properties or simple cause-and-effect relationships, such as ice being hard or stencils making neat letters, are generally easier (level 2).} & \sepwithci{-0.109}{-0.588}{0.390} & -0.065 & 24.8 \\[2pt]
14 & \hyp{Questions involving the spatial tracking of objects transferred between multiple actors tend to be harder.} & $0.263$$_{[\text{---}]}$ & --- & 0.4 \\[2pt]
15 & \hyp{Questions that rely on understanding implicit social norms or specific background knowledge, like the purpose of an etiquette school, tend to be harder (level 3).} & \sepwithci{-0.145}{-0.887}{0.618} & -0.086 & 4.3 \\[2pt]
16 & \hyp{Questions that rely on direct, explicit causal links stated in the text, such as a garment falling off because it is loose, are the easiest (level 1).} & \sepwithci{0.006}{-0.460}{0.483} & 0.004 & 27.2 \\[2pt]
17 & \hyp{Questions that require reasoning about physical constraints or highly context-dependent attributes of inanimate objects tend to be the hardest (level 3).} & \sepwithci{0.308}{-0.392}{1.029} & 0.183 & 7.5 \\[2pt]
18 & \hyp{Questions where the correct answer relies on understanding the negation of a typical expectation, such as a curable illness eliciting less sympathy, tend to be harder.} & $-0.670$$_{[\text{---}]}$ & --- & 0.4 \\[2pt]
19 & \hyp{Questions that involve straightforward, direct cause-and-effect relationships without complex contrasting conditions are generally easier and have a difficulty level of 1.} & \sepwithci{-0.243}{-0.677}{0.197} & -0.144 & 37.0 \\[2pt]
20 & \hyp{Questions that require multi-step logical deduction involving contrast words like 'although' or 'while' are more likely to have a difficulty level of 3.} & \sepwithci{-0.284}{-1.069}{0.481} & -0.169 & 5.5 \\[2pt]
\end{longtable}
\endgroup

\clearpage
\begingroup
\scriptsize
\setlength{\tabcolsep}{3pt}
\renewcommand{\arraystretch}{1.14}
\setlength{\LTcapwidth}{\textwidth}
\begin{longtable}{@{}c>{\raggedright\arraybackslash}p{0.56\textwidth}ccc@{}}
\caption{The 20 explained HypotheSAEs features on GSM8K. SEP is the mean IRT difficulty of matched minus unmatched test items (95\% bootstrap CI); Cohen's $d$ is the standardized difference, and coverage is the matched-item percentage. Matching means SAE activation $>0$, not LLM-judged agreement with the explanation. Undefined statistics are shown as ---; CI and $d$ are omitted when either group has fewer than two items.}\label{tab:all-hypotheses-hypothesaes-gsm8k}\\
\toprule
\textbf{\#} & \multicolumn{1}{c}{\textbf{Hypotheses}} &
$\mathbf{SEP}_{\scriptscriptstyle[95\%\,\mathrm{CI}]}$ &
\textbf{Cohen's $d$} & \textbf{Cov. (\%)} \\
\midrule

\multicolumn{5}{c}{\textbf{GSM8K}}\\\midrule
\endfirsthead
\multicolumn{5}{l}{\footnotesize HypotheSAEs -- GSM8K (continued)}\\
\toprule
\textbf{\#} & \multicolumn{1}{c}{\textbf{Hypotheses}} &
$\mathbf{SEP}_{\scriptscriptstyle[95\%\,\mathrm{CI}]}$ &
\textbf{Cohen's $d$} & \textbf{Cov. (\%)} \\
\midrule

\endhead
\midrule\multicolumn{5}{r}{\scriptsize Continued on next page}\\\endfoot
\bottomrule\endlastfoot
1 & \hyp{no coherent item-level property}\par\textnormal{Trigger: the final digit of the last number in an arithmetic expression, immediately preceding the equals sign.} & --- & --- & 100.0 \\[2pt]
2 & \hyp{no coherent item-level property}\par\textnormal{Trigger: The space token immediately following the '\#\#\#\#' marker before the final answer in GSM8K solutions.} & \sepwithci{-0.579}{-1.576}{0.434} & -0.318 & 94.0 \\[2pt]
3 & \hyp{The question involves specific times of day formatted with a colon.}\par\textnormal{Trigger: The colon ':' used in time formats (e.g., 5:00, 10:00).} & \sepwithci{1.523}{0.546}{2.600} & 0.855 & 8.0 \\[2pt]
4 & \hyp{no coherent item-level property}\par\textnormal{Trigger: A space character immediately preceding a number.} & \sepwithci{1.072}{0.561}{1.576} & 0.612 & 39.0 \\[2pt]
5 & \hyp{no coherent item-level property}\par\textnormal{Trigger: The equals sign (' =') immediately preceding a calculator annotation (' <<') in the reference solution.} & \sepwithci{0.136}{-0.356}{0.624} & 0.074 & 53.0 \\[2pt]
6 & \hyp{no coherent item-level property}\par\textnormal{Trigger: the first digit of a multi-digit number, particularly in the context of mathematical equations or quantities} & \sepwithci{0.533}{0.037}{1.033} & 0.295 & 56.0 \\[2pt]
7 & \hyp{no coherent item-level property}\par\textnormal{Trigger: the comma token ',' in lists or clauses} & \sepwithci{0.410}{-0.173}{0.992} & 0.226 & 69.5 \\[2pt]
8 & \hyp{no coherent item-level property}\par\textnormal{Trigger: digits inside the result part of a calculator annotation (<<...=...>>)} & \sepwithci{1.074}{0.411}{1.748} & 0.603 & 19.0 \\[2pt]
9 & \hyp{The reference solution contains a mathematical expression followed by '=\$<<' or '=<<' where the expression ends in a multi-digit number or a number with a decimal.}\par\textnormal{Trigger: The last digit of a number immediately preceding the '=\$<<' or '=<<' calculator annotation in GSM8K solutions.} & \sepwithci{0.690}{-0.340}{1.725} & 0.379 & 93.0 \\[2pt]
10 & \hyp{The reference solution contains mathematical expressions with parentheses or brackets.}\par\textnormal{Trigger: the closing bracket or parenthesis in a mathematical expression, or the final character before it} & \sepwithci{0.970}{0.489}{1.467} & 0.550 & 52.0 \\[2pt]
11 & \hyp{The reference solution contains an intermediate or final calculation resulting in a number starting with 18 or 19.}\par\textnormal{Trigger: The '=' token inside the << >> calculator annotation when the result starts with '18' or '19'.} & \sepwithci{0.822}{0.318}{1.326} & 0.460 & 63.5 \\[2pt]
12 & \hyp{The question or reference solution involves numbers with zeros, such as decimals (e.g., 0.03, 1.04) or hundreds (e.g., 105, 106, 2000).}\par\textnormal{Trigger: The digit '0' or a decimal point '.' when it is part of a number containing a zero, especially in the context of decimals or hundreds.} & $1.258$$_{[\text{---}]}$ & --- & 99.5 \\[2pt]
13 & \hyp{no coherent item-level property}\par\textnormal{Trigger: Tokens that are part of the narrative or descriptive text in math word problems, often nouns or prepositions following numbers or quantities.} & \sepwithci{0.136}{-0.768}{0.972} & 0.074 & 88.5 \\[2pt]
14 & \hyp{no coherent item-level property}\par\textnormal{Trigger: Sentence-ending punctuation (periods) and subsequent whitespace (spaces or newlines) in the problem description or solution text.} & \sepwithci{0.653}{0.123}{1.187} & 0.362 & 36.0 \\[2pt]
15 & \hyp{The question involves financial loss, negative values, or calculating a difference/change in money.}\par\textnormal{Trigger: Tokens related to financial loss, negative values, or the start of the answer section in problems involving money or negative numbers.} & \sepwithci{0.695}{-0.163}{1.501} & 0.382 & 89.5 \\[2pt]
16 & \hyp{The question involves a character planning, intending, or being able to do something in the future.}\par\textnormal{Trigger: The tokens ' to', ' will', and ' can' when used as auxiliary verbs or infinitives indicating future action or intent.} & \sepwithci{0.973}{0.394}{1.559} & 0.547 & 25.0 \\[2pt]
17 & \hyp{The question involves calculating time, dates, or durations.}\par\textnormal{Trigger: Prepositions and temporal markers indicating time or sequence, such as 'at', 'before', 'on', 'from', and 'late'.} & \sepwithci{1.331}{0.498}{2.202} & 0.752 & 13.5 \\[2pt]
18 & \hyp{The reference solution involves setting up and solving algebraic equations or performing arithmetic calculations with explicit operators.}\par\textnormal{Trigger: mathematical operators like +, -, and * in algebraic expressions or arithmetic calculations} & --- & --- & 100.0 \\[2pt]
19 & \hyp{no coherent item-level property}\par\textnormal{Trigger: words introducing a specific subset, condition, or variable assignment in a math word problem, such as 'with', 'total', 'of', 'the', 'weigh', 'has', 'red', 'charges', 'equal'} & \sepwithci{0.710}{-0.277}{1.627} & 0.390 & 93.5 \\[2pt]
20 & \hyp{no coherent item-level property}\par\textnormal{Trigger: The comma token ',' used as a separator in lists or as a thousands separator in numbers.} & \sepwithci{1.711}{1.393}{2.051} & 0.942 & 98.5 \\[2pt]
\end{longtable}
\endgroup

\clearpage
\begingroup
\scriptsize
\setlength{\tabcolsep}{3pt}
\renewcommand{\arraystretch}{1.14}
\setlength{\LTcapwidth}{\textwidth}
\begin{longtable}{@{}c>{\raggedright\arraybackslash}p{0.56\textwidth}ccc@{}}
\caption{The 20 explained HypotheSAEs features on BBH-structured. SEP is the mean IRT difficulty of matched minus unmatched test items (95\% bootstrap CI); Cohen's $d$ is the standardized difference, and coverage is the matched-item percentage. Matching means SAE activation $>0$, not LLM-judged agreement with the explanation. Undefined statistics are shown as ---; CI and $d$ are omitted when either group has fewer than two items.}\label{tab:all-hypotheses-hypothesaes-bbh-structured}\\
\toprule
\textbf{\#} & \multicolumn{1}{c}{\textbf{Hypotheses}} &
$\mathbf{SEP}_{\scriptscriptstyle[95\%\,\mathrm{CI}]}$ &
\textbf{Cohen's $d$} & \textbf{Cov. (\%)} \\
\midrule

\multicolumn{5}{c}{\textbf{BBH-structured}}\\\midrule
\endfirsthead
\multicolumn{5}{l}{\footnotesize HypotheSAEs -- BBH-structured (continued)}\\
\toprule
\textbf{\#} & \multicolumn{1}{c}{\textbf{Hypotheses}} &
$\mathbf{SEP}_{\scriptscriptstyle[95\%\,\mathrm{CI}]}$ &
\textbf{Cohen's $d$} & \textbf{Cov. (\%)} \\
\midrule

\endhead
\midrule\multicolumn{5}{r}{\scriptsize Continued on next page}\\\endfoot
\bottomrule\endlastfoot
1 & \hyp{The reference solution involves calculating a sum, average, or identifying a specific numerical quantity.}\par\textnormal{Trigger: The space or first digit of a number that represents the result of an arithmetic calculation or a specific numerical value in the reasoning process.} & \sepwithci{0.522}{0.277}{0.769} & 0.453 & 31.1 \\[2pt]
2 & \hyp{The reference solution contains a list of relative ordering constraints or properties separated by commas, culminating in an 'and'.}\par\textnormal{Trigger: The comma separating the penultimate and final items in a list of sequential logical deductions or conditions, typically right before 'and'.} & --- & --- & 100.0 \\[2pt]
3 & \hyp{The question asks to sort items by alphabetic order or find the second heaviest item.}\par\textnormal{Trigger: The token 'abetic' in the phrase 'alphabetic order' or the token ' second' in the phrase 'The second heaviest'.} & \sepwithci{0.869}{0.109}{1.707} & 0.745 & 3.6 \\[2pt]
4 & \hyp{The question has at least six multiple-choice options (reaching option F or beyond).}\par\textnormal{Trigger: The open parenthesis '(' for option F, or the name 'Gwen' in option D.} & \sepwithci{0.520}{0.234}{0.816} & 0.450 & 27.9 \\[2pt]
5 & \hyp{The question involves logical deduction over ordered objects, specifically identifying the position of an object in a sequence.}\par\textnormal{Trigger: the word 'the' in phrases like 'from the left' or 'from the right' within multiple-choice options} & \sepwithci{0.326}{-0.185}{0.828} & 0.278 & 93.2 \\[2pt]
6 & \hyp{The reference solution directly quotes or refers to a specific statement in the prompt that explicitly gives the answer.}\par\textnormal{Trigger: The word 'explicitly' in the phrase 'explicitly states' or 'explicitly establishes' within the explanation of a logical deduction problem.} & \sepwithci{0.346}{-0.521}{1.168} & 0.294 & 2.5 \\[2pt]
7 & \hyp{The question involves identifying or filtering objects based on their attributes (e.g., name, color, age) from a given set or table.}\par\textnormal{Trigger: The word 'named' or 'is' when identifying a specific object or entity from a table or description.} & --- & --- & 100.0 \\[2pt]
8 & \hyp{no coherent item-level property}\par\textnormal{Trigger: The tokens ' alph' (in 'alphabetic') and ' electronic' (in 'electronic store').} & \sepwithci{0.282}{-0.185}{0.893} & 0.240 & 5.4 \\[2pt]
9 & \hyp{The reference solution explicitly states the answer using a phrase like 'is the [superlative]' or 'is [name]'}\par\textnormal{Trigger: the word 'the' or 'is' immediately preceding the superlative or name that answers the question in the reference solution} & \sepwithci{0.352}{-0.526}{1.199} & 0.300 & 5.4 \\[2pt]
10 & \hyp{The question asks 'How many [adjective] things do you see' or the reference solution evaluates conditions with 'does not meet'.}\par\textnormal{Trigger: The token ' you' in the phrase 'How many [adjective] things do you see' or the token ' not' in 'does not meet'.} & \sepwithci{0.689}{0.392}{0.981} & 0.608 & 68.9 \\[2pt]
11 & \hyp{no coherent item-level property}\par\textnormal{Trigger: Tokens like '-', 'have', 'the', and ')' that appear in the middle of a sentence or phrase in the reference solution or question.} & \sepwithci{-0.296}{-0.760}{0.171} & -0.252 & 7.5 \\[2pt]
12 & \hyp{The reference solution involves identifying the leftmost object in a sequence.}\par\textnormal{Trigger: the word 'the' immediately preceding 'leftmost' or 'first' in the context of identifying the leftmost object in a sequence} & \sepwithci{-1.482}{-1.995}{-0.698} & -1.281 & 2.1 \\[2pt]
13 & \hyp{no coherent item-level property}\par\textnormal{Trigger: punctuation or short words immediately following the first sentence or phrase of the 'Answer:' section in multiple-choice reasoning tasks} & \sepwithci{-0.720}{-1.475}{0.071} & -0.613 & 1.1 \\[2pt]
14 & \hyp{The reference solution uses the phrase 'establishes that' to explain the logical deductions.}\par\textnormal{Trigger: The word 'that' following the word 'establishes' in the explanation of a logical deduction problem.} & \sepwithci{-0.637}{-1.018}{-0.250} & -0.552 & 84.3 \\[2pt]
15 & \hyp{The question asks 'how many' objects remain or meet a condition, resulting in a long list of numerical options (A through Q).}\par\textnormal{Trigger: The newline character `\textbackslash{}n` immediately preceding a multiple-choice option letter in parentheses, specifically in long lists of options (e.g., A through Q).} & \sepwithci{2.290}{1.768}{2.815} & 1.972 & 99.3 \\[2pt]
16 & \hyp{no coherent item-level property}\par\textnormal{Trigger: The hyphen in hyphenated words like 'non-black' or 'fourth-newest', or the 'ie' in 'scrunchiephone'.} & \sepwithci{0.715}{0.444}{0.986} & 0.635 & 66.4 \\[2pt]
17 & \hyp{The reference solution uses relative clauses to filter or describe items from a table or list.}\par\textnormal{Trigger: relative pronouns like 'which' or 'that', or past participles like 'listed', introducing a relative clause in the explanation of a reasoning problem} & \sepwithci{0.552}{-0.731}{1.624} & 0.471 & 2.5 \\[2pt]
18 & \hyp{The reference solution involves determining a sorted order or sequence of items.}\par\textnormal{Trigger: The period at the end of a sentence that establishes a sorted order or sequence of items.} & --- & --- & 100.0 \\[2pt]
19 & \hyp{The reference solution involves multi-step logical deduction to determine the relative order or position of objects.}\par\textnormal{Trigger: The word 'must' or 'to' in the context of logical deduction steps in the reference solution.} & \sepwithci{1.636}{1.335}{1.936} & 1.400 & 99.3 \\[2pt]
20 & \hyp{no coherent item-level property}\par\textnormal{Trigger: Tokens like 'the', 'green', 'as', 'for', 'fidget', 'mauve', 'at', 'saw' in the context of listing objects, schedules, or attributes in the reference solution or question.} & \sepwithci{-0.028}{-0.421}{0.375} & -0.024 & 3.6 \\[2pt]
\end{longtable}
\endgroup

\clearpage
\begingroup
\scriptsize
\setlength{\tabcolsep}{3pt}
\renewcommand{\arraystretch}{1.14}
\setlength{\LTcapwidth}{\textwidth}
\begin{longtable}{@{}c>{\raggedright\arraybackslash}p{0.56\textwidth}ccc@{}}
\caption{The 20 explained HypotheSAEs features on Winogrande. SEP is the mean IRT difficulty of matched minus unmatched test items (95\% bootstrap CI); Cohen's $d$ is the standardized difference, and coverage is the matched-item percentage. Matching means SAE activation $>0$, not LLM-judged agreement with the explanation. Undefined statistics are shown as ---; CI and $d$ are omitted when either group has fewer than two items.}\label{tab:all-hypotheses-hypothesaes-winogrande}\\
\toprule
\textbf{\#} & \multicolumn{1}{c}{\textbf{Hypotheses}} &
$\mathbf{SEP}_{\scriptscriptstyle[95\%\,\mathrm{CI}]}$ &
\textbf{Cohen's $d$} & \textbf{Cov. (\%)} \\
\midrule

\multicolumn{5}{c}{\textbf{Winogrande}}\\\midrule
\endfirsthead
\multicolumn{5}{l}{\footnotesize HypotheSAEs -- Winogrande (continued)}\\
\toprule
\textbf{\#} & \multicolumn{1}{c}{\textbf{Hypotheses}} &
$\mathbf{SEP}_{\scriptscriptstyle[95\%\,\mathrm{CI}]}$ &
\textbf{Cohen's $d$} & \textbf{Cov. (\%)} \\
\midrule

\endhead
\midrule\multicolumn{5}{r}{\scriptsize Continued on next page}\\\endfoot
\bottomrule\endlastfoot
1 & \hyp{The reference solution explains the reasoning by attributing a state of being or property to one of the characters (e.g., being blind, Jewish, hungry, skilled, poor).}\par\textnormal{Trigger: The copula verbs 'is', 'are', 'was', 'be' or the noun 'kids' when describing a property or state of a person in the explanation of a Winograd schema.} & \sepwithci{0.297}{-0.528}{1.064} & 0.176 & 93.3 \\[2pt]
2 & \hyp{The question involves a scenario set in a workplace, business, or commercial context.}\par\textnormal{Trigger: Tokens related to businesses, workplaces, or commercial entities (like 'factory', 'plant', 'supermarket', 'company', 'companies', 'merchant', 'office') and their preceding determiners/modifiers.} & \sepwithci{-0.208}{-1.070}{0.651} & -0.123 & 2.4 \\[2pt]
3 & \hyp{no coherent item-level property}\par\textnormal{Trigger: The feature fires on nouns or names that are the object of a preposition or part of a noun phrase in the explanation section of a Winograd schema problem.} & \sepwithci{-0.383}{-1.182}{0.563} & -0.227 & 2.4 \\[2pt]
4 & \hyp{The reference solution uses a complex sentence structure with relative clauses, modal verbs, or infinitive phrases to explain the reasoning.}\par\textnormal{Trigger: Tokens like 'would', 'which', 'to', 'that', 'can', or commas that introduce a relative clause, modal verb, or infinitive phrase explaining a causal or conditional relationship in the reasoning.} & \sepwithci{0.263}{-0.774}{1.286} & 0.156 & 4.3 \\[2pt]
5 & \hyp{no coherent item-level property}\par\textnormal{Trigger: Tokens related to quantity, degree, or specific actions (e.g., 'any', 'his', 'a', 'many', 'less', 'cure', 'pose', 'of') in the context of reasoning explanations.} & \sepwithci{-0.020}{-1.275}{1.636} & -0.012 & 2.0 \\[2pt]
6 & \hyp{no coherent item-level property}\par\textnormal{Trigger: The closing quote mark (") immediately following the correct answer word in the explanation.} & \sepwithci{-0.235}{-0.640}{0.178} & -0.139 & 41.3 \\[2pt]
7 & \hyp{no coherent item-level property}\par\textnormal{Trigger: Various function words and verbs in the middle of the sentence, such as 'had', 'ordered', 'into', 'it', 'arrested', 'and', 'for', 'of', 'ran', 'but', 'even'.} & \sepwithci{-0.961}{-2.068}{0.118} & -0.571 & 1.6 \\[2pt]
8 & \hyp{The reference solution contains a conjunction ('and' or 'or') linking two items or concepts.}\par\textnormal{Trigger: The word 'and' or 'or' used as a conjunction between two nouns or noun phrases.} & \sepwithci{-0.502}{-0.983}{0.007} & -0.299 & 17.7 \\[2pt]
9 & \hyp{no coherent item-level property}\par\textnormal{Trigger: Tokens related to specific objects or actions in the sentence, such as 'rolls', 'roller', 'toilet', 'the', 'a', 's', 'but', 'cinnamon'.} & \sepwithci{-0.997}{-1.603}{-0.366} & -0.595 & 4.7 \\[2pt]
10 & \hyp{The question involves reasoning about quantities, amounts, or volumes of substances or items.}\par\textnormal{Trigger: words related to quantity, volume, or amount, such as 'amount', 'volume', 'portions', and 'much'} & \sepwithci{0.304}{-0.655}{1.225} & 0.180 & 3.1 \\[2pt]
11 & \hyp{The question involves a scenario where someone wants, agrees, or intends to do something.}\par\textnormal{Trigger: Tokens related to wanting, agreeing, or intending to do something (e.g., 'to', 'for', 'their', 'willing', 'favored', 'wanted').} & \sepwithci{0.109}{-1.304}{1.792} & 0.065 & 2.8 \\[2pt]
12 & \hyp{The question has the blank at the very beginning of the second sentence or clause, immediately following a period.}\par\textnormal{Trigger: The underscore token ('\_' or ' \_') representing the blank in a Winograd-schema sentence.} & \sepwithci{0.123}{-0.300}{0.548} & 0.073 & 27.6 \\[2pt]
13 & \hyp{no coherent item-level property}\par\textnormal{Trigger: Tokens that are parts of words related to specific concepts like 'jew' (Jewish/jeweled), 'fever', 'pun', 'finishing', or 'ugl' (uglier).} & $-0.588$$_{[\text{---}]}$ & --- & 0.4 \\[2pt]
14 & \hyp{The reference solution explicitly refers to the 'first' or 'second' clause, part, or sentence.}\par\textnormal{Trigger: The words 'first' and 'second', typically referring to parts of a sentence or sequence.} & \sepwithci{-0.528}{-0.944}{-0.099} & -0.316 & 34.6 \\[2pt]
15 & \hyp{no coherent item-level property}\par\textnormal{Trigger: Pronouns or names referring to the subject of the sentence, often in the explanation section of a Winograd schema problem.} & \sepwithci{-0.608}{-1.304}{0.167} & -0.362 & 5.5 \\[2pt]
16 & \hyp{no coherent item-level property}\par\textnormal{Trigger: Tokens related to cooking, food, or specific ingredients, as well as contrastive conjunctions like 'whereas'.} & \sepwithci{1.282}{-1.143}{3.701} & 0.761 & 0.8 \\[2pt]
17 & \hyp{The question involves a scenario related to work, employment, commuting, or performing a specific task/job.}\par\textnormal{Trigger: the word 'to' or other words related to working, commuting, or performing an action, often in the context of employment or tasks} & \sepwithci{0.197}{-0.234}{0.643} & 0.117 & 37.4 \\[2pt]
18 & \hyp{no coherent item-level property}\par\textnormal{Trigger: Tokens like 'was', 'is', 'which', and commas in the explanation section of a Winograd schema problem.} & \sepwithci{-0.303}{-0.877}{0.304} & -0.180 & 15.4 \\[2pt]
19 & \hyp{The question sentence ends with a blank immediately followed by a period.}\par\textnormal{Trigger: The period token '.' at the end of a sentence with a blank at the end.} & \sepwithci{3.053}{2.749}{3.353} & 1.833 & 0.8 \\[2pt]
20 & \hyp{no coherent item-level property}\par\textnormal{Trigger: Tokens that are part of the reasoning chain explaining the correct answer, often comparative or descriptive words like 'more', 'less', 'maximizing', 'the', 'is', 'now', 'an', 'his'.} & \sepwithci{0.008}{-0.421}{0.452} & 0.005 & 36.2 \\[2pt]
\end{longtable}
\endgroup

\clearpage
\begingroup
\scriptsize
\setlength{\tabcolsep}{3pt}
\renewcommand{\arraystretch}{1.14}
\setlength{\LTcapwidth}{\textwidth}
\begin{longtable}{@{}c>{\raggedright\arraybackslash}p{0.56\textwidth}ccc@{}}
\caption{All 20 final NLParam hypotheses on GSM8K, in exported order. SEP is the mean IRT difficulty of matched minus unmatched test items (95\% bootstrap CI); Cohen's $d$ is the standardized difference, and coverage is the matched-item percentage. Undefined statistics are shown as ---; CI and $d$ are omitted when either group has fewer than two items.}\label{tab:all-hypotheses-nlparam-gsm8k}\\
\toprule
\textbf{\#} & \multicolumn{1}{c}{\textbf{Hypotheses}} &
$\mathbf{SEP}_{\scriptscriptstyle[95\%\,\mathrm{CI}]}$ &
\textbf{Cohen's $d$} & \textbf{Cov. (\%)} \\
\midrule

\multicolumn{5}{c}{\textbf{GSM8K}}\\\midrule
\endfirsthead
\multicolumn{5}{l}{\footnotesize NLParam -- GSM8K (continued)}\\
\toprule
\textbf{\#} & \multicolumn{1}{c}{\textbf{Hypotheses}} &
$\mathbf{SEP}_{\scriptscriptstyle[95\%\,\mathrm{CI}]}$ &
\textbf{Cohen's $d$} & \textbf{Cov. (\%)} \\
\midrule

\endhead
\midrule\multicolumn{5}{r}{\scriptsize Continued on next page}\\\endfoot
\bottomrule\endlastfoot
1 & \hyp{requires calculating an average; specifically, the question asks for the mean or average of a set of values} & \sepwithci{1.514}{0.413}{2.500} & 0.835 & 2.5 \\[2pt]
2 & \hyp{involves unit conversions; specifically, the problem requires converting between different units of measurement (e.g., minutes to hours, kb to Mb, yards to miles)} & \sepwithci{0.781}{-0.225}{1.926} & 0.430 & 8.0 \\[2pt]
3 & \hyp{contains a narrative with physical actions causing subsequent events; specifically, the text describes an action (like throwing an object) that directly causes another event (like more objects falling)} & $-1.026$$_{[\text{---}]}$ & --- & 0.5 \\[2pt]
4 & \hyp{involves multi-step arithmetic with more than three operations; specifically, the solution requires performing four or more distinct mathematical operations to reach the final answer} & \sepwithci{1.574}{1.114}{2.040} & 0.954 & 46.0 \\[2pt]
5 & \hyp{uses decimal numbers; specifically, the text contains non-integer numerical values such as 1.5, 1.25, or 2.50 in the problem description} & \sepwithci{0.045}{-0.630}{0.765} & 0.025 & 9.0 \\[2pt]
6 & \hyp{features water or liquid volume as the primary subject; specifically, the problem involves measuring, transferring, or calculating amounts of water in gallons or liters} & \sepwithci{-0.260}{-1.000}{0.482} & -0.142 & 1.0 \\[2pt]
7 & \hyp{has a shorter problem description; specifically, the question text is relatively brief and contains fewer sentences compared to other problems} & \sepwithci{-1.296}{-1.990}{-0.623} & -0.730 & 87.0 \\[2pt]
8 & \hyp{involves multiple people or entities with comparative relationships; specifically, the problem defines one person's amount or time in terms of another's (e.g., twice as long, \$0.50 more)} & \sepwithci{0.026}{-0.581}{0.645} & 0.014 & 20.5 \\[2pt]
9 & \hyp{requires tracking multiple distinct items with different prices; specifically, the text involves calculating the total cost of several different goods or services, each with its own unit price} & \sepwithci{0.084}{-0.739}{0.930} & 0.046 & 8.0 \\[2pt]
10 & \hyp{involves multiple entities with relative values; specifically, the problem defines the value of one entity in terms of another (e.g., '800 pounds more than', '4 times what Grace weighs')} & \sepwithci{-0.112}{-0.653}{0.434} & -0.061 & 27.5 \\[2pt]
11 & \hyp{involves geometric or spatial concepts; specifically, the problem mentions shapes, angles, or dimensions like length, width, or height} & \sepwithci{0.469}{-0.283}{1.235} & 0.257 & 7.5 \\[2pt]
12 & \hyp{requires tracking time or schedules; specifically, the text involves calculating durations, start times, or end times across a timeline} & \sepwithci{0.446}{-0.525}{1.468} & 0.244 & 9.0 \\[2pt]
13 & \hyp{involves multiple distinct entities or items; specifically, the problem requires tracking quantities for three or more different types of objects or people (e.g., MTB, BMX, and Trekking bikes; bicycles, tricycles, and unicycles)} & \sepwithci{-0.110}{-0.688}{0.475} & -0.060 & 20.0 \\[2pt]
14 & \hyp{requires calculating rates or unit prices; specifically, the text involves quantities expressed as 'per hour', 'per week', 'apiece', or 'each'} & \sepwithci{0.089}{-0.417}{0.600} & 0.049 & 47.0 \\[2pt]
15 & \hyp{involves fractions or percentages; specifically, the question requires calculating or reasoning with fractions, ratios, or percentages} & \sepwithci{0.617}{0.092}{1.138} & 0.342 & 40.0 \\[2pt]
16 & \hyp{requires converting units or rates; specifically, the problem involves translating between different units of measurement, time, or percentages to solve the problem} & \sepwithci{0.781}{-0.234}{1.929} & 0.430 & 8.0 \\[2pt]
17 & \hyp{requires categorizing a single type of item into multiple sub-types; specifically, the problem involves tracking a generic item (like apples or trees) that is split into specific categories (e.g., large/small, tallest/shortest/average)} & \sepwithci{0.012}{-0.631}{0.711} & 0.007 & 12.0 \\[2pt]
18 & \hyp{requires tracking items that are both added and removed; specifically, the problem involves a scenario where items are gained and then some are lost, given away, or thrown} & \sepwithci{0.589}{-0.704}{1.953} & 0.323 & 5.5 \\[2pt]
19 & \hyp{requires multi-step reasoning with intermediate variables; specifically, the problem cannot be solved in one or two simple arithmetic steps and requires keeping track of multiple intermediate quantities} & \sepwithci{1.406}{0.944}{1.872} & 0.831 & 58.0 \\[2pt]
20 & \hyp{involves large numbers; specifically, the problem requires arithmetic operations on numbers in the thousands} & \sepwithci{1.214}{0.507}{1.974} & 0.684 & 16.0 \\[2pt]
\end{longtable}
\endgroup

\clearpage
\begingroup
\scriptsize
\setlength{\tabcolsep}{3pt}
\renewcommand{\arraystretch}{1.14}
\setlength{\LTcapwidth}{\textwidth}
\begin{longtable}{@{}c>{\raggedright\arraybackslash}p{0.56\textwidth}ccc@{}}
\caption{All 20 final NLParam hypotheses on BBH-structured, in exported order. SEP is the mean IRT difficulty of matched minus unmatched test items (95\% bootstrap CI); Cohen's $d$ is the standardized difference, and coverage is the matched-item percentage. Undefined statistics are shown as ---; CI and $d$ are omitted when either group has fewer than two items.}\label{tab:all-hypotheses-nlparam-bbh-structured}\\
\toprule
\textbf{\#} & \multicolumn{1}{c}{\textbf{Hypotheses}} &
$\mathbf{SEP}_{\scriptscriptstyle[95\%\,\mathrm{CI}]}$ &
\textbf{Cohen's $d$} & \textbf{Cov. (\%)} \\
\midrule

\multicolumn{5}{c}{\textbf{BBH-structured}}\\\midrule
\endfirsthead
\multicolumn{5}{l}{\footnotesize NLParam -- BBH-structured (continued)}\\
\toprule
\textbf{\#} & \multicolumn{1}{c}{\textbf{Hypotheses}} &
$\mathbf{SEP}_{\scriptscriptstyle[95\%\,\mathrm{CI}]}$ &
\textbf{Cohen's $d$} & \textbf{Cov. (\%)} \\
\midrule

\endhead
\midrule\multicolumn{5}{r}{\scriptsize Continued on next page}\\\endfoot
\bottomrule\endlastfoot
1 & \hyp{requires determining the position of an object relative to others; specifically, the question asks for the position of an object (e.g., 'third from the left', 'rightmost') based on a set of relative ordering constraints} & \sepwithci{0.309}{0.009}{0.609} & 0.264 & 27.5 \\[2pt]
2 & \hyp{involves logical deduction of a sequence; specifically, the text requires ordering a set of objects based on a series of relative constraints} & \sepwithci{-0.048}{-0.315}{0.223} & -0.041 & 54.3 \\[2pt]
3 & \hyp{requires multi-step logical deduction to determine a sequence; specifically, the text provides a set of constraints about the relative order or age of objects and asks to identify the position of a specific object} & \sepwithci{0.134}{-0.142}{0.411} & 0.114 & 49.3 \\[2pt]
4 & \hyp{includes an example sentence explaining how to read the data; specifically, the text contains a phrase like '} & \sepwithci{-0.021}{-0.627}{0.624} & -0.018 & 6.4 \\[2pt]
5 & \hyp{has a multiple-choice format with more than five options; specifically, the text presents a question followed by six or more lettered options} & \sepwithci{0.454}{0.158}{0.737} & 0.393 & 37.5 \\[2pt]
6 & \hyp{requires determining the oldest or newest item in a sequence; specifically, the question asks to identify the item at the extreme end of an ordered list} & \sepwithci{-0.596}{-0.984}{-0.177} & -0.513 & 12.5 \\[2pt]
7 & \hyp{contains distractor objects; specifically, the text lists items that are irrelevant to the final question being asked, requiring the model to ignore them} & \sepwithci{-0.019}{-0.377}{0.341} & -0.016 & 22.9 \\[2pt]
8 & \hyp{requires deducing the order of a sequence of objects; specifically, the task involves arranging items based on relative position clues} & \sepwithci{-0.048}{-0.318}{0.218} & -0.041 & 54.3 \\[2pt]
9 & \hyp{contains the phrase 'extremely simple questions'; specifically, the task description explicitly states it is about answering extremely simple questions about objects} & \sepwithci{-0.069}{-0.456}{0.318} & -0.058 & 18.6 \\[2pt]
10 & \hyp{involves spatial reasoning with relative positions; specifically, the text requires deducing the exact order of objects based on clues like 'left of', 'rightmost', or 'second from the left'} & \sepwithci{-0.242}{-0.553}{0.060} & -0.207 & 24.3 \\[2pt]
11 & \hyp{involves multiple relative constraints; specifically, the text provides several statements comparing the positions or ages of objects (e.g., 'older than', 'to the right of') that must be combined} & \sepwithci{-0.037}{-0.310}{0.242} & -0.032 & 53.9 \\[2pt]
12 & \hyp{options are complete sentences; specifically, the multiple-choice options provided are full sentences rather than single words, numbers, or short phrases} & \sepwithci{-0.048}{-0.313}{0.218} & -0.041 & 54.3 \\[2pt]
13 & \hyp{requires identifying a gap in a schedule; specifically, the text requires finding an unaccounted time period between known activities to determine when an event could have occurred} & \sepwithci{0.144}{-0.076}{0.373} & 0.122 & 20.7 \\[2pt]
14 & \hyp{describes a fixed arrangement task; specifically, the text explicitly states that objects are 'arranged in a fixed order' or asks to deduce the position of an item in a sequence} & \sepwithci{-0.048}{-0.318}{0.221} & -0.041 & 54.3 \\[2pt]
15 & \hyp{describes a logical deduction task; specifically, the text explicitly introduces the problem as 'A logical deduction task which requires deducing the order of a sequence of objects'} & \sepwithci{-0.048}{-0.323}{0.221} & -0.041 & 54.3 \\[2pt]
16 & \hyp{requires reasoning about prices or costs; specifically, the text involves comparing the expense of different items, such as fruits at a stand} & \sepwithci{-0.212}{-0.718}{0.306} & -0.180 & 7.5 \\[2pt]
17 & \hyp{describes objects on a surface; specifically, the text begins by listing items located on a 'table', 'floor', or 'nightstand'} & \sepwithci{-0.093}{-0.494}{0.301} & -0.079 & 17.9 \\[2pt]
18 & \hyp{requires tracking multiple attributes of objects simultaneously; specifically, the text lists objects with both color and type attributes and asks a question that depends on filtering by one or both attributes} & \sepwithci{0.016}{-0.373}{0.407} & 0.014 & 19.3 \\[2pt]
19 & \hyp{contains multiple objects to order; specifically, the text describes a set of three, five, or seven objects that must be arranged in a fixed sequence} & \sepwithci{-0.048}{-0.321}{0.216} & -0.041 & 54.3 \\[2pt]
20 & \hyp{uses relative positional language; specifically, the text contains relational phrases such as 'to the left of', 'to the right of', 'above', or 'below' to establish relationships between entities} & \sepwithci{-0.101}{-0.381}{0.169} & -0.086 & 40.7 \\[2pt]
\end{longtable}
\endgroup

\clearpage
\begingroup
\scriptsize
\setlength{\tabcolsep}{3pt}
\renewcommand{\arraystretch}{1.14}
\setlength{\LTcapwidth}{\textwidth}
\begin{longtable}{@{}c>{\raggedright\arraybackslash}p{0.56\textwidth}ccc@{}}
\caption{All 20 final NLParam hypotheses on Winogrande, in exported order. SEP is the mean IRT difficulty of matched minus unmatched test items (95\% bootstrap CI); Cohen's $d$ is the standardized difference, and coverage is the matched-item percentage. Undefined statistics are shown as ---; CI and $d$ are omitted when either group has fewer than two items.}\label{tab:all-hypotheses-nlparam-winogrande}\\
\toprule
\textbf{\#} & \multicolumn{1}{c}{\textbf{Hypotheses}} &
$\mathbf{SEP}_{\scriptscriptstyle[95\%\,\mathrm{CI}]}$ &
\textbf{Cohen's $d$} & \textbf{Cov. (\%)} \\
\midrule

\multicolumn{5}{c}{\textbf{Winogrande}}\\\midrule
\endfirsthead
\multicolumn{5}{l}{\footnotesize NLParam -- Winogrande (continued)}\\
\toprule
\textbf{\#} & \multicolumn{1}{c}{\textbf{Hypotheses}} &
$\mathbf{SEP}_{\scriptscriptstyle[95\%\,\mathrm{CI}]}$ &
\textbf{Cohen's $d$} & \textbf{Cov. (\%)} \\
\midrule

\endhead
\midrule\multicolumn{5}{r}{\scriptsize Continued on next page}\\\endfoot
\bottomrule\endlastfoot
1 & \hyp{requires reasoning about inverse relationships; specifically, the text requires understanding that an increase in one quantity (e.g., eating more) implies a decrease or lack in another (e.g., having a smaller lunch or being hungry)} & \sepwithci{0.096}{-0.864}{1.203} & 0.057 & 3.9 \\[2pt]
2 & \hyp{involves negation in the premise; specifically, the text contains words like 'not', 'didn't', or 'unlike' to establish a negative condition for one of the subjects} & \sepwithci{-0.225}{-0.666}{0.233} & -0.134 & 18.5 \\[2pt]
3 & \hyp{uses dimensional or spatial adjectives; specifically, the sentence containing the blank uses words like 'small', 'big', 'broad', or 'confined' to describe the missing entity} & \sepwithci{0.300}{-0.399}{1.014} & 0.178 & 10.2 \\[2pt]
4 & \hyp{requires inferring the cause of a state or action from a contrasting pair of individuals; specifically, the text contrasts two people and requires deducing which one is responsible for a specific outcome or state based on their described characteristics} & \sepwithci{-0.229}{-0.635}{0.200} & -0.136 & 28.7 \\[2pt]
5 & \hyp{involves contrasting actions or states between two individuals; specifically, the text describes two people doing different things or having different characteristics, and the blank must be resolved by matching the correct person to the concluding action or state} & \sepwithci{-0.115}{-0.525}{0.308} & -0.068 & 36.2 \\[2pt]
6 & \hyp{involves a scenario where one person is acting upon or communicating with another person} & \sepwithci{0.407}{-0.052}{0.880} & 0.242 & 27.2 \\[2pt]
7 & \hyp{involves inanimate objects rather than human actors; specifically, the options provided for the blank are non-living things, physical items, or spaces rather than people's names} & \sepwithci{0.024}{-0.408}{0.468} & 0.014 & 39.0 \\[2pt]
8 & \hyp{requires understanding environmental or hygiene concepts; specifically, the text involves reasoning about cleanliness, sterility, pollution, or recycling to resolve the blank} & \sepwithci{0.368}{-0.405}{1.198} & 0.218 & 5.5 \\[2pt]
9 & \hyp{involves comparing two inanimate objects or locations; specifically, the text asks to resolve a pronoun or blank referring to one of two non-human entities (e.g., rooms, vehicles, containers)} & \sepwithci{0.062}{-0.370}{0.502} & 0.037 & 37.8 \\[2pt]
10 & \hyp{involves interpersonal relationships or social dynamics; specifically, the text describes interactions between two named individuals, such as giving gifts, breaking hearts, or doing deeds in someone else's name} & \sepwithci{0.452}{-0.028}{0.939} & 0.270 & 28.3 \\[2pt]
11 & \hyp{requires resolving a pronoun or blank to a person's name; specifically, the options provided are names of people rather than inanimate objects or concepts} & \sepwithci{-0.097}{-0.525}{0.331} & -0.058 & 59.1 \\[2pt]
12 & \hyp{contains a subordinate clause starting with 'because' or 'as' that explains a preference or outcome; specifically, the sentence structure relies on a causal conjunction to link the blank to the correct entity} & \sepwithci{0.065}{-0.344}{0.474} & 0.039 & 54.7 \\[2pt]
13 & \hyp{contains a contrastive conjunction; specifically, the text uses words like 'although', 'but', or 'rather than' to set up the logical relationship between the clauses} & \sepwithci{0.039}{-0.380}{0.477} & 0.023 & 33.9 \\[2pt]
14 & \hyp{involves reasoning about interpersonal relationships or actions; specifically, the sentence describes an interaction, comparison, or event involving two distinct individuals} & \sepwithci{-0.056}{-0.492}{0.377} & -0.033 & 60.2 \\[2pt]
15 & \hyp{requires resolving a pronoun or blank to a subject based on contrasting preferences or behaviors; specifically, the text contrasts two individuals' actions or preferences and requires identifying which one aligns with a given trait or subsequent action} & \sepwithci{-0.000}{-0.414}{0.432} & -0.000 & 30.7 \\[2pt]
16 & \hyp{features a 'but not' construction; specifically, the text contrasts two individuals using the phrase 'but not' to indicate that an action or state applies to one but not the other} & \sepwithci{0.110}{-0.883}{1.145} & 0.065 & 3.5 \\[2pt]
17 & \hyp{requires reasoning about the physical properties or conditions of inanimate objects; specifically, the text asks the reader to infer which object possesses a certain physical trait (e.g., being used, brand new, having faulty screws, being soft, lasting longer) based on the context provided} & \sepwithci{0.061}{-0.417}{0.550} & 0.036 & 28.0 \\[2pt]
18 & \hyp{requires resolving pronoun or blank reference based on physical properties; specifically, the text requires identifying the correct noun based on physical attributes like 'big', 'thin', or 'soft'} & \sepwithci{0.296}{-0.221}{0.836} & 0.176 & 22.4 \\[2pt]
19 & \hyp{requires resolving a pronoun or blank to a person based on social roles or interpersonal dynamics; specifically, the text requires inferring who is the employer, who is offering a gift, or who is asking for help based on the context of the interaction} & \sepwithci{0.392}{-0.097}{0.888} & 0.234 & 26.0 \\[2pt]
20 & \hyp{involves multiple objects with distinct attributes; specifically, the text describes two inanimate objects and assigns different physical attributes or actions to them, requiring the model to map the correct attribute to the correct object} & \sepwithci{0.189}{-0.277}{0.655} & 0.112 & 29.9 \\[2pt]
\end{longtable}
\endgroup

\clearpage
\begingroup
\scriptsize
\setlength{\tabcolsep}{3pt}
\renewcommand{\arraystretch}{1.14}
\setlength{\LTcapwidth}{\textwidth}
\begin{longtable}{@{}c>{\raggedright\arraybackslash}p{0.56\textwidth}ccc@{}}
\caption{All 44 hypotheses exported by the Edisa baseline on GSM8K with a maximum of 50 hypotheses, in exported order. Repeated hypothesis texts are retained. Statistics use all 200 test items. Negative SEP indicates lower mean difficulty among matched items. SEP is the mean IRT difficulty of matched minus unmatched test items (95\% bootstrap CI); Cohen's $d$ is the standardized difference, and coverage is the matched-item percentage. Undefined statistics are shown as ---; CI and $d$ are omitted when either group has fewer than two items.}\label{tab:all-hypotheses-gmm-gsm8k}\\
\toprule
\textbf{\#} & \multicolumn{1}{c}{\textbf{Hypotheses}} &
$\mathbf{SEP}_{\scriptscriptstyle[95\%\,\mathrm{CI}]}$ &
\textbf{Cohen's $d$} & \textbf{Cov. (\%)} \\
\midrule

\multicolumn{5}{c}{\textbf{GSM8K}}\\\midrule
\endfirsthead
\multicolumn{5}{l}{\footnotesize Edisa -- GSM8K (continued)}\\
\toprule
\textbf{\#} & \multicolumn{1}{c}{\textbf{Hypotheses}} &
$\mathbf{SEP}_{\scriptscriptstyle[95\%\,\mathrm{CI}]}$ &
\textbf{Cohen's $d$} & \textbf{Cov. (\%)} \\
\midrule

\endhead
\midrule\multicolumn{5}{r}{\scriptsize Continued on next page}\\\endfoot
\bottomrule\endlastfoot
1 & \hyp{The question requires calculating a total by summing multiple intermediate values derived from rates, prices, or percentages} & \sepwithci{0.197}{-0.345}{0.751} & 0.108 & 26.0 \\[2pt]
2 & \hyp{The question is a math word problem that can be solved using exactly two basic arithmetic operations} & \sepwithci{-1.583}{-2.095}{-1.054} & -0.927 & 22.0 \\[2pt]
3 & \hyp{The question involves calculating the age of one or more individuals based on relative age differences and time shifts} & \sepwithci{0.802}{0.346}{1.254} & 0.439 & 1.0 \\[2pt]
4 & \hyp{The question requires calculating a final value by determining and combining at least two intermediate quantities} & \sepwithci{1.222}{0.751}{1.693} & 0.707 & 61.0 \\[2pt]
5 & \hyp{The question is a multi-step math word problem that requires accounting for implicit entities, rounding rules, or multi-period summations} & \sepwithci{3.144}{0.823}{5.391} & 1.771 & 2.0 \\[2pt]
6 & \hyp{The question is a multi-step arithmetic word problem} & --- & --- & 100.0 \\[2pt]
7 & \hyp{The question is a multi-step arithmetic word problem that requires translating a complex or slightly ambiguous real-world scenario into a sequence of mathematical operations} & \sepwithci{1.971}{1.458}{2.492} & 1.213 & 23.5 \\[2pt]
8 & \hyp{The question involves calculating a total amount based on rates or per-unit quantities} & \sepwithci{-0.061}{-0.572}{0.432} & -0.033 & 52.0 \\[2pt]
9 & \hyp{The question is a multi-step math word problem that requires calculating intermediate values to find the final answer} & $0.429$$_{[\text{---}]}$ & --- & 99.5 \\[2pt]
10 & \hyp{The question is a multi-step arithmetic word problem that requires calculating intermediate values to find the final answer} & $0.429$$_{[\text{---}]}$ & --- & 99.5 \\[2pt]
11 & \hyp{The question requires tracking multiple quantities or a single quantity through a sequence of at least three distinct arithmetic operations or changes} & \sepwithci{1.470}{0.940}{1.984} & 0.856 & 76.0 \\[2pt]
12 & \hyp{The question requires solving a multi-step mathematical word problem involving relationships between multiple quantities} & \sepwithci{0.302}{-1.000}{1.578} & 0.165 & 96.0 \\[2pt]
13 & \hyp{The question is a multi-step arithmetic word problem that requires calculating intermediate values to find a final total or remaining amount} & \sepwithci{-0.371}{-1.009}{0.249} & -0.204 & 77.5 \\[2pt]
14 & \hyp{The question involves calculating a total amount of money based on an hourly rate or time-based rate} & \sepwithci{0.498}{-0.742}{1.967} & 0.272 & 5.0 \\[2pt]
15 & \hyp{The question is a multi-step math word problem requiring arithmetic operations} & --- & --- & 100.0 \\[2pt]
16 & \hyp{requires multiplying a quantity by a given factor or rate to find a total before performing further addition or subtraction} & \sepwithci{-0.166}{-0.674}{0.337} & -0.091 & 48.0 \\[2pt]
17 & \hyp{The question involves calculating or using rates or ratios (e.g., per hour, per minute, per gallon)} & \sepwithci{0.498}{-0.003}{0.996} & 0.275 & 59.5 \\[2pt]
18 & \hyp{The question can be solved using exactly two basic arithmetic operations} & \sepwithci{-1.531}{-2.044}{-1.013} & -0.894 & 22.5 \\[2pt]
19 & \hyp{The question is a multi-step arithmetic word problem} & --- & --- & 100.0 \\[2pt]
20 & \hyp{The question is a multi-step math word problem that can be solved using a sequence of basic arithmetic operations} & --- & --- & 100.0 \\[2pt]
21 & \hyp{The question involves calculating quantities over a sequence of days or time periods} & \sepwithci{0.617}{-0.078}{1.334} & 0.340 & 19.0 \\[2pt]
22 & \hyp{The question is a multi-step arithmetic word problem requiring the calculation of a final or original quantity based on a series of given conditions} & --- & --- & 100.0 \\[2pt]
23 & \hyp{The question requires calculating a final total or specific value by first determining intermediate quantities using multiplication, division, or percentages, and then adding them together} & \sepwithci{0.401}{-0.122}{0.926} & 0.220 & 37.5 \\[2pt]
24 & \hyp{The question requires exactly two or three basic arithmetic operations to solve} & \sepwithci{-1.483}{-1.952}{-1.019} & -0.887 & 52.5 \\[2pt]
25 & \hyp{requires solving a multi-step arithmetic word problem involving multiple parts or categories} & \sepwithci{0.421}{-0.121}{0.952} & 0.231 & 64.0 \\[2pt]
26 & \hyp{The question is a multi-step arithmetic word problem} & --- & --- & 100.0 \\[2pt]
27 & \hyp{The question involves calculating the age of one or more individuals based on relative age differences and time shifts} & \sepwithci{0.802}{0.351}{1.253} & 0.439 & 1.0 \\[2pt]
28 & \hyp{The question requires exactly two or three basic arithmetic operations to solve} & \sepwithci{-1.483}{-1.937}{-1.015} & -0.887 & 52.5 \\[2pt]
29 & \hyp{The question requires calculating an average, a rate over time, or a volume} & \sepwithci{1.524}{0.615}{2.445} & 0.854 & 7.5 \\[2pt]
30 & \hyp{involves calculating percentages of a total number of items} & \sepwithci{0.207}{-0.720}{1.151} & 0.113 & 8.5 \\[2pt]
31 & \hyp{The question is a multi-step arithmetic word problem} & --- & --- & 100.0 \\[2pt]
32 & \hyp{The question is a math word problem that can be solved using two or three basic arithmetic operations} & \sepwithci{-1.951}{-2.446}{-1.477} & -1.235 & 67.0 \\[2pt]
33 & \hyp{The question is a multi-step math word problem requiring at least four intermediate calculations to reach the final answer} & \sepwithci{1.613}{1.158}{2.086} & 0.981 & 43.0 \\[2pt]
34 & \hyp{The question is a basic arithmetic word problem requiring two or three simple operations to solve} & \sepwithci{-1.483}{-1.957}{-1.029} & -0.887 & 52.5 \\[2pt]
35 & \hyp{The question is a multi-step arithmetic word problem} & --- & --- & 100.0 \\[2pt]
36 & \hyp{The question involves calculating distance, rate, or time for a vehicle} & \sepwithci{0.169}{-1.678}{1.836} & 0.092 & 2.5 \\[2pt]
37 & \hyp{The question involves calculating the total or unit cost of items, often factoring in a promotion, discount, or bulk pricing} & \sepwithci{0.108}{-0.487}{0.709} & 0.059 & 17.5 \\[2pt]
38 & \hyp{The question is a multi-step arithmetic word problem} & --- & --- & 100.0 \\[2pt]
39 & \hyp{The question is a multi-step math word problem requiring multiple arithmetic operations to solve} & --- & --- & 100.0 \\[2pt]
40 & \hyp{The question is a multi-step math word problem that requires performing multiple arithmetic operations to find the final answer} & --- & --- & 100.0 \\[2pt]
41 & \hyp{The question involves calculating the number of students in a school or classroom setting} & \sepwithci{0.465}{-1.387}{2.290} & 0.254 & 2.0 \\[2pt]
42 & \hyp{The question requires calculating a total quantity or cost by multiplying a rate or unit amount by a number of items or time periods} & \sepwithci{-0.102}{-0.604}{0.395} & -0.056 & 49.5 \\[2pt]
43 & \hyp{The question is a multi-step math word problem requiring basic arithmetic operations to track quantities across different entities} & \sepwithci{-0.238}{-0.735}{0.261} & -0.130 & 48.5 \\[2pt]
44 & \hyp{The question involves calculating the number of people in a specific group or tracking changes in a population of people (e.g., students, passengers)} & \sepwithci{0.591}{-0.418}{1.624} & 0.324 & 10.5 \\[2pt]
\end{longtable}
\endgroup

\clearpage
\begingroup
\scriptsize
\setlength{\tabcolsep}{3pt}
\renewcommand{\arraystretch}{1.14}
\setlength{\LTcapwidth}{\textwidth}
\begin{longtable}{@{}c>{\raggedright\arraybackslash}p{0.56\textwidth}ccc@{}}
\caption{All 47 hypotheses exported by the Edisa baseline on BBH-structured with a maximum of 50 hypotheses, in exported order. Repeated hypothesis texts are retained. Statistics use all 280 test items. Negative SEP indicates lower mean difficulty among matched items. SEP is the mean IRT difficulty of matched minus unmatched test items (95\% bootstrap CI); Cohen's $d$ is the standardized difference, and coverage is the matched-item percentage. Undefined statistics are shown as ---; CI and $d$ are omitted when either group has fewer than two items.}\label{tab:all-hypotheses-gmm-bbh-structured}\\
\toprule
\textbf{\#} & \multicolumn{1}{c}{\textbf{Hypotheses}} &
$\mathbf{SEP}_{\scriptscriptstyle[95\%\,\mathrm{CI}]}$ &
\textbf{Cohen's $d$} & \textbf{Cov. (\%)} \\
\midrule

\multicolumn{5}{c}{\textbf{BBH-structured}}\\\midrule
\endfirsthead
\multicolumn{5}{l}{\footnotesize Edisa -- BBH-structured (continued)}\\
\toprule
\textbf{\#} & \multicolumn{1}{c}{\textbf{Hypotheses}} &
$\mathbf{SEP}_{\scriptscriptstyle[95\%\,\mathrm{CI}]}$ &
\textbf{Cohen's $d$} & \textbf{Cov. (\%)} \\
\midrule

\endhead
\midrule\multicolumn{5}{r}{\scriptsize Continued on next page}\\\endfoot
\bottomrule\endlastfoot
1 & \hyp{The question asks for the position of an object whose position is explicitly stated in the provided text, requiring no actual logical deduction} & \sepwithci{-0.987}{-1.274}{-0.703} & -0.872 & 12.1 \\[2pt]
2 & \hyp{The question requires deducing the left-to-right order of a sequence of books based on relative position clues} & \sepwithci{-0.014}{-0.574}{0.568} & -0.012 & 5.4 \\[2pt]
3 & \hyp{The question asks to identify the available time gap in a person's schedule when they could have visited a specific location} & \sepwithci{0.144}{-0.087}{0.377} & 0.122 & 20.7 \\[2pt]
4 & \hyp{The question asks for the position of an object, and the correct answer is explicitly stated verbatim in the provided text, requiring no actual logical deduction} & \sepwithci{-0.987}{-1.271}{-0.700} & -0.872 & 12.1 \\[2pt]
5 & \hyp{asks for the color of the left-most item in a list of objects} & \sepwithci{-1.099}{-1.317}{-0.882} & -0.936 & 0.7 \\[2pt]
6 & \hyp{The question is a logical deduction task that requires determining the relative age order of a set of vehicles at an antique car show} & \sepwithci{0.559}{0.031}{1.084} & 0.480 & 11.1 \\[2pt]
7 & \hyp{The question is a logical deduction task that requires deducing the finishing order of five golfers in a tournament based on relative positioning statements} & \sepwithci{0.167}{-0.726}{1.015} & 0.142 & 2.5 \\[2pt]
8 & \hyp{The question is a logical deduction task that requires deducing the price order of seven fruits based on a set of constraints} & \sepwithci{0.685}{-0.351}{1.347} & 0.583 & 1.1 \\[2pt]
9 & \hyp{The correct answer to the question is explicitly stated verbatim in the provided text} & \sepwithci{-1.229}{-1.470}{-0.994} & -1.145 & 18.9 \\[2pt]
10 & \hyp{asks the user to count the number of objects that are neither of two specified colors} & \sepwithci{0.365}{0.039}{0.793} & 0.310 & 1.1 \\[2pt]
11 & \hyp{asks a question based on a provided table of penguins and their attributes} & \sepwithci{-0.115}{-0.714}{0.550} & -0.097 & 6.1 \\[2pt]
12 & \hyp{The question asks for the position of a specific person, and the correct answer is explicitly stated verbatim in the provided text, requiring no actual logical deduction} & \sepwithci{-1.101}{-1.545}{-0.678} & -0.954 & 4.6 \\[2pt]
13 & \hyp{The question asks for information based on a provided table of penguins and their attributes} & \sepwithci{-0.115}{-0.727}{0.537} & -0.097 & 6.1 \\[2pt]
14 & \hyp{The question asks to identify the position of an object in a sequence where the correct answer is explicitly stated in the prompt, requiring no actual logical deduction} & \sepwithci{-0.972}{-1.263}{-0.682} & -0.857 & 11.8 \\[2pt]
15 & \hyp{asks to identify the color of an object from a provided list where the color is explicitly stated in the text} & \sepwithci{-1.219}{-1.671}{-0.694} & -1.058 & 3.9 \\[2pt]
16 & \hyp{The question requires identifying an available time slot for an event based on a given schedule of activities and a closing time constraint} & \sepwithci{0.144}{-0.086}{0.372} & 0.122 & 20.7 \\[2pt]
17 & \hyp{The question requires counting the number of animals in a provided text-based table that meet a specific age condition} & \sepwithci{0.192}{-0.268}{0.695} & 0.163 & 1.1 \\[2pt]
18 & \hyp{The question is a logical deduction task that requires ordering five fruits by price based on a set of constraints} & \sepwithci{0.368}{-0.624}{1.328} & 0.313 & 2.5 \\[2pt]
19 & \hyp{The question requires identifying an available time slot in a person's daily schedule to determine when they could have visited a specific location} & \sepwithci{0.144}{-0.084}{0.366} & 0.122 & 20.7 \\[2pt]
20 & \hyp{The question asks to verify the color of a specific object based on a provided list of objects and their colors} & \sepwithci{-1.730}{-2.192}{-1.213} & -1.528 & 3.6 \\[2pt]
21 & \hyp{The question is a logical deduction task that requires deducing the order of a sequence of birds on a branch} & \sepwithci{0.059}{-0.335}{0.471} & 0.050 & 12.9 \\[2pt]
22 & \hyp{The question requires sorting a list of names alphabetically and identifying the name at a specific position in the sorted list} & \sepwithci{2.214}{0.757}{3.671} & 1.905 & 0.7 \\[2pt]
23 & \hyp{The question requires determining the possible time window during which a person could have visited a specific location by eliminating times when they were known to be elsewhere} & \sepwithci{0.144}{-0.084}{0.370} & 0.122 & 20.7 \\[2pt]
24 & \hyp{asks the user to calculate the number of remaining objects of a certain color or type after a specific category of items is removed from a described collection} & \sepwithci{0.466}{-0.225}{1.226} & 0.398 & 5.4 \\[2pt]
25 & \hyp{The question asks to identify the object at a specific position in a sequence, and the correct answer is explicitly stated as a premise in the text} & \sepwithci{-0.965}{-1.227}{-0.699} & -0.858 & 15.0 \\[2pt]
26 & \hyp{The question is a logical deduction task that requires determining the relative order of three colored books on a shelf} & \sepwithci{-1.107}{-1.459}{-0.722} & -0.950 & 2.1 \\[2pt]
27 & \hyp{The question asks to identify the oldest or newest object, and the answer is explicitly stated as a premise in the provided text} & \sepwithci{-1.814}{-2.286}{-1.347} & -1.554 & 0.7 \\[2pt]
28 & \hyp{The correct answer to the question is explicitly stated verbatim in the provided text} & \sepwithci{-1.229}{-1.476}{-0.980} & -1.145 & 18.9 \\[2pt]
29 & \hyp{asks the user to identify the color of an object based on its spatial position relative to another object in a described row} & \sepwithci{0.408}{0.021}{0.752} & 0.348 & 5.7 \\[2pt]
30 & \hyp{The question requires answering a counting or reasoning question based on a provided table of animals and their attributes} & \sepwithci{-0.021}{-0.625}{0.636} & -0.018 & 6.4 \\[2pt]
31 & \hyp{The question is a logical deduction task that requires deducing the order of a sequence of fruits based on their relative prices} & \sepwithci{-0.058}{-0.637}{0.538} & -0.050 & 6.1 \\[2pt]
32 & \hyp{The question is a logical deduction task that requires determining the most expensive fruit among a set of three fruits based on their relative prices} & \sepwithci{-0.360}{-0.974}{0.166} & -0.306 & 1.4 \\[2pt]
33 & \hyp{asks the user to count the number of objects that do not have a specific color to the left of a given object in a sequence} & --- & --- & 0.0 \\[2pt]
34 & \hyp{The question requires extracting numerical data from a provided text-based table and performing arithmetic operations such as averaging, summing, or counting} & \sepwithci{0.350}{-0.121}{0.842} & 0.298 & 2.9 \\[2pt]
35 & \hyp{The question involves answering a query based on a provided table of penguins and their attributes} & \sepwithci{-0.021}{-0.618}{0.626} & -0.018 & 6.4 \\[2pt]
36 & \hyp{is a logical deduction task that requires deducing the finishing order of seven golfers in a tournament based on a set of relative and absolute position constraints} & \sepwithci{0.389}{-0.239}{0.972} & 0.332 & 6.4 \\[2pt]
37 & \hyp{The question asks the user to determine the time window during which a person could have visited a location by finding the unaccounted time in their schedule} & \sepwithci{0.144}{-0.089}{0.370} & 0.122 & 20.7 \\[2pt]
38 & \hyp{The question is a logical deduction task that requires determining the finishing order of three golfers in a tournament based on relative statements} & \sepwithci{-1.035}{-1.345}{-0.696} & -0.891 & 3.6 \\[2pt]
39 & \hyp{involves answering a question based on a provided table of penguins and their attributes} & \sepwithci{-0.115}{-0.715}{0.557} & -0.097 & 6.1 \\[2pt]
40 & \hyp{The question requires extracting information or computing a simple minimum, maximum, or count from a provided text-based table of animals and their attributes} & \sepwithci{-0.093}{-0.712}{0.582} & -0.079 & 6.1 \\[2pt]
41 & \hyp{The question is a logical deduction task that requires deducing the order of a set of fruits based on their relative prices} & \sepwithci{0.005}{-0.606}{0.607} & 0.004 & 5.7 \\[2pt]
42 & \hyp{The question is a logical deduction task that requires deducing the order of a sequence of vehicles based on their relative ages} & \sepwithci{1.366}{0.879}{1.835} & 1.220 & 7.5 \\[2pt]
43 & \hyp{asks to identify the color of an object based on its relative position in a row of objects} & \sepwithci{0.197}{-0.188}{0.574} & 0.168 & 6.8 \\[2pt]
44 & \hyp{The question asks the user to calculate the remaining number of objects of a specific color or type after removing a subset of objects from a given list} & \sepwithci{0.466}{-0.226}{1.224} & 0.398 & 5.4 \\[2pt]
45 & \hyp{The question is a logical deduction task that requires determining the oldest or newest of three vehicles based on their relative ages} & \sepwithci{-1.103}{-1.515}{-0.648} & -0.945 & 1.8 \\[2pt]
46 & \hyp{The question is a logical deduction task that requires determining the relative order of three birds on a branch} & \sepwithci{-0.841}{-1.300}{-0.359} & -0.722 & 3.9 \\[2pt]
47 & \hyp{The question is a logical deduction task that requires determining the order of five birds on a branch based on a set of relative positional clues} & \sepwithci{1.150}{0.703}{1.643} & 0.987 & 2.1 \\[2pt]
\end{longtable}
\endgroup

\clearpage
\begingroup
\scriptsize
\setlength{\tabcolsep}{3pt}
\renewcommand{\arraystretch}{1.14}
\setlength{\LTcapwidth}{\textwidth}
\begin{longtable}{@{}c>{\raggedright\arraybackslash}p{0.56\textwidth}ccc@{}}
\caption{All 37 hypotheses exported by the Edisa baseline on Winogrande with a maximum of 50 hypotheses, in exported order. Repeated hypothesis texts are retained. Statistics use all 254 test items. Negative SEP indicates lower mean difficulty among matched items. SEP is the mean IRT difficulty of matched minus unmatched test items (95\% bootstrap CI); Cohen's $d$ is the standardized difference, and coverage is the matched-item percentage. Undefined statistics are shown as ---; CI and $d$ are omitted when either group has fewer than two items.}\label{tab:all-hypotheses-gmm-winogrande}\\
\toprule
\textbf{\#} & \multicolumn{1}{c}{\textbf{Hypotheses}} &
$\mathbf{SEP}_{\scriptscriptstyle[95\%\,\mathrm{CI}]}$ &
\textbf{Cohen's $d$} & \textbf{Cov. (\%)} \\
\midrule

\multicolumn{5}{c}{\textbf{Winogrande}}\\\midrule
\endfirsthead
\multicolumn{5}{l}{\footnotesize Edisa -- Winogrande (continued)}\\
\toprule
\textbf{\#} & \multicolumn{1}{c}{\textbf{Hypotheses}} &
$\mathbf{SEP}_{\scriptscriptstyle[95\%\,\mathrm{CI}]}$ &
\textbf{Cohen's $d$} & \textbf{Cov. (\%)} \\
\midrule

\endhead
\midrule\multicolumn{5}{r}{\scriptsize Continued on next page}\\\endfoot
\bottomrule\endlastfoot
1 & \hyp{The question asks the model to choose between two options to fill in a blank or resolve a pronoun in a given sentence based on contextual clues} & --- & --- & 100.0 \\[2pt]
2 & \hyp{The question asks the user to choose the correct entity from two options to fill in a blank in a given sentence} & --- & --- & 100.0 \\[2pt]
3 & \hyp{The question requires resolving a missing word or pronoun in a sentence by choosing between two provided options} & --- & --- & 100.0 \\[2pt]
4 & \hyp{The question requires resolving an ambiguous pronoun or blank in a sentence to one of two given noun options} & --- & --- & 100.0 \\[2pt]
5 & \hyp{The question requires resolving a pronoun or blank in a sentence by choosing between two provided options, similar to a Winograd Schema Challenge} & --- & --- & 100.0 \\[2pt]
6 & \hyp{The question asks the user to choose which of two given options correctly fills in a blank to resolve a coreference in a provided sentence} & --- & --- & 100.0 \\[2pt]
7 & \hyp{The question asks the model to resolve a blank in a sentence to one of two given entities mentioned in the sentence} & --- & --- & 100.0 \\[2pt]
8 & \hyp{The question asks the model to choose the correct entity from two options to fill in a blank in a given sentence, testing pronoun or entity resolution} & --- & --- & 100.0 \\[2pt]
9 & \hyp{The question asks the user to choose the right option to fill in a blank in a given sentence from two provided options} & --- & --- & 100.0 \\[2pt]
10 & \hyp{The question asks to fill in a blank in a sentence by choosing between two given options} & --- & --- & 100.0 \\[2pt]
11 & \hyp{asks the model to choose the correct option to fill in a blank in a sentence} & --- & --- & 100.0 \\[2pt]
12 & \hyp{The question asks the model to choose the correct word from two provided options to fill in a blank in a given sentence} & --- & --- & 100.0 \\[2pt]
13 & \hyp{The question requires resolving an ambiguous pronoun or reference in a sentence to one of two given options} & \sepwithci{-0.262}{-1.970}{1.421} & -0.155 & 98.4 \\[2pt]
14 & \hyp{The question asks the user to choose which of two options correctly fills in a blank in a given sentence} & --- & --- & 100.0 \\[2pt]
15 & \hyp{The question asks the user to choose the correct option to fill in a blank or resolve a reference in a given sentence from two provided options} & --- & --- & 100.0 \\[2pt]
16 & \hyp{The question asks the model to fill in a blank in a sentence by choosing between two provided entities, testing coreference resolution} & \sepwithci{0.003}{-1.546}{1.569} & 0.002 & 99.2 \\[2pt]
17 & \hyp{The question asks to fill in a blank in a sentence by choosing between two entities mentioned in the sentence} & --- & --- & 100.0 \\[2pt]
18 & \hyp{The question asks the model to resolve a missing reference or pronoun in a sentence by choosing between two given names} & \sepwithci{-0.114}{-0.558}{0.321} & -0.068 & 59.4 \\[2pt]
19 & \hyp{The question requires filling in a blank in a sentence by choosing between two provided options} & --- & --- & 100.0 \\[2pt]
20 & \hyp{The question requires filling in a blank in a sentence by choosing between two provided options} & --- & --- & 100.0 \\[2pt]
21 & \hyp{The question requires filling in a blank in a sentence by choosing the correct entity from two provided options based on contextual clues} & --- & --- & 100.0 \\[2pt]
22 & \hyp{The question asks the model to choose the right option to fill in a blank ('\_') in a given sentence from two provided options} & --- & --- & 100.0 \\[2pt]
23 & \hyp{The question requires filling in a blank in a sentence by choosing between two provided options} & --- & --- & 100.0 \\[2pt]
24 & \hyp{The question requires filling in a blank in a sentence by choosing between two provided options} & --- & --- & 100.0 \\[2pt]
25 & \hyp{The question asks the user to resolve a blank in a sentence by choosing between two given options, testing pronoun or entity resolution} & \sepwithci{0.003}{-1.553}{1.547} & 0.002 & 99.2 \\[2pt]
26 & \hyp{The question asks the user to choose the correct option to fill in a blank in a given sentence} & --- & --- & 100.0 \\[2pt]
27 & \hyp{The question asks the model to fill in a blank in a sentence by choosing between two entities mentioned in the sentence} & --- & --- & 100.0 \\[2pt]
28 & \hyp{The question asks the user to choose the right option to fill in a blank in a given sentence from two provided options} & --- & --- & 100.0 \\[2pt]
29 & \hyp{The question asks to fill in a blank in a sentence by choosing between two given options} & --- & --- & 100.0 \\[2pt]
30 & \hyp{The question asks the model to resolve an ambiguous reference or missing word in a sentence by choosing between two provided options} & --- & --- & 100.0 \\[2pt]
31 & \hyp{The question asks the user to choose the correct option to fill in a blank in a given sentence from two provided noun or name options} & --- & --- & 100.0 \\[2pt]
32 & \hyp{The question asks the user to choose the correct word from two given options to fill in a blank in a sentence} & --- & --- & 100.0 \\[2pt]
33 & \hyp{The question requires resolving an ambiguous pronoun or blank to one of two entities mentioned in a sentence} & --- & --- & 100.0 \\[2pt]
34 & \hyp{The question asks the model to choose the correct option to fill in a blank in a given sentence, resolving a coreference or logical relationship between two provided options} & --- & --- & 100.0 \\[2pt]
35 & \hyp{The question asks the model to resolve an ambiguous reference or blank in a sentence by choosing between two given options} & --- & --- & 100.0 \\[2pt]
36 & \hyp{The question asks the user to fill in a blank in a sentence by choosing between two provided options} & --- & --- & 100.0 \\[2pt]
37 & \hyp{The question asks the model to fill in a blank in a sentence by choosing between two given options} & --- & --- & 100.0 \\[2pt]
\end{longtable}
\endgroup

\clearpage
\begingroup
\scriptsize
\setlength{\tabcolsep}{3pt}
\renewcommand{\arraystretch}{1.14}
\setlength{\LTcapwidth}{\textwidth}
\begin{longtable}{@{}c>{\raggedright\arraybackslash}p{0.56\textwidth}ccc@{}}
\caption{The 40 hypotheses selected by our main-table model on GSM8K, in selection order. SEP is the mean IRT difficulty of matched minus unmatched test items (95\% bootstrap CI); Cohen's $d$ is the standardized difference, and coverage is the matched-item percentage. Undefined statistics are shown as ---; CI and $d$ are omitted when either group has fewer than two items.}\label{tab:all-hypotheses-ours-gsm8k}\\
\toprule
\textbf{\#} & \multicolumn{1}{c}{\textbf{Hypotheses}} &
$\mathbf{SEP}_{\scriptscriptstyle[95\%\,\mathrm{CI}]}$ &
\textbf{Cohen's $d$} & \textbf{Cov. (\%)} \\
\midrule

\multicolumn{5}{c}{\textbf{GSM8K}}\\\midrule
\endfirsthead
\multicolumn{5}{l}{\footnotesize Ours -- GSM8K (continued)}\\
\toprule
\textbf{\#} & \multicolumn{1}{c}{\textbf{Hypotheses}} &
$\mathbf{SEP}_{\scriptscriptstyle[95\%\,\mathrm{CI}]}$ &
\textbf{Cohen's $d$} & \textbf{Cov. (\%)} \\
\midrule

\endhead
\midrule\multicolumn{5}{r}{\scriptsize Continued on next page}\\\endfoot
\bottomrule\endlastfoot
1 & \hyp{The question contains implicit information or requires the solver to infer a missing value from a combination of given constraints (e.g., working backwards from a total or a difference).} & \sepwithci{1.112}{0.546}{1.703} & 0.623 & 15.0 \\[2pt]
2 & \hyp{The question involves multi-step bookkeeping where intermediate quantities must be calculated and then used as inputs for subsequent calculations, rather than a simple linear sequence of operations.} & \sepwithci{1.498}{1.043}{1.954} & 0.899 & 51.0 \\[2pt]
3 & \hyp{The reference answer contains a logical error or contradiction, such as ignoring a stated value and summing the wrong numbers, making it difficult for a correct model to match the expected output.} & \sepwithci{3.696}{3.037}{4.361} & 2.063 & 1.0 \\[2pt]
4 & \hyp{The question requires more than three distinct arithmetic operations to reach the final answer.} & \sepwithci{0.735}{0.185}{1.311} & 0.408 & 26.0 \\[2pt]
5 & \hyp{Requires at least five distinct arithmetic operations to reach the final answer.} & \sepwithci{1.147}{0.567}{1.732} & 0.648 & 20.0 \\[2pt]
6 & \hyp{The question requires setting up and solving an algebraic equation with an unknown variable, rather than just performing a sequence of arithmetic operations.} & \sepwithci{1.428}{0.482}{2.338} & 0.790 & 4.0 \\[2pt]
7 & \hyp{The question requires tracking and comparing multiple entities across different timeframes or conditions (e.g., Jon vs. James, James vs. cousin, super ticket vs. regular ticket).} & \sepwithci{0.101}{-1.119}{1.284} & 0.055 & 5.0 \\[2pt]
8 & \hyp{The question contains complex sentence structures with multiple conditions, nested clauses, or information presented out of chronological order.} & $0.507$$_{[\text{---}]}$ & --- & 0.5 \\[2pt]
9 & \hyp{The question involves rates or unit conversions that are interrupted or require combining multiple different units of measure (e.g., gigabytes per minute with a time interruption, or feet per second combined with spacing in feet).} & \sepwithci{2.216}{1.559}{2.877} & 1.220 & 1.0 \\[2pt]
10 & \hyp{The question involves multi-step unit conversions or rate calculations (e.g., converting between hours and minutes, or calculating rates based on other rates).} & \sepwithci{1.284}{0.451}{2.212} & 0.713 & 6.5 \\[2pt]
11 & \hyp{The question requires calculating a rate, ratio, or percentage from given quantities, rather than just applying a given rate or percentage.} & \sepwithci{2.035}{1.331}{2.776} & 1.174 & 9.0 \\[2pt]
12 & \hyp{Requires calculating a rate or unit cost from a given quantity and time/amount before applying it to a new quantity.} & \sepwithci{2.078}{1.095}{3.109} & 1.176 & 5.5 \\[2pt]
13 & \hyp{Requires spatial reasoning or understanding the physical layout of an object (e.g., pages on a folded sheet of paper).} & \sepwithci{1.029}{0.045}{2.003} & 0.563 & 1.0 \\[2pt]
14 & \hyp{The question contains a distractor number or a number that is not directly used in the primary calculation but defines a constraint (e.g., 'six groups of equal size' where the size itself isn't needed, just the number of groups).} & \sepwithci{1.433}{0.493}{2.385} & 0.803 & 8.5 \\[2pt]
15 & \hyp{The question requires tracking and updating multiple distinct categories or sub-categories of items across a sequence of events.} & \sepwithci{2.096}{0.422}{4.568} & 1.157 & 1.5 \\[2pt]
16 & \hyp{The question involves calculating a rate of change or unit rate based on a comparative relationship (e.g., '4 weeks to lose the same amount... in a single week').} & \sepwithci{1.909}{0.813}{3.090} & 1.075 & 5.5 \\[2pt]
17 & \hyp{Requires calculating a difference between an estimated/expected value and a true/actual value.} & $4.605$$_{[\text{---}]}$ & --- & 0.5 \\[2pt]
18 & \hyp{The question involves unit conversions or geometric calculations (e.g., volume, area) before the main arithmetic can be completed.} & \sepwithci{-0.011}{-0.716}{0.743} & -0.006 & 3.5 \\[2pt]
19 & \hyp{Requires calculating total cost and total revenue to find profit.} & \sepwithci{0.966}{0.431}{1.581} & 0.529 & 2.0 \\[2pt]
20 & \hyp{Requires calculating percentages of a total amount.} & \sepwithci{0.166}{-0.561}{0.930} & 0.091 & 11.5 \\[2pt]
21 & \hyp{Requires converting between minutes and hours to aggregate total time across different activities.} & --- & --- & 0.0 \\[2pt]
22 & \hyp{The question requires calculating a final remaining amount or difference from a stated initial capacity, total, or budget.} & \sepwithci{-0.036}{-0.747}{0.744} & -0.019 & 14.5 \\[2pt]
23 & \hyp{The question contains distractor information or complex conditional logic that must be parsed to determine which numbers are actually relevant to the final calculation (e.g., a lawyer donating three times the sum of specific other donations).} & \sepwithci{0.549}{-0.418}{1.515} & 0.301 & 3.0 \\[2pt]
24 & \hyp{The question requires working backwards from a final state or a target average to find an initial value or a missing component (e.g., finding the initial number of fries, the initial points, or the required score on a final test).} & \sepwithci{1.155}{0.411}{1.957} & 0.637 & 4.5 \\[2pt]
25 & \hyp{The question involves calculating a final value based on a combination of fixed costs/quantities and variable costs/quantities (e.g., studio rental fee plus per-student fee, base salary plus per-mile rate and bonus).} & \sepwithci{1.388}{0.285}{2.723} & 0.763 & 2.0 \\[2pt]
26 & \hyp{The question requires calculating a net profit or rate by subtracting an ongoing cost from an ongoing revenue before dividing a fixed cost by the net rate.} & --- & --- & 0.0 \\[2pt]
27 & \hyp{Requires converting between different units of time (such as seconds and minutes) to calculate a total duration.} & $-1.456$$_{[\text{---}]}$ & --- & 0.5 \\[2pt]
28 & \hyp{Requires tracking and calculating values for four or more distinct entities or categories.} & \sepwithci{1.408}{0.589}{2.254} & 0.787 & 8.0 \\[2pt]
29 & \hyp{The question requires calculating a percentage of a quantity.} & \sepwithci{0.232}{-0.449}{0.933} & 0.127 & 14.5 \\[2pt]
30 & \hyp{Requires adding an implicit +1 to account for the host or the original subject in a count of people.} & $1.305$$_{[\text{---}]}$ & --- & 0.5 \\[2pt]
31 & \hyp{The question requires working backwards from a total or a final state to find an intermediate or initial value, or solving an implicit algebraic equation where the unknown is not just the final result of a forward calculation.} & \sepwithci{0.884}{0.273}{1.516} & 0.488 & 9.5 \\[2pt]
32 & \hyp{The question requires setting up an algebraic equation with a fractional coefficient (e.g., 2/3 * x = 12) to solve for an unknown.} & \sepwithci{2.797}{2.376}{3.168} & 1.556 & 1.5 \\[2pt]
33 & \hyp{Requires calculating a baseline allowance or threshold, then determining the quantity of items that exceed it, and applying a rate only to the excess.} & \sepwithci{0.705}{0.089}{1.350} & 0.389 & 14.5 \\[2pt]
34 & \hyp{Requires calculating a value based on a comparative phrase involving both a multiplier and an offset (e.g., 'more than twice the amount' or 'more than 5 times as much').} & \sepwithci{-0.023}{-0.912}{0.902} & -0.013 & 4.0 \\[2pt]
35 & \hyp{Requires arithmetic operations with negative numbers.} & $0.894$$_{[\text{---}]}$ & --- & 0.5 \\[2pt]
36 & \hyp{The question requires calculating an intermediate value that is not explicitly asked for but is necessary to find the final answer, often involving a sequence of dependent calculations where the output of one step is the input to the next (e.g., calculating the number of people at different times of day to find the total, calculating the time taken for different parts of a triathlon to find the final run time).} & \sepwithci{0.038}{-0.912}{0.954} & 0.021 & 93.5 \\[2pt]
37 & \hyp{The question requires calculating a percentage or fraction of a previously calculated or updated value, rather than a static initial value.} & \sepwithci{0.120}{-0.632}{0.951} & 0.066 & 13.5 \\[2pt]
38 & \hyp{Requires calculating a remaining budget after subtracting costs for four or more distinct line items.} & $3.561$$_{[\text{---}]}$ & --- & 0.5 \\[2pt]
39 & \hyp{Requires interpreting a rate applied to a fractional or partial unit (e.g., 'or part thereof').} & --- & --- & 0.0 \\[2pt]
40 & \hyp{The question involves a multi-step chain of reasoning where the result of one calculation is used as the input for the next calculation, often involving more than three distinct steps.} & \sepwithci{1.651}{0.984}{2.315} & 0.957 & 16.0 \\[2pt]
\end{longtable}
\endgroup

\clearpage
\begingroup
\scriptsize
\setlength{\tabcolsep}{3pt}
\renewcommand{\arraystretch}{1.14}
\setlength{\LTcapwidth}{\textwidth}
\begin{longtable}{@{}c>{\raggedright\arraybackslash}p{0.56\textwidth}ccc@{}}
\caption{The 12 hypotheses selected by our main-table model on BBH-structured, in selection order. SEP is the mean IRT difficulty of matched minus unmatched test items (95\% bootstrap CI); Cohen's $d$ is the standardized difference, and coverage is the matched-item percentage. Undefined statistics are shown as ---; CI and $d$ are omitted when either group has fewer than two items.}\label{tab:all-hypotheses-ours-bbh-structured}\\
\toprule
\textbf{\#} & \multicolumn{1}{c}{\textbf{Hypotheses}} &
$\mathbf{SEP}_{\scriptscriptstyle[95\%\,\mathrm{CI}]}$ &
\textbf{Cohen's $d$} & \textbf{Cov. (\%)} \\
\midrule

\multicolumn{5}{c}{\textbf{BBH-structured}}\\\midrule
\endfirsthead
\multicolumn{5}{l}{\footnotesize Ours -- BBH-structured (continued)}\\
\toprule
\textbf{\#} & \multicolumn{1}{c}{\textbf{Hypotheses}} &
$\mathbf{SEP}_{\scriptscriptstyle[95\%\,\mathrm{CI}]}$ &
\textbf{Cohen's $d$} & \textbf{Cov. (\%)} \\
\midrule

\endhead
\midrule\multicolumn{5}{r}{\scriptsize Continued on next page}\\\endfoot
\bottomrule\endlastfoot
1 & \hyp{The question asks for a fact, property, or position that is explicitly stated in the text (e.g., 'The plums are the cheapest', 'What color is the textbook?'), requiring only direct lookup rather than logical deduction.} & \sepwithci{-1.278}{-1.521}{-1.032} & -1.206 & 20.0 \\[2pt]
2 & \hyp{The question asks to identify an entity at a specific non-extreme ordinal position (e.g., 'second-to-last', 'third from the right') rather than at an extreme end (e.g., 'leftmost', 'cheapest').} & \sepwithci{0.565}{0.257}{0.875} & 0.492 & 28.2 \\[2pt]
3 & \hyp{The question requires sorting a list of strings (e.g., names) alphabetically to determine the answer, rather than performing a simple numerical minimum/maximum operation or direct lookup.} & \sepwithci{2.214}{0.757}{3.670} & 1.905 & 0.7 \\[2pt]
4 & \hyp{The question requires identifying an object based on its relative position to another specific object (e.g., 'directly to the right of', 'furthest from') rather than its absolute position in the sequence (e.g., 'left-most', 'right-most').} & \sepwithci{0.505}{0.118}{0.888} & 0.432 & 6.1 \\[2pt]
5 & \hyp{The question requires reasoning about relative spatial or temporal positions (e.g., 'to the right of', 'between what times') rather than absolute attributes (e.g., 'is the jug mauve?').} & \sepwithci{0.738}{0.462}{1.009} & 0.658 & 62.1 \\[2pt]
6 & \hyp{The question requires performing an operation over multiple elements, such as finding an unaccounted time gap in a schedule, summing values in a table, or counting rows after a deletion, rather than retrieving a single stated value.} & \sepwithci{0.449}{0.187}{0.718} & 0.387 & 27.5 \\[2pt]
7 & \hyp{The problem involves a larger number of entities (e.g., seven objects) to order or track, increasing the state space compared to problems with fewer entities (e.g., three objects).} & \sepwithci{1.297}{0.944}{1.650} & 1.189 & 13.2 \\[2pt]
8 & \hyp{The question requires counting or performing arithmetic operations on objects after applying a filtering condition.} & \sepwithci{0.521}{0.107}{0.954} & 0.447 & 10.7 \\[2pt]
9 & \hyp{The question requires sorting items alphabetically or numerically before identifying a specific item, rather than retrieving an item based on its original order or explicit value.} & \sepwithci{2.214}{0.758}{3.671} & 1.905 & 0.7 \\[2pt]
10 & \hyp{The question requires deducing the answer by combining multiple constraints or steps, whereas the easier question asks for a fact that is explicitly stated in the text.} & \sepwithci{-0.179}{-1.071}{0.705} & -0.152 & 3.6 \\[2pt]
11 & \hyp{Questions are harder when the answer choices require distinguishing between many plausible options (e.g., 7 to 17 options) rather than a small set of options (e.g., 2 to 5 options).} & \sepwithci{0.645}{0.331}{0.951} & 0.566 & 29.3 \\[2pt]
12 & \hyp{The logical deduction task requires ordering a sequence of five or seven objects, rather than a simple sequence of only three objects.} & \sepwithci{1.366}{1.081}{1.635} & 1.335 & 23.6 \\[2pt]
\end{longtable}
\endgroup

\clearpage
\begingroup
\scriptsize
\setlength{\tabcolsep}{3pt}
\renewcommand{\arraystretch}{1.14}
\setlength{\LTcapwidth}{\textwidth}
\begin{longtable}{@{}c>{\raggedright\arraybackslash}p{0.56\textwidth}ccc@{}}
\caption{The 44 hypotheses selected by our main-table model on Winogrande, in selection order. SEP is the mean IRT difficulty of matched minus unmatched test items (95\% bootstrap CI); Cohen's $d$ is the standardized difference, and coverage is the matched-item percentage. Undefined statistics are shown as ---; CI and $d$ are omitted when either group has fewer than two items.}\label{tab:all-hypotheses-ours-winogrande}\\
\toprule
\textbf{\#} & \multicolumn{1}{c}{\textbf{Hypotheses}} &
$\mathbf{SEP}_{\scriptscriptstyle[95\%\,\mathrm{CI}]}$ &
\textbf{Cohen's $d$} & \textbf{Cov. (\%)} \\
\midrule

\multicolumn{5}{c}{\textbf{Winogrande}}\\\midrule
\endfirsthead
\multicolumn{5}{l}{\footnotesize Ours -- Winogrande (continued)}\\
\toprule
\textbf{\#} & \multicolumn{1}{c}{\textbf{Hypotheses}} &
$\mathbf{SEP}_{\scriptscriptstyle[95\%\,\mathrm{CI}]}$ &
\textbf{Cohen's $d$} & \textbf{Cov. (\%)} \\
\midrule

\endhead
\midrule\multicolumn{5}{r}{\scriptsize Continued on next page}\\\endfoot
\bottomrule\endlastfoot
1 & \hyp{The sentence structure involves a complex causal chain where the blank refers to an entity that is the cause of a state described earlier in the sentence, requiring the model to infer the cause from the effect.} & \sepwithci{0.676}{-0.262}{1.729} & 0.402 & 4.3 \\[2pt]
2 & \hyp{The easier question can be solved using strong lexical associations, common collocations, or direct causal links (e.g., 'tough stain', 'IOS' to 'iPhone', 'lazy' to 'failed').} & \sepwithci{-0.889}{-1.583}{-0.069} & -0.532 & 7.5 \\[2pt]
3 & \hyp{The question requires multi-step causal reasoning to connect the premise to the resolution, where the solver must infer an unstated intermediate state or action (e.g., hating a food means eating less of it, which leaves more to harvest; or accepting a challenge implies believing one can win), rather than relying on direct semantic associations or attribute matching (e.g., ice is cold, or an animal rights activist dislikes leather).} & \sepwithci{0.916}{0.449}{1.393} & 0.559 & 26.4 \\[2pt]
4 & \hyp{The question can be resolved using basic selectional restrictions or inherent physical properties of the entities (e.g., a cloth is soft, a video is boring, a color is bright) rather than requiring causal reasoning about the events described.} & \sepwithci{-0.872}{-1.456}{-0.220} & -0.524 & 11.8 \\[2pt]
5 & \hyp{The harder question involves resolving a pronoun based on an action that caused regret or a negative emotional reaction (e.g., feeling bad and vowing not to do it again).} & --- & --- & 0.0 \\[2pt]
6 & \hyp{The harder question requires reasoning about implicit physical, biological, or social mechanisms (e.g., relative hardness for scratching, evergreen vs. deciduous traits, professional vs. amateur competence, or the implications of acting in someone else's name), whereas the easier question relies on direct lexical associations or common sense (e.g., library for a paper, productive means hard worker).} & --- & --- & 0.0 \\[2pt]
7 & \hyp{The question requires multi-step causal reasoning where the blank is resolved by inferring an implicit intermediate state (e.g., fewer items means they take up more space, or asking about a wedding implies the other person is the groom).} & \sepwithci{0.386}{-0.038}{0.817} & 0.230 & 31.9 \\[2pt]
8 & \hyp{Questions are harder when the causal relationship requires inferring a missing intermediate step or implicit consequence (e.g., 'cows hated eating the corn' -> 'farmer had more corn to harvest', 'sunglasses were poor' -> 'eyes could not see clearly') rather than a direct, explicit cause-and-effect relationship.} & \sepwithci{0.427}{0.026}{0.835} & 0.255 & 46.9 \\[2pt]
9 & \hyp{The question requires complex social or interpersonal reasoning (such as inferring roles, intentions, or dynamics) rather than relying on direct physical properties or straightforward cause-and-effect.} & \sepwithci{0.881}{0.257}{1.503} & 0.531 & 13.8 \\[2pt]
10 & \hyp{The sentence contains a semantic contradiction or counter-intuitive resolution where the correct referent violates typical commonsense expectations or requires strict adherence to syntactic structure over semantic plausibility.} & --- & --- & 0.0 \\[2pt]
11 & \hyp{The question requires resolving the pronoun or blank based on a counter-intuitive, anti-stereotypical, or unexpected causal relationship (e.g., having more of a crop because animals hated eating it, or an item failing to fit because the inner layer was too large rather than the outer layer being too small).} & \sepwithci{2.640}{0.569}{4.557} & 1.588 & 1.2 \\[2pt]
12 & \hyp{The harder question requires understanding implicit social, economic, or legal relationships (e.g., the cost implications of a gift vs. a card, or the distinction between a homeowner and a tenant).} & \sepwithci{-0.448}{-1.329}{0.426} & -0.266 & 0.8 \\[2pt]
13 & \hyp{The question can be answered using direct lexical or semantic associations between the clue and the correct option (e.g., 'chainsaw' and 'tree cutter', 'wood' and 'rotten', 'video' and 'boring'), which makes it easier, whereas harder questions require reasoning about the specific situational dynamics or causal relationships described in the sentence.} & \sepwithci{-0.571}{-1.043}{-0.083} & -0.342 & 22.8 \\[2pt]
14 & \hyp{The harder question requires resolving a reference based on explicit possession or attribution in the text rather than general causal reasoning about traits.} & --- & --- & 0.0 \\[2pt]
15 & \hyp{The easier question can be solved by directly matching a straightforward descriptive trait, profession, or physical state to the person exhibiting the corresponding behavior, without needing to infer complex interactions.} & \sepwithci{-0.604}{-1.176}{0.069} & -0.360 & 7.5 \\[2pt]
16 & \hyp{The easier question uses explicit contrast structures (e.g., 'instead of', 'unlike') where the blank is resolved by mapping a straightforward positive or negative trait directly to the accepted or rejected entity.} & \sepwithci{0.398}{-0.825}{2.399} & 0.236 & 1.2 \\[2pt]
17 & \hyp{The question requires multi-step causal or physical reasoning about the implicit properties of objects or actions (e.g., that bronzer is a fast way to tan, or that soaking an object transfers liquid from the container) rather than relying on direct lexical associations.} & \sepwithci{0.266}{-0.178}{0.720} & 0.158 & 30.7 \\[2pt]
18 & \hyp{The question requires complex situational or social commonsense reasoning to infer roles and goals (e.g., understanding what a 'ride along' implies or the implications of doing something 'in place of' someone else).} & \sepwithci{0.236}{-0.486}{0.998} & 0.140 & 9.4 \\[2pt]
19 & \hyp{The question requires inferring an implicit goal or intention behind an action involving multiple agents, locations, or domains, rather than relying on explicit causal markers.} & \sepwithci{-0.516}{-1.117}{0.067} & -0.306 & 2.4 \\[2pt]
20 & \hyp{The question requires multi-step reasoning or specific domain knowledge to connect the clue to the correct entity (e.g., knowing that 'saccharine' means sweet and fruits are sweet, or that snowboarding requires going up a hill, or that commodities are bought when priced low), rather than relying on direct, everyday commonsense associations (e.g., returning an item because it is the wrong size, or changing a filter because it is dirty).} & \sepwithci{0.319}{-0.345}{1.003} & 0.189 & 8.7 \\[2pt]
21 & \hyp{The question requires reasoning about physical, spatial, or quantitative relationships where the correct answer depends on an inverse correlation (e.g., having more of one item implies the other item takes up more space, or transferring a substance to one object means the source object has less).} & \sepwithci{0.832}{0.145}{1.530} & 0.498 & 10.2 \\[2pt]
22 & \hyp{The easier question can be resolved largely through direct semantic selectional restrictions or strong lexical associations between the predicate and the correct entity (e.g., a video is 'boring', a street is 'long', birds fly to 'bird seed'), without needing to integrate the full causal logic of the sentence.} & \sepwithci{-0.998}{-1.857}{0.096} & -0.597 & 5.9 \\[2pt]
23 & \hyp{The harder questions often require the model to infer the antecedent based on the logical consequence of an action that is not explicitly stated but must be deduced from world knowledge (e.g., a person who fled a raid avoids arrest, a person who asks a question lacks knowledge).} & \sepwithci{0.627}{0.214}{1.036} & 0.378 & 52.4 \\[2pt]
24 & \hyp{The easier question relies on direct physical constraints, basic definitions, or highly conventional event scripts (e.g., a door being too small for a bed, a cashier making less money, returning an item to the person who loaned it).} & \sepwithci{-0.762}{-2.075}{1.423} & -0.452 & 1.6 \\[2pt]
25 & \hyp{The sentence structure involves a comparison where the blank refers to the entity that is less prominent or requires reasoning about the negative/lesser side of the comparison.} & \sepwithci{0.170}{-0.493}{0.865} & 0.101 & 9.4 \\[2pt]
26 & \hyp{The correct resolution of the blank violates typical semantic or physical expectations, such as a person being 'too short to climb over' a fence rather than the fence being too short, forcing the model to override strong lexical priors.} & \sepwithci{2.640}{0.567}{4.539} & 1.588 & 1.2 \\[2pt]
27 & \hyp{The question requires a multi-step inference combining a stated premise with implicit background knowledge about a role or action (e.g., inferring that a worse baker makes a mess, or that a cashier accepts real bills and real bills have sharp lines).} & \sepwithci{0.445}{0.023}{0.859} & 0.266 & 36.6 \\[2pt]
28 & \hyp{The easier question contains explicit causal conjunctions ('because', 'since', 'as') that directly link a stated preference, action, or outcome to an attribute of the correct answer.} & \sepwithci{-0.659}{-1.349}{0.088} & -0.393 & 9.8 \\[2pt]
29 & \hyp{The harder questions frequently involve implicit causal reasoning where the blank refers to the cause of a negative outcome or failure (e.g., 'was not enough', 'lacked the aptitude', 'was boring'), requiring the model to infer which entity possesses the negative attribute.} & \sepwithci{0.205}{-0.264}{0.699} & 0.121 & 23.2 \\[2pt]
30 & \hyp{The correct resolution contradicts a strong local syntactic or semantic expectation, such as a counter-intuitive physical description ('the thief was too short' rather than the fence) or an unexpected source of a trait.} & $4.577$$_{[\text{---}]}$ & --- & 0.4 \\[2pt]
31 & \hyp{The harder questions require resolving a pronoun or blank to an entity that is the object or patient of the preceding clause, whereas easier questions often resolve to the subject or agent.} & \sepwithci{0.110}{-0.311}{0.518} & 0.065 & 39.4 \\[2pt]
32 & \hyp{The question requires resolving a reference based on implicit social roles, physical constraints, or metonymy rather than direct descriptions (e.g., knowing that a 'best man' implies the other person is the groom, or describing a 'sprain' as 'fat'), which increases the difficulty.} & \sepwithci{0.442}{-0.069}{0.982} & 0.263 & 18.1 \\[2pt]
33 & \hyp{The question requires inverse or counter-intuitive causal reasoning, where the correct answer depends on an indirect relationship (e.g., inferring that having more of one item means the other item is bigger, or that socks being 'too large' prevents boots from fitting, rather than the boots being too small).} & \sepwithci{1.029}{0.460}{1.613} & 0.626 & 16.1 \\[2pt]
34 & \hyp{The correct answer requires resolving a pronoun or implicit reference to the entity that was NOT chosen, NOT affected, or NOT the primary subject of the main clause (e.g., the cheaper item, the person who didn't get sick, the vehicle not driven).} & \sepwithci{-0.225}{-0.744}{0.289} & -0.133 & 15.0 \\[2pt]
35 & \hyp{The easier question contains strong, direct lexical or causal cues linking the blank to the correct entity (e.g., 'returned' strongly implies the other person 'loaned' it; 'scary' directly explains why someone went on a ride less).} & \sepwithci{-1.950}{-2.239}{-1.667} & -1.165 & 1.2 \\[2pt]
36 & \hyp{The harder question involves a potentially counter-intuitive or highly specific causal attribution that defies typical expectations, making the reference resolution much more difficult for models to predict correctly.} & --- & --- & 0.0 \\[2pt]
37 & \hyp{The correct answer requires inferring a contrast or counter-intuitive relationship, such as a concession or an indirect consequence, rather than a direct causal link.} & \sepwithci{0.737}{-0.098}{1.683} & 0.438 & 2.8 \\[2pt]
38 & \hyp{The sentence requires reasoning about a comparative degree or inverse relationship (e.g., 'more than', 'less') to determine which entity possesses the trait that resolves the blank.} & \sepwithci{-0.106}{-0.545}{0.346} & -0.063 & 25.6 \\[2pt]
39 & \hyp{The question requires inverse reasoning or understanding a negative constraint, where the correct referent is identified by its association with an undesirable outcome, a lack of a trait, or an avoided option (e.g., 'instead of', 'unlike', 'sounded quieter', 'too short to climb over').} & \sepwithci{0.266}{-0.193}{0.728} & 0.158 & 25.6 \\[2pt]
40 & \hyp{The sentence involves complex interpersonal dynamics where one person's action or trait directly causes the other person's reaction, requiring the reader to carefully assign distinct roles (e.g., cause vs. experiencer) to resolve the pronoun.} & \sepwithci{0.742}{0.199}{1.292} & 0.447 & 19.7 \\[2pt]
41 & \hyp{The easier question involves resolving a reference based on physical properties or direct physical actions (e.g., melting, falling), whereas the harder question involves abstract states, intentions, or social situations.} & \sepwithci{-0.249}{-1.539}{0.963} & -0.148 & 1.6 \\[2pt]
42 & \hyp{The correct answer requires resolving a pronoun or blank to an entity that is the subject of a subordinate clause, where the causal relationship is counterintuitive or requires overriding a strong local association (e.g., 'cows hated eating the corn' leading to more corn, or 'sprain is too fat' instead of 'foot').} & $4.577$$_{[\text{---}]}$ & --- & 0.4 \\[2pt]
43 & \hyp{The question involves complex causal reasoning about constraints and degrees of difficulty (e.g., deducing that having less money makes a budget harder to follow).} & $2.197$$_{[\text{---}]}$ & --- & 0.4 \\[2pt]
44 & \hyp{The easier question relies on direct lexical associations or straightforward commonsense properties (e.g., cashier = less money, lack of light = dark room, sneaky = coming up behind).} & \sepwithci{-1.340}{-1.629}{-1.022} & -0.798 & 1.6 \\[2pt]
\end{longtable}
\endgroup



\end{document}